\documentclass[letterpaper]{article} 
\usepackage{aaai2027}  
\usepackage[hyphens]{url}  
\usepackage{graphicx} 
\usepackage{natbib}  
\usepackage{caption} 
\usepackage{algorithm}
\usepackage{algorithmic}

\usepackage{newfloat}
\usepackage{listings}
\DeclareCaptionStyle{ruled}{labelfont=normalfont,labelsep=colon,strut=off} 
\floatstyle{ruled}
\newfloat{listing}{tb}{lst}{}
\floatname{listing}{Listing}

\usepackage{booktabs}
\usepackage{longtable}
\usepackage{supertabular}

\usepackage{multirow}
\usepackage{colortbl}
\usepackage{makecell}
\usepackage{array}
\usepackage{amsmath}
\usepackage{amssymb}
\usepackage{textcomp}

\usepackage{capt-of}
\usepackage{pifont}

\usepackage{enumitem}
\usepackage{placeins}

\usepackage{subcaption}

\title{Beyond RGB: Benchmarking and Enhancing MLLMs for Hyperspectral Image Understanding via Training-Free Reasoning Framework}

\author{
    Xinyu Zhang\textsuperscript{\rm 1}\equalcontrib,
    Zurong Mai\textsuperscript{\rm 1}\equalcontrib,
    Qingmei Li\textsuperscript{\rm 2}\equalcontrib,
    Xiaoya Fan\textsuperscript{\rm 5},
    Zjin Liao\textsuperscript{\rm 1},
    Haoyuan Liang\textsuperscript{\rm 1},
    Yibin Wen\textsuperscript{\rm 1},
    Yuhang Chen\textsuperscript{\rm 1},
    Chan Tsz Ho\textsuperscript{\rm 1},
    Bi Tianyuan\textsuperscript{\rm 1},
    Ruifeng Su\textsuperscript{\rm 1},
    Zihao Qiang\textsuperscript{\rm 1},
    Juepeng Zheng\textsuperscript{\rm 1,6}\corresponding,
    Jianxi Huang\textsuperscript{\rm 3,4},
    Yutong Lu\textsuperscript{\rm 1,6},
    Haohuan Fu\textsuperscript{\rm 2,6}
}
\affiliations{
    \textsuperscript{\rm 1}Sun Yat-sen University\\
    \textsuperscript{\rm 2}Tsinghua Shenzhen International Graduate School\\
    \textsuperscript{\rm 3}China Agricultural University\\
    \textsuperscript{\rm 4}Southwest Jiaotong University\\
    \textsuperscript{\rm 5}Southwest University\\
    \textsuperscript{\rm 6}National Supercomputing Center in Shenzhen\\
    \equalcontrib Equal contribution.\\
    \corresponding Corresponding author.
}

\begin{document}

\maketitle

\begin{abstract}
Multimodal Large Language Models (MLLMs) have achieved strong performance on RGB image understanding, yet their ability to use spectral evidence beyond the visible range remains largely unexplored. Hyperspectral imagery (HSI) provides dense spectral measurements that reveal material and environmental cues unavailable in RGB, but current MLLMs cannot directly ingest high-dimensional HSI. To study this gap, we introduce \textbf{HM-Bench}, an evidence-grounded benchmark for hyperspectral image understanding with MLLMs. HM-Bench contains 19,337 question--answer pairs from 2,178 hyperspectral samples across 13 task categories, covering general perception, spectral reasoning, and spatial--spectral reasoning. To make HSI accessible to the native image--text interfaces of existing MLLMs, we further propose \textbf{VSR$^{2}$}, a training-free \underline{\textbf{V}}isual--\underline{\textbf{S}}pectral--\underline{\textbf{R}}eport \underline{\textbf{R}}easoning framework. VSR$^{2}$ represents each HSI sample with three aligned views: an RGB image for visual semantics and spatial context, a PCA-based image for dominant spectral variation, and a structured report for quantitative spectral--spatial evidence. Under a controlled RGB-only versus VSR$^{2}$ evaluation protocol, experiments on 17 representative MLLMs show that HSI information improves average accuracy from 38.90\% to 40.95\%, with gains varying substantially across models and task categories. These results indicate that HSI information (beyond RGB) is useful for current MLLMs, while robust hyperspectral reasoning remains an open challenge.

\end{abstract}


\section{Introduction}
\begin{figure}[t]
  \centering
  \includegraphics[width=\columnwidth]{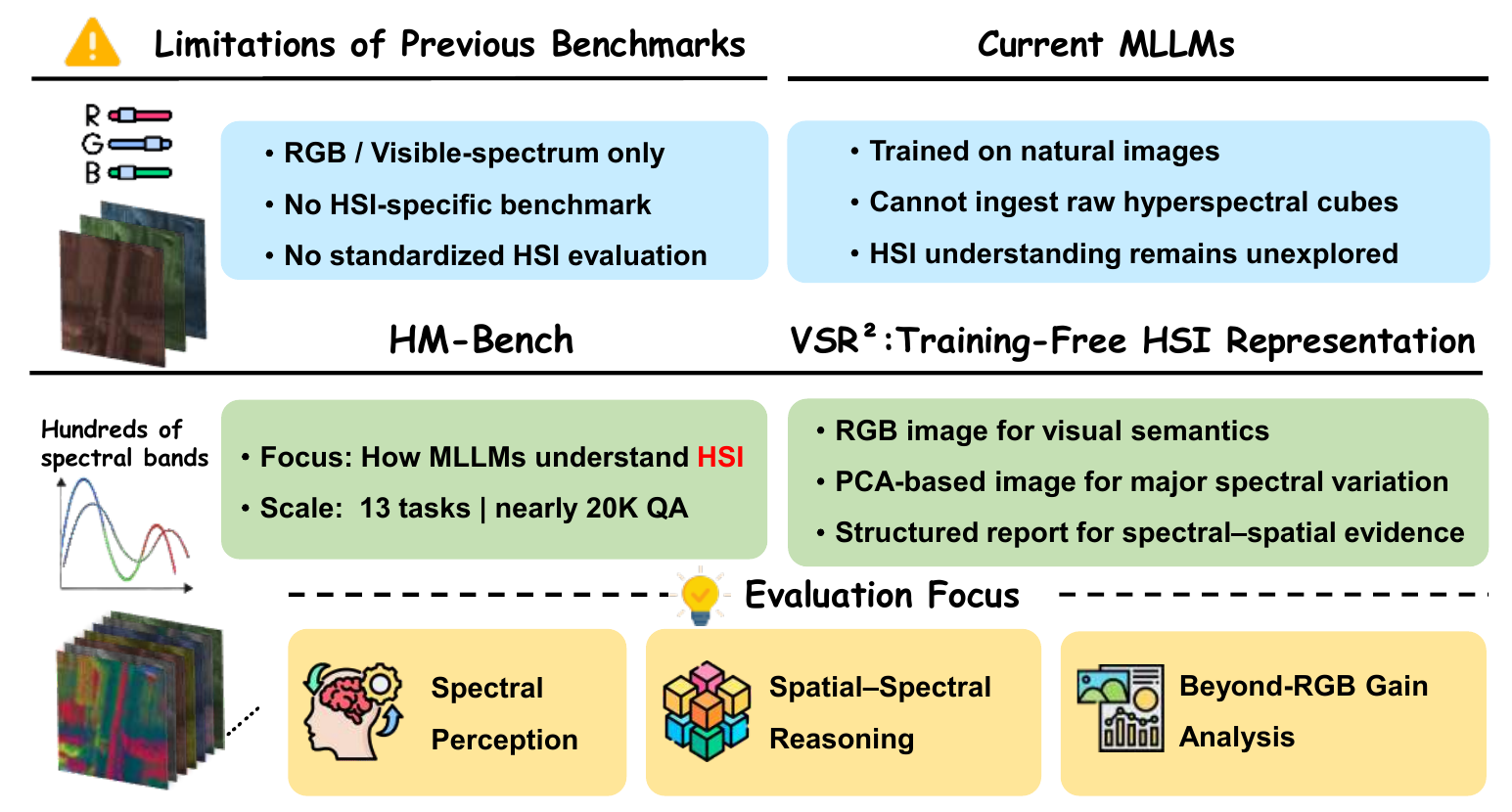}
    \caption{Limitations of current MLLMs and benchmarks in hyperspectral understanding. We tackle these challenges by proposing HM-Bench and exploring VSR$^{2}$.}
  \label{fig:teaser}
\end{figure}

Recent advances in multimodal large language models (MLLMs)~\citep{yin2024survey} have substantially improved general image understanding, but most progress has been built and evaluated on natural RGB images. In contrast, hyperspectral imagery (HSI) is rarely incorporated into the training and evaluation of visual foundation models. HSI captures hundreds of narrow contiguous spectral bands encoding physical signatures related to material composition, vegetation health, water content, and surface properties~\citep{rasti2020feature}. Visually similar materials often yield nearly identical RGB appearances but display divergent reflectance signatures across red-edge, near-infrared, and other non-visible bands. The unique spectral characteristics render HSI a powerful tool for agricultural monitoring~\citep{zhang2016crop}, environmental assessment~\citep{stuart2019hyperspectral}, urban analysis~\citep{hong2021multimodal}, and planetary exploration. Drawing from the above insights, this work targets a fundamental question: \textbf{whether, which tasks, and how HSI information beyond RGB can benefit MLLMs in hyperspectral perception and spatial–spectral reasoning}, as illustrated in Fig.~\ref{fig:teaser}.

To systematically investigate this question, we introduce \textbf{HM-Bench}, an evidence-grounded benchmark comprising 19,337 question–answer pairs from 2,178 hyperspectral samples across 13 task categories and six capability dimensions. Distinct from existing RGB-centric \cite{liu2024mmbench, wang2025xlrs} or task-specific HSI \cite{hong2024spectralgpt, wang2025hypersigma} benchmarks, HM-Bench assesses general perceptual and reasoning abilities of MLLMs. Evaluations cover basic feature recognition, target quantification and spatial localization, as well as advanced challenges including material composition interpretation, status evaluation, and multitemporal change detection. Reliable answers require spectral cues including subtle reflectance variations, absorption signatures and mixed-pixel traits, which cannot be fully captured by RGB imagery. We build the benchmark via an evidence-based hybrid approach. Objective questions stem from calibrated annotations, geometric measurements, and spectral statistics. For reasoning-oriented questions, we use MLLMs for initial generation and then enforce rigorous expert verification and deterministic consistency checks. All answers can thereby be traced to explicit spatial–spectral evidence.

Constructing the benchmark alone does not enable HSI evaluation for current MLLMs, as they cannot ingest raw high-dimensional HSI cubes, and reducing HSI to RGB discards critical information \cite{li2024llava}. To address this, we propose \textbf{VSR$^2$} (\textbf{\underline{V}}isual–\underline{\textbf{S}}pectral–\underline{\textbf{R}}eport \underline{\textbf{R}}easoning), a training-free framework that exposes complementary hyperspectral evidence through native MLLM interfaces. VSR$^2$ represents each sample via three aligned views: a \textit{visual view} (RGB synthesized from visible bands) preserving scene semantics and spatial structure; a \textit{spectral view} (PCA-based composite) capturing primary cross-band variations; and a \textit{report view} (structured text) summarizing global spectral statistics, representative indices, regional differences, and reflectance trends. Lightweight fusion prompts guide the model to examine RGB first, then consult spectral and report views when additional evidence is needed.

Evaluating 17 representative MLLMs under RGB-only and VSR$^2$ settings reveals that HSI understanding depends more on visual representation strength and multimodal reasoning ability than on remote sensing specialization alone. Task-specific analyses further suggest that the three views are complementary to some extent: RGB inputs are generally stronger on visually grounded semantic and spatial tasks, PCA views tend to be more effective for tasks sensitive to cross-band variation, and structured reports provide complementary cues for tasks involving explicit quantitative evidence. Overall, VSR$^2$ improves the average accuracy from 38.90\% to 40.95\%, indicating a measurable benefit from our training-free reasoning framework.

Our main contributions are summarized as follows:
\begin{itemize}[leftmargin=7.5pt]
\item We construct HM-Bench, an evidence-driven hyperspectral benchmark enabling systematic evaluation from fundamental perception to complex spatial-spectral reasoning.
\item We propose VSR$^2$, a training-free framework converting hyperspectral cubes into complementary RGB, PCA, and structured report representations with lightweight prompts for cross-view reasoning.
\item We conduct comprehensive evaluation of 17 MLLMs, revealing whether, where, and how hyperspectral information improves performance, providing insights into training-free hyperspectral understanding.
\end{itemize}

\section{Related Work}
\subsection{Benchmarks for MLLMs}

The rapid progress of MLLMs has led to a wide range of benchmarks for multimodal understanding and reasoning, including MMBench~\citep{liu2024mmbench}, MM-Vet~\citep{yu2024mmvet}, MMMU~\citep{yue2024mmmu}, and MathVista~\citep{lu2024mathvista}. These benchmarks evaluate diverse capabilities such as perception, knowledge grounding, and reasoning, but are built primarily on natural RGB and therefore provide little evidence about model behavior beyond the visible spectrum.

Several recent benchmarks study MLLMs under more challenging visual conditions. HR-Bench~\citep{wang2025hrbench} focuses on high-resolution images, BenchLMM~\citep{cai2024benchlmm} evaluates robustness under cross-style shifts, and IF-Bench~\citep{zhang2026ifbench} targets infrared understanding. Findings from the previous work demonstrate that strong RGB performance does not necessarily transfer to specialized visual domains. Our work extends this line of research to hyperspectral imagery, where the challenge is not only domain shift, but also how to expose beyond-RGB information through the image-text interface of current MLLMs.

\subsection{Remote Sensing MLLMs and Benchmarks}

General-purpose MLLMs have also inspired remote sensing models such as RSGPT~\citep{hu2025rsgpt}, GeoChat~\citep{kuckreja2024geochat}, SkyEyeGPT~\citep{zhan2025skyeyegpt}, EarthGPT~\citep{zhang2024earthgpt}, and LHRS-Bot~\citep{muhtar2024lhrsbot}, which support tasks including captioning, VQA, and grounding. On the benchmark side, VRSBench~\citep{li2024vrsbench}, UrBench~\citep{zhou2025urbench}, AgroBench~\citep{shinoda2025agrobench}, AgroMind~\citep{li2025can}, XLRS-Bench~\citep{wang2025xlrs}, and RSHR-Bench~\citep{dang2025benchmark} evaluate remote sensing reasoning abilities such as localization, object relations, and scene understanding.

Despite these advances, existing remote sensing MLLMs and benchmarks remain centered on RGB imagery. This is a major limitation for modalities such as hyperspectral imagery, which captures dense spectral measurements beyond visible light and provides material cues unavailable in RGB. As summarized in Table~\ref{tab:benchmark_comparison}, hyperspectral understanding requires not only a dedicated benchmark, but also an MLLM-compatible way to present spectral information.

\begin{table}[t]
\centering
\small
\setlength{\tabcolsep}{3pt}
\renewcommand{\arraystretch}{1.0}
\caption{Comparison between existing benchmarks and HM-Bench. ``BR-Eval'' indicates support for beyond-RGB evaluation, and ``TF-Enh'' indicates support for training-free input enhancement for MLLMs.}
\label{tab:benchmark_comparison}
\resizebox{\columnwidth}{!}{%
\begin{tabular}{lccccc}
\toprule
\textbf{Benchmark} & \textbf{Input Modality} & \textbf{QA} & \textbf{Tasks} & \textbf{BR-Eval} & \textbf{TF-Enh} \\
\midrule
MMBench~\citep{liu2024mmbench}        & RGB                 & $\sim$3k & 20 & \ding{55} & \ding{55} \\
VRSBench~\citep{li2024vrsbench}       & RGB                 & 12.3k & 10 & \ding{55} & \ding{55} \\
AgroBench~\citep{shinoda2025agrobench}      & RGB                 & 4.3k  & 7  & \ding{55} & \ding{55} \\
AgroMind~\citep{li2025can}       & RGB                 & 28k   & 13 & \ding{55} & \ding{55} \\
XLRS-Bench~\citep{wang2025xlrs}     & RGB                 & 32k   & 16 & \ding{55} & \ding{55} \\
RSHR-Bench~\citep{dang2025benchmark}     & RGB                 & 8.2k  & 13 & \ding{55} & \ding{55} \\
\midrule
\rowcolor{orange!15}
\textbf{HM-Bench (Ours)}
  & HSI cube $\rightarrow$ 3 views
  & 19.3k & 13 & \ding{51} & \ding{51} \\
\bottomrule
\end{tabular}%
}
\end{table}

\subsection{Hyperspectral Image Understanding}

Hyperspectral image understanding has long been studied in remote sensing, including classification, target detection, anomaly detection, unmixing, and change detection~\citep{li2019deep}. Because hyperspectral data is high-dimensional and spectrally rich, effective analysis typically relies on joint spatial-spectral modeling. Deep models such as CNNs, Transformers~\citep{hong2021spectralformer},and recent foundation models including SpectralGPT~\citep{hong2024spectralgpt} and HyperSIGMA~\citep{wang2025hypersigma} have achieved strong performance on downstream HSI tasks.

However, these methods are designed for task-specific prediction on hyperspectral cubes rather than open-ended language interaction. They are therefore not directly suitable for evaluating general-purpose MLLMs. Moreover, raw hyperspectral data is incompatible with current image-text interfaces, making input representation a key challenge. In contrast, HM-Bench and VSR$^{2}$ provide a dedicated benchmark and a training-free protocol for exposing hyperspectral information to existing MLLMs.

\begin{figure}[ht]
  \centering
  \includegraphics[width=\columnwidth]{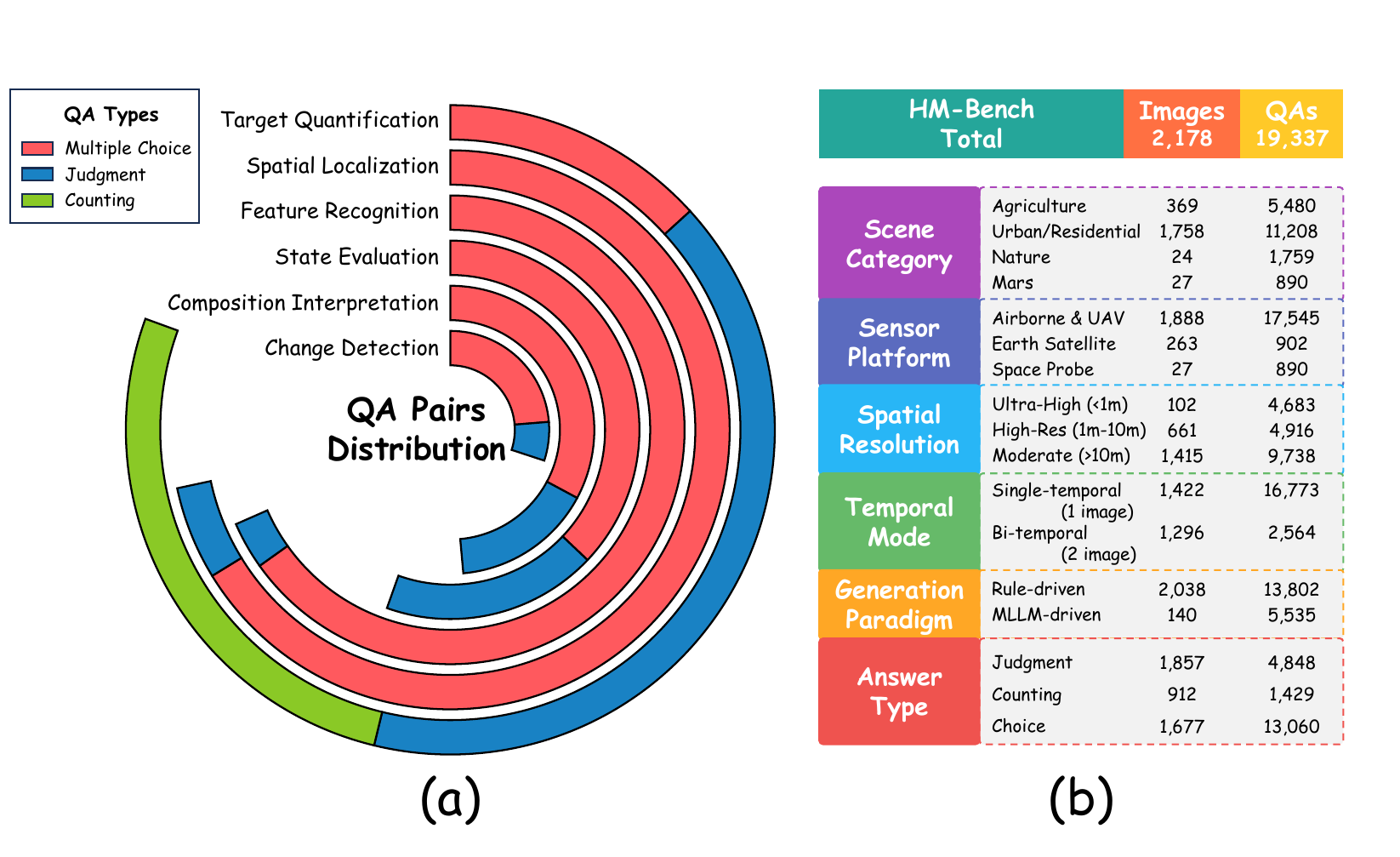}
    \caption{Statistical overview of HM-Bench. (a) Question-type distribution across six task scenarios. (b) Image and QA-pair counts by scene category, sensor platform, spatial resolution, temporal mode, generation paradigm, and answer type. (UAV: Unmanned Aerial Vehicle.)}
  \label{fig:dataset_statistics}
\end{figure}

\section{HM-Bench and the VSR$^{2}$ Framework}

\subsection{Overview}

We introduce \textbf{HM-Bench}, a benchmark for studying whether and how beyond-RGB hyperspectral information benefits MLLMs. Comprising 19,337 QA pairs from 2,178 hyperspectral samples across 13 task categories (Fig.~\ref{fig:dataset_statistics}), HM-Bench covers capabilities from general perception to hyperspectral reasoning. Multiple samples and questions can originate from the same scene; source-level aggregation is reported in Appendix~\ref{app:source_stratification}. Its construction follows an evidence-grounded hybrid workflow, as illustrated in Fig.~\ref{fig:frame+pipline}(a).

To make HSI data compatible with off-the-shelf MLLMs, we propose \textbf{VSR\(^{2}\)}, a training-free processing protocol. Each hyperspectral sample is converted into three aligned views: RGB, PCA, and a structured report (Fig.~\ref{fig:frame+pipline}(b)). Together with HM-Bench, it provides a controlled testbed for evaluating single- and multi-view hyperspectral reasoning.


\subsection{Data Sources and Preprocessing}

\begin{figure*}[t]
    \centering
    \includegraphics[width=\textwidth]{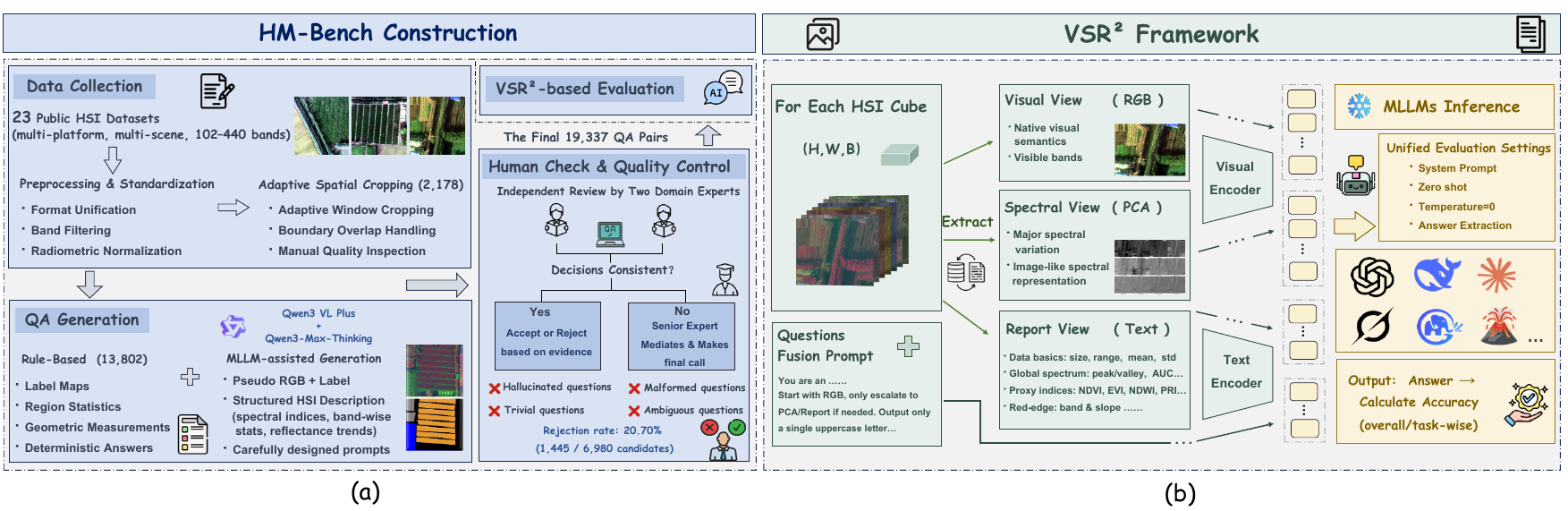}
    \caption{
    \textbf{HM-Bench construction and VSR$^{2}$ framework overview.}
    \textbf{(a)} HM-Bench construction pipeline. Built from 23 public hyperspectral datasets through preprocessing, adaptive cropping, QA generation, and expert verification, yielding 19,337 evidence-grounded QA pairs.
    \textbf{(b)} Overview of VSR$^{2}$. Each hyperspectral cube is converted into three aligned views, enabling controlled zero-shot evaluation of single-view and multi-view input settings.
    }
    \label{fig:frame+pipline}
\end{figure*}

As shown in Fig.~\ref{fig:frame+pipline} (a), HM-Bench is built from 23 public hyperspectral datasets, including \emph{Indian Pines}~\citep{baumgardner2015220}, \emph{Salinas}~\citep{hou2021hyperspectral}, \emph{Xiongan}~\citep{yi2020aerial}, \emph{Houston}~\citep{debes2014hyperspectral,xu2019advanced}, \emph{Washington DC Mall}~\citep{multispec_dc}, \emph{Hermiston}~\citep{hou2021hermiston}, \emph{Bay Area}~\citep{cooper2020bayarea}, \emph{Santa Barbara}~\citep{lopez2019gpu}, \emph{Pavia University}~\citep{hou2021hyperspectral}, \emph{Pavia Centre}~\citep{hou2021hyperspectral}, \emph{Botswana}~\citep{hou2021hyperspectral}, \emph{Urban}~\citep{zhu2017hyperspectral}, \emph{Kennedy Space Center}~\citep{kennedy2024ksc}, the \emph{WHU-Hi} series~\citep{zhong2020whu}, and the Martian \emph{MARS} suite~\citep{li2022stepwise}. Their published specifications span spaceborne, airborne, UAV, and extra-planetary platforms, with 48--440 bands, wavelength coverage from $0.38\,\mu m$ to $3.8\,\mu m$, spatial resolution from 0.043\,m to 30\,m, and scenes ranging from agriculture and urban areas to natural landscapes and Martian geomorphology (see Appendix~\ref{app:dataset} for details).

To improve cross-dataset consistency, we convert all scenes to a unified $(H, W, B)$ format, remove invalid bands affected by noise or atmospheric absorption, and apply radiometric normalization. We then align label maps with HSI scenes and correct clearly anomalous regions. For unlabeled or boundary-ambiguous areas, we derive auxiliary annotations using spectral clustering, physically grounded indices (e.g., NDVI and MNDWI), SAM-based spectral matching, and neighborhood filtering. Finally, we use adaptive cropping to convert large scenes into samples, with overlapping crops near boundaries to preserve informative content, followed by manual removal of low-quality patches dominated by background, severe noise, or spectral distortion.

\subsection{Task Taxonomy and Benchmark Composition}

\begin{figure*}[hbpt]
  \centering
  \hfill
  \includegraphics[width=\textwidth]{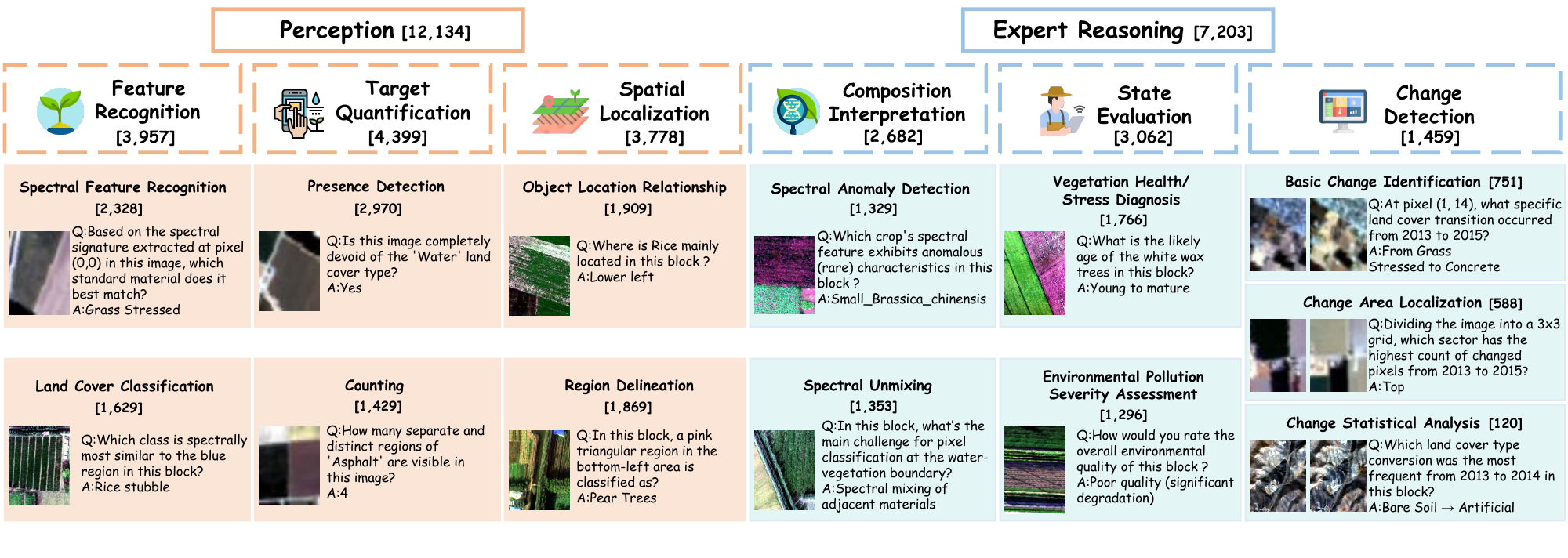}
  \caption{Hierarchical task taxonomy of HM-Bench, showing 13 hyperspectral tasks under perception and reasoning, together with data statistics and representative VQA examples.}
  \label{fig:task_samples}
\end{figure*}

HM-Bench follows a hierarchical design from basic perception to advanced HSI reasoning. As illustrated in Fig.~\ref{fig:task_samples}, it is organized into two high-level domains, \textbf{Perception} and \textbf{Reasoning}. 
Within Perception, we define three capability dimensions. \emph{Feature recognition} includes Spectral Feature Recognition (SFR) and Land Cover Classification (LCC); \emph{target quantification} includes Presence Detection (PD) and Counting (CS); and \emph{spatial localization} includes Object Location Relationship (OLR) and Region Delineation (RD).
Within Reasoning, we define three capability dimensions as well. \emph{Composition interpretation} includes Spectral Anomaly Detection (SAD) and Spectral Unmixing (SU); \emph{state evaluation} includes Vegetation Health/Stress Diagnosis (VH) and Environmental Pollution Severity Assessment (EPSA); and \emph{change detection} includes Basic Change Identification (BCI), Change Area Localization (CAL), and Change Statistical Analysis (CSA).

These 13 task categories span judgment, multiple-choice, and counting formats. Lower-level tasks focus on object existence, category recognition, counting, and localization, while higher-level tasks require composition interpretation, state evaluation, and cross-temporal reasoning from spectral or spatial-spectral evidence. Many reasoning tasks are constructed to require spectral or spatial-spectral evidence beyond standard RGB appearance. In total, HM-Bench contains 2,178 samples and 19,337 QA pairs, spanning sensing scales from centimeter-level UAV imagery to orbital and extra-planetary observations. Detailed distributions are shown in Fig.~\ref{fig:dataset_statistics}; further details are provided in Appendix \ref{app:benchmark_details}.

\subsection{Benchmark Construction}

HM-Bench is built through three stages: hyperspectral preprocessing, QA generation, and expert-centered quality control (Fig.~\ref{fig:frame+pipline} (a)). We distinguish whether a reference answer is valid in the original HSI and annotations from whether the RGB, PCA, and report views actually preserve enough evidence to infer that answer; the task-level evidence checks are specified in Appendix~\ref{app:evidence_sufficiency}.
\paragraph{Generation principle.}
All generation and auditing are conducted by domain experts. We distinguish two QA types: (1) \emph{objective} QA pairs, whose answers are deterministically derived from labels, geometry, or spectral statistics; and (2) \emph{reasoning-oriented} QA pairs, which require richer language but must be supported by explicit spectral-spatial evidence.
\paragraph{Rule-based generation for objective questions.}
For tasks with objectively verifiable answers, such as land-cover existence, category proportion, target counting, and localization-related queries, we adopt a rule-based generation pipeline. All questions and answers are derived from existing source annotations or their deterministic derivatives. QAs are programmatically derived from calibrated label maps, region statistics, and geometric measurements. Because these are grounded in deterministic metadata rather than subjective interpretation, we use a \emph{generation + expert self-inspection} protocol instead of redundant dual-pass agreement scoring.

\paragraph{MLLM-assisted generation for reasoning questions.}
For higher-level tasks, such as composition interpretation, state evaluation, and other spatial-spectral reasoning problems, we use MLLMs only as controlled language generators rather than annotation authorities. For each sample, we extract structured hyperspectral descriptors, including spectral indices, band-wise statistics, reflectance trends, and scene-level composition cues, and inject them into designed prompts to generate candidate questions and reference answers. These outputs improve linguistic diversity and reasoning depth, but are never included without expert verification.
The task instructions and evidence fields vary slightly by QA type. The generator returns only question, options, and answer; scripts add paths and taxonomy metadata after generation, as illustrated in Appendix~\ref{app:benchmark_details}.

\paragraph{Anti-shortcut design.}
A core principle of HM-Bench is that many reasoning tasks should not be solvable from RGB appearance alone. We therefore explicitly construct questions that depend on non-visible hyperspectral evidence, such as red-edge shifts, absorption features, material-specific reflectance responses, and cross-band composition patterns.

\paragraph{Expert verification and rejection analysis.}
To ensure QA quality, mitigate hallucinations and avoid circular dependency, we subject all MLLM-generated QA candidates to blind expert auditing by reviewers who are blinded to their generation source before inclusion. Following the expert verification protocol illustrated in Fig.~\ref{fig:frame+pipline} (a), each is independently reviewed by two domain experts against the underlying HSI evidence. If the two reviewers reach consistent judgments, the decision is finalized; otherwise, a third expert is introduced for adjudication. Expert reviewers verify each candidate based on spectral-spatial evidence and reject any sample that is unsupported, ambiguous, trivial, or improperly formatted.

In total, we audited {6,980} MLLM-generated candidates and rejected {1,445} (20.70\%). Rejections fall into four categories: hallucination (34.7\%), triviality (33.1\%), format issues (28.7\%), and others (3.5\%). After filtering, {5,535} expert-verified MLLM-assisted QA pairs are retained. A final QA pair is included only if its answer can be traced to calibrated labels, geometric measurements, spectral indices, band-level statistics, or expert-confirmed spectral interpretation.

\paragraph{Answer format control.}
For multiple-choice questions, prompts explicitly require the model to output {only} the option letter, minimizing post-processing ambiguity. Manual inspection shows near-perfect compliance in zero-shot evaluation, indicating that benchmark results are not materially affected by answer-extraction errors.

Overall, this multi-stage curation strategy combines the scalability of automatic generation, the linguistic flexibility of MLLM-assisted formulation, and the reliability of strict expert verification, resulting in a benchmark that is both diverse and evidence-grounded.

\subsection{VSR$^{2}$: Visual--Spectral--Report Reasoning}

To expose hyperspectral information to off-the-shelf MLLMs without retraining, we propose \textbf{VSR$^{2}$}, a training-free framework that converts each HSI sample into three aligned views (Fig.~\ref{fig:frame+pipline} (b)): an RGB image, a PCA-based spectral view, and a structured textual report. These views are complementary: the RGB view preserves natural visual semantics and spatial layout, the PCA view provides an image-compatible summary of dominant spectral variation, and the report supplies explicit quantitative spectral-spatial evidence through the language interface. More details are in Appendix \ref{app:vsr}.
\paragraph{RGB visual view.}
We derive an RGB image from visible-spectrum bands using sensor-aware band selection rather than a fixed pseudo-color mapping. For each dataset, we choose the three bands whose central wavelengths are closest to standard red, green, and blue ranges, using the nearest visible bands when exact matches are unavailable. The selected bands are normalized and composed into a standard three-channel image. This yields a physically grounded RGB baseline that preserves scene appearance and spatial layout while avoiding arbitrary false-color rendering.

\begin{figure*}[t]
    \centering
    \includegraphics[width=\textwidth]{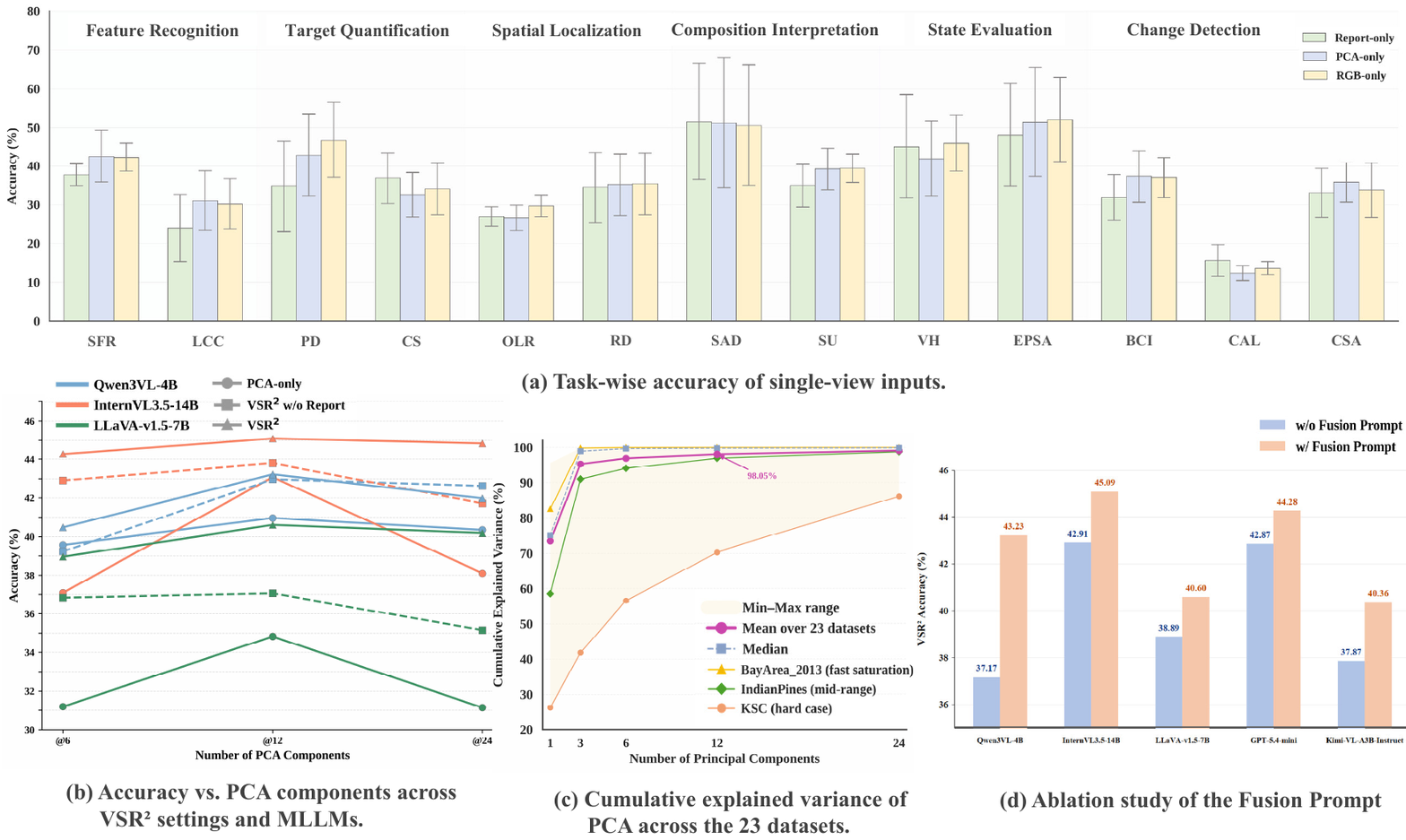}
    \caption{
    Main results, task-wise analysis, and ablations of VSR$^{2}$ on HM-Bench. The subfigures present task-wise single-view performance, model-level sensitivity to the number of PCA components, dataset-level PCA explained variance, and the effect of the fusion prompt. In (a), bars are means over 17 models and error bars show one standard deviation; CS, VH, and EPSA correspond to Count, VHS, and EPS in Table~\ref{tab:full_results}.
    }
    \label{fig:result}
\end{figure*}

\paragraph{PCA spectral view.}
We summarize spectral variation using PCA. Given an HSI cube of size $H \times W \times B$, we reshape it so that each pixel is represented by a $B$-dimensional spectrum, perform PCA along the spectral dimension, and render the resulting component maps as normalized grayscale images. For HM-Bench, we retain the top 12 principal components and arrange them into a standardized layout of three rows and four columns. This choice balances spectral coverage and visual compactness: across the 23 source datasets, the first 12 components explain over 95\% cumulative variance in 22 datasets (Fig.~\ref{fig:result}(c)), and we further validate this design against 6-, 12-, and 24-component variants (Fig.~\ref{fig:result}(b)).

\begin{table*}[htbp]
\centering
\small
\setlength{\tabcolsep}{4pt}
\definecolor{groupgray}{RGB}{240,240,240}

\caption{\textbf{Overall performance under different VSR$^{2}$ input settings on HM-Bench.} Question-weighted accuracy (\%) under a unified zero-shot scoring protocol. The Average row is the unweighted mean of the 17 displayed model scores, excluding Random; means are recomputed from the values displayed to two decimals. \textbf{Bold} indicates the best result per column (input setting); \underline{underlined} indicates the second-best per column. The Random row is excluded from ranking.}
\label{tab:overall_results}

\begin{tabular}{l *{6}{wc{1.3cm}}}
\toprule
\textbf{Model} & \textbf{Report} & \textbf{PCA} & \textbf{RGB} & \textbf{w/o PCA} & \textbf{w/o Report} & \textbf{VSR$^{2}$} \\
\midrule
Random & 25.13 & 25.13 & 25.13 & 25.13 & 25.13 & 25.13 \\

\multicolumn{7}{@{}l}{\cellcolor{groupgray}\small\textit{Closed-source Models}} \\[2pt]
Claude-Sonnet-4-6~\cite{anthropic2026claudesonnet46}    & 36.57 & 39.48 & 40.95 & 39.65 & 43.11 & 44.01 \\
GPT-5.4-Mini~\cite{openai2026gpt54mini}                  & 37.45 & \underline{42.35} & 39.88 & 40.58 & 43.70 & \underline{44.28} \\
Grok-4~\cite{de2025grok}                            & 31.68 & 32.78 & 34.67 & 38.70 & \underline{43.74} & 39.50 \\

\multicolumn{7}{@{}l}{\cellcolor{groupgray}\small\textit{Open-source Models}} \\[2pt]
InternVL2-8B~\cite{chen2024internvl}         & 36.17 & 38.09 & 36.63 & 37.90 & 38.35 & 39.12 \\
InternVL2.5-4B~\cite{chen2024internvl}       & 32.30 & 34.83 & 36.74 & 34.62 & 34.01 & 35.92 \\
InternVL3-9B~\cite{chen2024internvl}         & 37.18 & 42.14 & 41.74 & 40.00 & 42.86 & 40.47 \\
InternVL3-14B~\cite{chen2024internvl}        & 37.26 & 38.09 & 42.48 & 40.11 & 42.60 & 40.05 \\
InternVL3.5-14B~\cite{chen2024internvl}     & \textbf{39.52} & \textbf{43.08} & \underline{42.88} & \textbf{42.98} & \textbf{43.80}& \textbf{45.09} \\
Kimi-VL-A3B-Instruct~\cite{team2025kimi}    & 37.57 & 38.93 & 40.24 & 39.13 & 40.58 & 40.36 \\
Qwen2.5-VL-7B~\cite{bai2023qwen}            & 31.03 & 29.36 & 38.19 & 37.88 & 39.30 & 39.41 \\
Qwen3-VL-4B~\cite{bai2023qwen}              & 35.93 & 40.96 & 40.08 & 39.34 & 42.97 & 43.23 \\
LLaVA-Next-8B~\cite{li2024llava}            & 36.71 & 39.03 & 37.39 & 38.96 & 39.95 & 39.28 \\
LLaVA-v1.5-7B~\cite{liu2024improved}        & 32.30 & 34.83 & 31.05 & 35.10 & 37.06 & 40.60 \\
Geochat-7B~\cite{kuckreja2024geochat}       & 34.06 & 35.27 & 37.03 & 36.01 & 40.78 & 39.60 \\
GeoLLaVA-8K~\cite{wang2025geollava}         & \underline{38.76} & 37.79 & 39.27 & 39.57 & 40.80 & 41.94 \\
GLM-4.6V-Flash-9B~\cite{hong2025glm}        & 37.72 & 42.06 & \textbf{44.46} & \underline{41.96} & 43.26 & 43.61 \\
Nemotron-Nano-VL-8B~\cite{deshmukh2025nvidia} & 37.00 & 38.67 & 37.54 & 39.47 & 36.56 & 39.61 \\
\midrule

Average & 35.84 & 38.10 & 38.90 & 38.94 & 40.79 & 40.95 \\
\bottomrule
\end{tabular}
\end{table*}

\paragraph{Report view.}
The third view is a structured textual report built from quantitative spectral-spatial descriptors, including basic data properties, global spectral statistics, proxy indices, local grid-level variation, reflectance trends, and PCA summaries. The generation prompt receives these numerical descriptors, without ground-truth labels, class proportions, questions, or answers, and instructs the generator to avoid semantic class claims. Because generated language can still introduce interpretive cues or errors, we check the resulting reports and test the effect of removing descriptive wording in Appendix~\ref{app:report_generation} and Appendix~\ref{app:report_ablation}.

\paragraph{Fusion prompt.}
VSR$^2$ uses a common lightweight prompt template with clauses adapted to the available views. In the full configuration, it asks the model to first inspect RGB for scene semantics, then consult PCA and the report when spectral or quantitative evidence is needed. It contains no task-specific demonstrations, labels, or answer hints.

\subsection{Evaluation Protocol}

We conduct controlled zero-shot evaluation on HM-Bench using 17 MLLMs, including 3 closed-source and 14 open-source models. We evaluate six input settings covering single-view, partial multi-view, and full VSR$^{2}$ exposure. For each model, we report both overall accuracy and task-wise performance across six capability dimensions and 13 task categories. All settings use the same questions, answer scoring, and base instruction template, with modality-specific clauses adapted to the available views. Thus, RGB-only versus full VSR$^2$ is a system-level comparison that also changes input quantity and prompt content; more tightly matched spectral-range and empty-report controls are described in Appendix~\ref{app:controlled_contributions}. All experiments are strictly training-free, without task-specific fine-tuning, adapter tuning, or additional HSI pretraining. Implementation details, including prompt templates and scoring criteria, are provided in Appendix~\ref{app:evaluation}.


\section{Experiments}

Table~\ref{tab:overall_results} reports overall accuracy across VSR$^{2}$ variants, while Fig.~\ref{fig:result} breaks down task-wise single-view performance and other ablations. All models use the same questions, answer extraction, and scoring rules. Because the six input settings differ in both supplied evidence and applicable prompt clauses, we interpret their overall differences as system-level effects and use matched controls to assess narrower spectral and report contributions (Appendix~\ref{app:controlled_contributions}).

\subsection{Results}

Overall, beyond-RGB hyperspectral information improves MLLM performance on HM-Bench, but the gains are modest and highly non-uniform. The effect depends both on the representation form and on the model’s ability to integrate multiple views. The results indicate that current MLLMs can benefit from hyperspectral cues, while robust cross-view reasoning remains an open challenge.

\paragraph{Beyond-RGB helps, but only modestly.}
Averaged over all models, full VSR$^{2}$ achieves the best mean accuracy at {40.95\%}, outperforming RGB-only (38.90\%), PCA-only (38.10\%), Report-only (35.84\%), VSR$^{2}$ w/o PCA (38.94\%), and VSR$^{2}$ w/o Report (40.79\%). Thus, current MLLMs can benefit from beyond-RGB HSI evidence, but the average gain over RGB-only is +2.05 percentage points, indicating limited zero-shot exploitation ability.

\paragraph{Full VSR$^{2}$ is best overall, but not universally.}
Full VSR$^{2}$ yields the best results for Claude-Sonnet-4-6, GPT-5.4-Mini, InternVL2-8B, InternVL3.5-14B, Qwen2.5-VL-7B, Qwen3-VL-4B, LLaVA-v1.5-7B, GeoLLaVA-8K, and Nemotron-Nano-VL-8B, showing that RGB, PCA, and report views are complementary overall. However, no single setting dominates universally: Grok-4, InternVL3-9B, InternVL3-14B, Kimi-VL-A3B-Instruct, LLaVA-Next-8B, and GeoChat-7B perform better under VSR$^{2}$ w/o Report than under full VSR$^{2}$. This suggests that adding more hyperspectral evidence helps only when the model can effectively coordinate heterogeneous views.

\paragraph{Visual spectral cues outperform text-only reports.}
Among single-view settings, RGB-only performs best on average ({38.90\%}), followed closely by PCA-only ({38.10\%}), while Report-only is weaker ({35.84\%}). Moreover, VSR$^{2}$ w/o Report already reaches {40.79\%}, nearly matching full VSR$^{2}$ (40.95\%). The result indicates that much of the gain comes from presenting spectral information in an image-compatible form, whereas the report view is less consistently useful and depends more on cross-view reasoning ability.

\paragraph{Report view benefits partial MLLMs.}
Full VSR$^{2}$ outperforms VSR$^{2}$ w/o Report for 11 of the 17 models, including Claude-Sonnet-4-6, GPT-5.4-Mini, InternVL2-8B, InternVL2.5-4B, InternVL3.5-14B, Qwen2.5-VL-7B, Qwen3-VL-4B, LLaVA-v1.5-7B, GeoLLaVA-8K, GLM-4.6V-Flash, and Nemotron-Nano-VL-8B, indicating that some MLLMs can further exploit explicit spectral statistics when combined with visual evidence. The non-uniform effect across models highlights that joint reasoning over RGB, PCA, and textual spectral summaries remains challenging.

\paragraph{Matched spectral and report controls.}
On three representative models, replacing visible-band PCA with full-spectrum PCA while keeping RGB and the two-image format improves overall accuracy by 0.86, 0.47, and 0.51 percentage points. Keeping both images and the report field fixed but replacing an empty report with its quantitative content adds 2.93, 1.41, and 0.15 points. These effects vary across tasks; in particular, report content reduces presence-detection accuracy on all three models. The complete task-wise contrasts and the scope of each comparison are given in Appendix~\ref{app:controlled_contributions}.

\paragraph{Task-dependent beyond-RGB performance.}
Fig.~\ref{fig:result}(a) compares Report-only, PCA-only, and RGB-only across the 13 tasks in six capability dimensions. The three representation views exhibit different strengths.
First, RGB-only remains strongest or most stable on many visually grounded tasks, especially PD, OLR, VH, and EPSA. 
Second, PCA-only is highly competitive and sometimes superior on tasks more directly tied to spectral variation, including SFR, LCC, BCI, and CSA. 
Third, Report-only is rarely the best overall, but is comparatively competitive on more quantitative or status-oriented tasks, such as CS and VH. 

\paragraph{Sensitivity to data source and report wording.}
To account for multiple questions derived from the same scene, we also aggregate the archived full VSR$^2$ predictions for 13 models by original scene group. The question-pooled accuracy is 39.71\%, while the equal-scene average is 36.66\% (95\% scene-level interval [30.8, 42.5]\%). A three-model control that removes generated interpretive wording from reports changes full VSR$^2$ accuracy by only $-0.26$ percentage points on average, although individual tasks vary. The report checks, complete ablation, and source-stratified results are detailed in Appendices~\ref{app:report_generation} and~\ref{app:more_results}.


\subsection{Ablation Studies}

To further analyze the design choices of VSR$^{2}$, we perform ablation studies on the PCA view and fusion prompt.

\paragraph{Effect of the number of PCA components.}
We vary the number of retained principal components and evaluate three representative open-source models under PCA-only, VSR$^{2}$ w/o Report, and full VSR$^{2}$. As shown in Fig.~\ref{fig:result}(b), retaining {12} components consistently gives the best performance across all three models and settings. For Qwen3-VL-4B, accuracy peaks at 40.96/42.97/43.23; for InternVL3.5-14B, 43.08/43.80/45.09; and for LLaVA-v1.5-7B, 34.83/37.06/40.60, under PCA-only, VSR$^{2}$ w/o Report, and full VSR$^{2}$, respectively. Compared with 6 components, 12 preserves richer spectral variation; compared with 24, it avoids redundant and visually cluttered inputs.

This choice is also supported by dataset-level PCA statistics in Fig.~\ref{fig:result}(c) and Table~\ref{tab:pca_variance_summary}. Across the 23 datasets, cumulative explained variance averaged over datasets reaches 95.25\% with 3 components, 96.91\% with 6, and 98.05\% with 12. With 12 components, 22/23 datasets exceed 95\% explained variance and 19/23 exceed 98\%, with KSC as the main outlier. Therefore, 12 components balance spectral fidelity and visual compactness, consistent with the model-level results.

\paragraph{Effect of the fusion prompt.}
We further test whether VSR$^{2}$ benefits only from richer inputs or also from explicit guidance on how to use them. The fusion prompt instructs the model to start from the RGB view and consult PCA and report views only when additional spectral evidence is needed, encouraging progressive cross-view reasoning. As shown in Fig.~\ref{fig:result}(d), the prompt consistently improves all five tested models: Qwen3-VL-4B (+6.06), InternVL3.5-14B (+2.18), LLaVA-v1.5-7B (+1.71), GPT-5.4-Mini (+1.41), and Kimi-VL-A3B-Instruct (+2.49), with a mean gain of 2.77 percentage points across these five models. This comparison suggests that the tested models benefit from explicit fusion guidance in addition to the auxiliary views.


\section{Conclusion}
To address HSI understanding for MLLMs, we introduce \textbf{HM-Bench} and \textbf{VSR$^{2}$}. HM-Bench provides 13 evidence-grounded tasks across six capability dimensions, while VSR$^{2}$ exposes HSI information to current MLLMs through RGB, PCA-based spectral, and structured report views. Under a controlled zero-shot protocol, experiments on 17 representative MLLMs show that beyond-RGB evidence improves average accuracy over RGB-only inputs, with model-compatible visual representations contributing most consistently.
The results further show that HSI reasoning is not a uniform capability. The gains remain modest, task-dependent, and architecture-dependent; text-based spectral reports are less consistently effective than visual spectral representations, while reliable three-view fusion remains challenging for current models. These findings establish HM-Bench as a testbed for advancing multimodal reasoning beyond RGB. Future work will focus on developing HSI-native representations and more effective cross-view fusion strategies, as well as adapting MLLMs to spectral reasoning through domain-specific training and evaluation.

\bibliography{aaai2027}


\appendix

\clearpage

\twocolumn[

\begin{center}
{\LARGE\bfseries
Beyond RGB: Benchmarking and Enhancing MLLMs for Hyperspectral Image Understanding via Training-Free Reasoning Framework
}

\vspace{0.3em}

{\large\bfseries
(Supplementary Material)
}
\end{center}

\vspace{1em}
]

\newcounter{appsection}
\renewcommand{\theappsection}{\Alph{appsection}}

\makeatletter
\@addtoreset{subsection}{appsection}
\makeatother

\renewcommand{\thesubsection}{\theappsection.\arabic{subsection}}

\newcommand{\appsection}[1]{%
    \refstepcounter{appsection}%
    \section*{\theappsection.\quad #1}%
}

\section*{Appendix}
This appendix provides additional details for HM-Bench and VSR$^2$ that are omitted from the main paper due to space constraints. The appendix is organized as follows:

\begin{itemize}[leftmargin=7.5pt]
    \item \textbf{Sec.~\ref{app:dataset}: Dataset Description} -- Detailed information about the 23 hyperspectral datasets used in HM-Bench.
    \item \textbf{Sec.~\ref{app:benchmark_details}: Details of HM-Bench} -- Detailed benchmark construction, including task taxonomy and QA generation.
    \item \textbf{Sec.~\ref{app:vsr}: Details of VSR$^2$} -- Detailed description of the VSR$^2$ framework, including the construction of the RGB, PCA, structured report views and the fusion prompt.
    \item \textbf{Sec.~\ref{app:evaluation}: Details of the Evaluation Setup} -- Evaluation settings, answer extraction, and scoring protocol.
    \item \textbf{Sec.~\ref{app:more_results}: Supplementary Results and Analysis} -- Extended experimental results and additional analyses.
    \item \textbf{Sec.~\ref{app:limitations}: Limitations and Future Work} -- Discussion of limitations and future directions.
    \item \textbf{Sec.~\ref{app:casestudy}: Case Study} -- Representative examples and qualitative analysis.
\end{itemize}

\appsection{Dataset Description}
\label{app:dataset}

To support a comprehensive evaluation of MLLMs in hyperspectral image understanding, HM-Bench integrates 23 publicly available hyperspectral datasets spanning diverse sensing platforms, scene types, and spectral configurations. These datasets cover a wide range of spatial resolutions, spectral band numbers, and semantic environments, providing broad coverage from centimeter-level UAV imagery to airborne, spaceborne, and planetary observations. Table~\ref{tab:dataset_stats} summarizes the detailed statistics of the included datasets.

\subsection{Multi-Platform and Multi-Scale Heterogeneity}
HM-Bench achieves an unprecedented platform coverage spanning "UAV-Airborne-Spaceborne-Deep Space". It includes UAV-borne sensors (e.g., WHU-Hi series equipped with Headwall Nano-Hyperspec~\citep{zhong2020whu}) for ultra-high-resolution precision agriculture, airborne sensors (e.g., AVIRIS~\citep{baumgardner2015220}, CASI~\citep{debes2014hyperspectral, xu2019advanced}, ROSIS~\citep{hou2021hyperspectral}) for classic large-scale mapping with high signal-to-noise ratios, and spaceborne or deep-space sensors (e.g., CRISM~\citep{li2022stepwise}, NASA EO-1~\citep{hou2021hyperspectral}) for macroscopic and planetary exploration. The diversity of sensing platforms naturally results in a massive cross-scale spatial resolution spanning from the centimeter level (0.043m, WHU-Hi-HongHu) to the decameter level (30m, Botswana~\citep{hou2021hyperspectral}). This cross-scale variation fundamentally challenges MLLMs: while ultra-high resolution tests fine-grained texture and local geometry perception~\citep{rasti2020feature}, low resolution forces MLLMs to abandon RGB-like shape reliance and deeply interpret "spectral fingerprints" for mixed pixel reasoning~\citep{zhu2017hyperspectral}.

\begin{figure}[htbp]
    \centering
    \includegraphics[width=0.7\columnwidth]{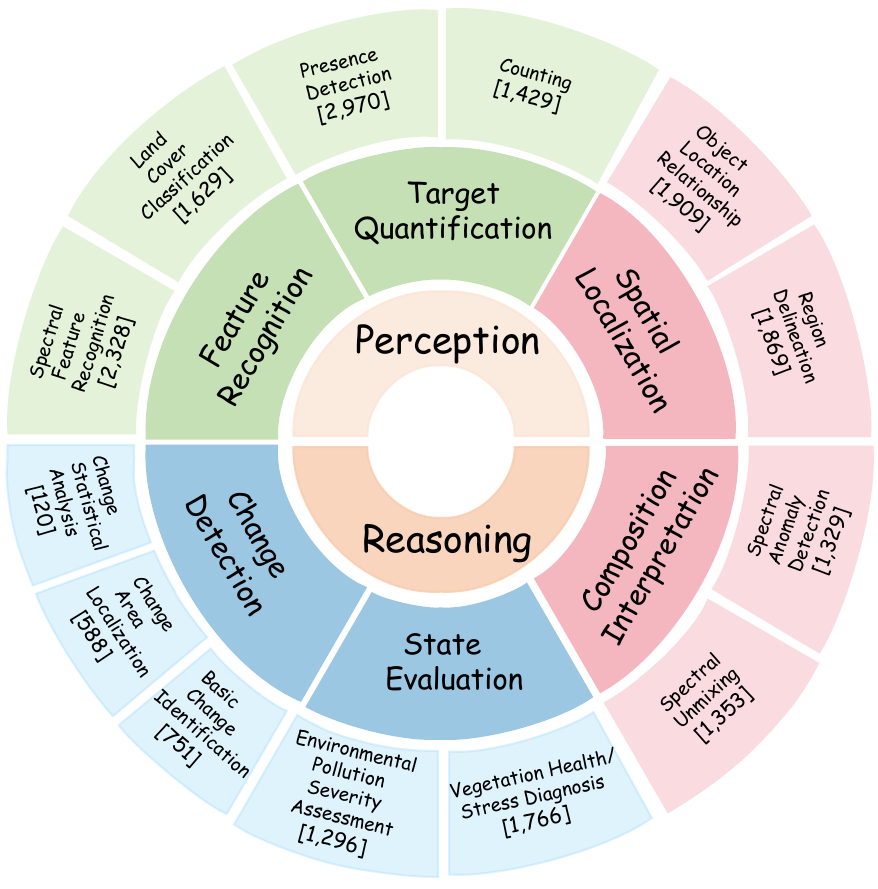}
    \vspace{-1em}
    \caption{Hierarchical task taxonomy of HM-Bench.}
    \label{fig:piechart}
\end{figure}

\begin{table*}[t]
  \caption{Detailed statistics of the HSI datasets included in HM-Bench.}
  \label{tab:dataset_stats}
  \centering
  \resizebox{\textwidth}{!}{
      \begin{tabular}{lccccccc}
        \toprule
        \textbf{Dataset} & \textbf{Sensor} & \textbf{Resolution} & \textbf{Bands} & \textbf{Wavelength} & \textbf{QAs} & \textbf{Patch Size} & \textbf{Scene Type} \\
        \midrule
        IndianPines~\citep{baumgardner2015220} & AVIRIS & 20m & 200 & 0.4 -- 2.5 $\mu$m & 450 & 40 $\times$ 40 & Agriculture/Nature \\
        Salinas~\citep{hou2021hyperspectral} & AVIRIS & 3.7m & 204 & 0.4 -- 2.5 $\mu$m & 1,080 & 64 $\times$ 64 & Agriculture \\
        Xiongan~\citep{yi2020aerial} & \makecell{VNIR Imaging Spectrometer (SITP)} & 0.5m & 250 & 0.4 -- 1 $\mu$m & 3,003 & 256 $\times$ 256 & Agriculture \\
        Houston 2013~\citep{debes2014hyperspectral} & CASI-1500 & 2.5m & 144 & 0.38 -- 1.05 $\mu$m & 501 & 100 $\times$ 100 & Urban/Nature \\
        Houston 2018~\citep{xu2019advanced} & CASI-1500 & 1m & 48 & 0.38 -- 1.05 $\mu$m & 501 & 100 $\times$ 100 & Urban/Nature \\
        WashingtonDC~\citep{multispec_dc} & HYDICE & --- & 191 & 0.4 -- 2.4 $\mu$m & 612 & 48 $\times$ 48 & Urban/Nature \\
        Hermiston (2004\&2007)~\citep{hou2021hermiston} & EO-1 & 30m & 242 & 0.4 -- 2.5 $\mu$m & 384 & 32 $\times$ 32 & Agriculture \\
        BayArea (2013\&2015)~\citep{cooper2020bayarea} & AVIRIS & 16.9m & 224 & 0.4 -- 2.5 $\mu$m & 6,000 & 32 $\times$ 32 & Urban/Nature \\
        SantaBarbara (2013\&2014)~\citep{lopez2019gpu} & AVIRIS & 15 -- 18m & 224 & 0.4 -- 2.6 $\mu$m & 820 & 100 $\times$ 100 & Urban/Nature \\
        Pavia Center~\citep{hou2021hyperspectral} & ROSIS & 1.3m & 102 & 0.43 -- 0.86 $\mu$m & 768 & 121 $\times$ 715 & Urban/Residential \\
        Pavia University~\citep{hou2021hyperspectral} & ROSIS & 1.3m & 103 & 0.43 -- 0.86 $\mu$m & 707 & 67 $\times$ 340 & Urban/Campus \\
        Kennedy Space Center (KSC)~\citep{kennedy2024ksc} & NASA AVIRIS & 18m & 176 & 0.4 -- 2.5 $\mu$m & 676 & 56 $\times$ 614 & Space Launch Site \\
        Botswana~\citep{hou2021hyperspectral} & NASA EO-1 & 30m & 145 & 0.4 -- 2.5 $\mu$m & 518 & 164 $\times$ 256 & Desert \\
        WHU-Hi-HanChuan~\citep{zhong2020whu} & Headwall Nano-Hyperspec & 0.109m & 274 & 0.4 -- 1 $\mu$m & 563 & 274 $\times$ 405 & Agricultural \\
        WHU-Hi-HongHu~\citep{zhong2020whu} & Headwall Nano-Hyperspec & 0.043m & 270 & 0.4 -- 1 $\mu$m & 565 & 270 $\times$ 313 & Lake \\
        WHU-Hi-LongKou~\citep{zhong2020whu} & Headwall Nano-Hyperspec & 0.463m & 270 & 0.4 -- 1 $\mu$m & 552 & 270 $\times$ 183 & Coastal Zone/Urban \\
        Urban~\citep{zhu2017hyperspectral} & HYDICE & 2m & 162 & 0.4 -- 2.5 $\mu$m & 747 & 34 $\times$ 207 & Urban/Residential \\
        MARS-Holden~\citep{li2022stepwise} & CRISM & 18m & 440 & 0.4 -- 3.8 $\mu$m & 290 & 46 $\times$ 595 & Mars Terrain \\
        MARS-Utopia~\citep{li2022stepwise} & CRISM & 18m & 432 & 0.4 -- 3.8 $\mu$m & 150 & 53 $\times$ 595 & Mars Terrain \\
        MARS-Nilifossae~\citep{li2022stepwise} & CRISM & 18m & 425 & 0.4 -- 3.8 $\mu$m & 450 & 53 $\times$ 595 & Mars Terrain \\
        \bottomrule
      \end{tabular}
    }
\end{table*}

\subsection{Multi-Scene and Multi-Domain Diversity}
The integrated datasets encompass diverse environments spanning macroscopic to microscopic scales, ensuring that the benchmark is not biased toward specific land covers. The scene taxonomy covers precision agriculture (e.g., Indian Pines~\citep{baumgardner2015220}, Salinas~\citep{hou2021hyperspectral}) involving complex crop classification and health monitoring, complex urban and residential landscapes (e.g., Houston~\citep{debes2014hyperspectral, xu2019advanced}, Washington DC~\citep{multispec_dc}) with dense buildings and structural occlusions, as well as natural terrains and ecology (e.g., Botswana~\citep{hou2021hyperspectral}, KSC~\citep{kennedy2024ksc}). Uniquely, HM-Bench incorporates extraterrestrial geomorphology (MARS series~\citep{li2022stepwise}), evaluating the capability of MLLMs to process unknown, extreme hyperspectral signatures beyond Earth. This extensive diversity reduces the risk of scenario overfitting and supports a more comprehensive evaluation of hyperspectral reasoning capabilities.

\paragraph{Visual and task coverage.}
The 19,337 question--answer pairs are derived from 2,178 hyperspectral samples; multiple questions can refer to the same sample, and samples from a source scene can overlap. Thus the number of questions measures evaluation instances, not independent scenes. After merging temporal acquisitions of the same location, the 19 source groups span 176 source--task combinations, and 14 groups contain questions from at least 10 of the 13 task categories. These counts describe coverage of the existing benchmark, not independent observations or generalization to unseen sources. Source-stratified results are given in Appendix~\ref{app:source_stratification}.

\paragraph{Pretraining exposure and question-only controls.}
HM-Bench constructs new question--answer pairs from measurements, annotations, and structured evidence associated with public imagery. New questions reduce the possibility of memorization of identical benchmark pairs, but cannot establish that an evaluated model has not encountered a source image or related material during pretraining. VSR$^2$ performs no benchmark-specific parameter updates. The question-only evaluation in Table~\ref{tab:spectral_range_question_only} measures performance from the supplied question and options without image or report inputs; its near-chance results on the three tested models do not by themselves rule out prior exposure or establish that every question requires non-visible evidence.

\subsection{Multi-Band and Multi-Dimensional Coverage}
Unlike standard visual benchmarks, HM-Bench retains multiband hyperspectral cubes rather than reducing all source data to RGB. The published datasets cover wavelengths from the visible range (0.38 $\mu$m) to short-wave infrared (SWIR, 2.5 $\mu$m), extending to 3.8 $\mu$m for Martian datasets~\citep{li2022stepwise}, with official band counts from 48 to 440. This spectral coverage captures physically meaningful properties that are invisible to conventional RGB sensors, including moisture content and vegetation red-edge responses.

\appsection{Details of HM-Bench}
\label{app:benchmark_details}

HM-Bench is designed to evaluate whether and how MLLMs can benefit from hyperspectral information beyond RGB. To this end, the benchmark covers a broad spectrum of capabilities, ranging from basic perception to spatial--spectral reasoning. Its construction follows an evidence-grounded workflow that combines expert-designed task taxonomy, deterministic rule-based generation, and model-assisted generation for complex reasoning questions.

\subsection{Hierarchical Task Taxonomy}
To rigorously evaluate the diverse capabilities of MLLMs, HM-Bench establishes a hierarchical taxonomy (detailed in Fig.\ref{fig:piechart}), comprising 13 distinct tasks categorized into two primary dimensions: Perception and Reasoning.

\textbf{Feature Recognition.}
This capability dimension includes two tasks: Spectral Feature Recognition (SFR), which requires models to identify specific material types (e.g., distinguishing between healthy and stressed grass) based on spectral information; and Land Cover Classification (LCC), which tests semantic partitioning of image regions using both spatial texture and high-dimensional spectral cues.

\textbf{Target Quantification.}
This capability dimension includes Presence Detection (PD), a binary task that determines the existence of specific land-cover categories (e.g., verifying the absence of water pixels); and Counting (CS), which requires models to enumerate independent object instances or estimate the total spatial area of specific categories.

\textbf{Spatial Localization.}
This capability dimension includes Object Location Relationship (OLR), which requires models to reason about the relative spatial orientation of targets (e.g., identifying the quadrant of a rice field); and Region Delineation (RD), which requires models to output the precise minimal bounding box of a target for fine-grained localization.

\textbf{Composition Interpretation.}
This capability dimension includes Spectral Anomaly Detection (SAD), which challenges models to identify rare spectral signatures or abnormal physicochemical indicators in complex backgrounds; and Spectral Unmixing (SU), which requires understanding mixed-pixel phenomena to resolve sub-pixel endmember components together with their relative abundances or concentrations.

\textbf{State Evaluation.}
This capability dimension includes Vegetation Health/Stress Diagnosis (VH), which leverages non-visible spectral features such as red-edge shifts and near-infrared reflectance to assess crop growth stages and stress levels; and Environmental Pollution Severity Assessment (EPSA), which requires models to quantify ecological degradation based on the absorption characteristics of water or soil.
\begin{table*}[t]
\centering
\caption{Representative rejection cases during MLLM-based QA generation.}
\label{tab:rejection_examples}
\renewcommand{\arraystretch}{1.25}
\begin{tabular}{>{\centering\arraybackslash}m{1.9cm}
                >{\raggedright\arraybackslash}m{3.7cm}
                >{\raggedright\arraybackslash}m{5.0cm}
                >{\raggedright\arraybackslash}m{5.2cm}}
\toprule
\textbf{Category} & 
\textbf{Definition} & 
\textbf{Rejected QA candidate} & 
\textbf{Reason for rejection} \\
\midrule

\textbf{Hallucination} &
The MLLM generates surface-plausible content that cannot be supported by the available hyperspectral evidence. &
\textbf{Task:} Vegetation Health/Stress Diagnosis

\textbf{Question:} Based on the spectral characteristics of the vegetation in this block, what is the estimated chlorophyll content?

\textbf{Options:} A. Low (SPAD $<$ 20) \newline
B. Moderate (SPAD 30--40) \newline
C. High (SPAD $>$ 45) \newline
D. Extremely high (SPAD $>$ 60)

\textbf{Answer:} C &
The question requests a quantitative chlorophyll estimate, but the available spectral-spatial evidence does not provide sufficient calibration information for biochemical retrieval. The MLLM incorrectly associates vegetation appearance with a specific physiological parameter and produces a seemingly plausible but unsupported quantitative claim.
\\
\midrule

\textbf{Triviality} &
The QA pair is correctly formatted but provides limited evaluation value because the answer can be obtained from overly obvious visual cues without requiring meaningful spectral-spatial analysis. &
\textbf{Task:} Presence Detection

\textbf{Question:} Does vegetation exist in this image patch?

\textbf{Options:} A. Yes \quad B. No

\textbf{Answer:} A &
The image patch is almost entirely occupied by visually dominant vegetation, making the answer directly observable from RGB appearance or simple visual inspection. Although the QA pair is valid, it does not require hyperspectral evidence or non-visible spectral information, resulting in limited capability discrimination.
\\
\midrule

\textbf{Format Issue} &
The generated content may be semantically meaningful but violates the predefined QA schema or evaluation protocol. &
\textbf{Task:} Spectral Unmixing

\textbf{Question:} Based on the spectral characteristics of pixel, which endmember combination is most likely?

\textbf{Options:} A. Concrete and Asphalt \newline
...
D. Healthy vegetation and Concrete

\textbf{Answer:} The pixel contains mixed vegetation and impervious surface signals due to transitional spectral responses. &
The benchmark requires the answer to be represented only by the option index (e.g., ``B''). The generated response contains additional free-form explanation, which violates the predefined answer format and prevents automatic evaluation.
\\

\bottomrule
\end{tabular}
\end{table*}
\textbf{Change Detection (Bi-temporal).}
This capability dimension evaluates multi-temporal hyperspectral reasoning through three tasks: Basic Change Identification (BCI), which identifies land-cover transitions; Change Area Localization (CAL), which identifies the specific grid sector exhibiting the most intensive changes; and Change Statistical Analysis (CSA), which infers macroscopic change trends (e.g., urbanization of bare soil) or quantifies global change metrics.

\subsection{Rule-based QA Generation}
The Rule-based Generation route serves as a deterministic logic engine that converts high-precision physical statistics and spatial-geometric parameters into structured QA pairs based on human-designed protocols. This pipeline ensures the absolute veracity of the ground truth through the following mechanisms:

\textbf{Logic Triggering and Template Selection.} Statistical profiles derived during pre-processing serve as the primary drivers for template dispatching~\citep{li2025can}. For instance, a land-cover class is designated as absent if its total pixel or connected component count $N=0$, and conversely, any positive count ($N>0$) prompts the system to invoke either "object counting" or "existence verification" templates.

\textbf{Semantic Variable Embedding.} Quantitative results calculated by the logic engine are treated as dynamic variables and embedded into predefined natural language slots~\citep{li2025can, wang2025xlrs}. In target quantization tasks, the system populates the templates with the exact number of objects or distinct regions identified. For spatial localization, the engine compares the geometric centroids $(x, y)$ of two target entities to compute a relative displacement vector. The direction of this vector is then automatically mapped to semantic orientation terms and filled into the directional description template to ensure spatial consistency.
\vspace{1em}

\textbf{Threshold-Based Hard Logic Determination.} To address the specific physicochemical indicators of HSI, such as Normalized Difference Vegetation Index (NDVI) or bi-temporal change intensity, the rule-based route employs rigorous physical thresholds to determine ground-truth labels~\citep{dang2025benchmark}. The system evaluates whether calculated spectral feature values fall within predefined intervals (e.g., mapping $NDVI > 0.6$ to "Healthy Vegetation" or values exceeding the $3\sigma$ criterion to "Significant Change"), thereby automatically selecting the correct multiple-choice option (A/B/C/D) based on objective spectral evidence.

\subsection{MLLM-based QA Generation}

We employ \textbf{Qwen3-VL-Plus}~\citep{bai2023qwen} and \textbf{Qwen3-Max-Thinking}~\citep{qwen3maxthinking} as the generation backbones for producing candidate QA pairs. Instead of directly providing raw hyperspectral cubes, which are not natively supported by current MLLMs, we construct structured spectral-spatial descriptions for each image block. The input consists of complementary information extracted from the hyperspectral data, including pseudo-RGB representations~\citep{kang2020hyperspectral}, spectral indices (e.g., NDVI and related vegetation indicators)~\citep{rasti2020feature}, band-wise statistics (mean, standard deviation, and extrema), reflectance trends, and task-specific physical descriptors. These structured representations provide the factual basis for generation while preventing the MLLMs from relying solely on visual appearance or language priors.

The generation prompts share a common structure but are adapted to each QA type. The task instructions specify what evidence to use and what kind of question to construct; the accompanying evidence also changes with the task. For example, vegetation-health questions use red-edge and vegetation-index measurements, whereas anomaly questions use deviations from background spectral statistics. The model drafts the question, options, and answer, which are then checked against the supplied evidence.

\textbf{Representative generation prompt.}
Figure~\ref{fig:prompt_template} illustrates the prompt structure for spectral
feature recognition. Other QA types retain the evidence-grounding and output
requirements but use slightly different task instructions and relevant
spectral--spatial fields. The QA-generation evidence packet can contain
annotations and class-wise statistics used to establish candidate answers;
it is distinct from the label-free report supplied to evaluated MLLMs in
VSR$^2$. The generator is asked to return only a list of questions, option
sets, and answer letters. A deterministic post-processing script attaches
image paths, item and taxonomy identifiers, and other benchmark metadata,
then checks the output schema before expert review.

\begin{figure}[!p]
  \centering
\begin{lstlisting}[basicstyle=\ttfamily\scriptsize,breaklines=true,frame=single,
                    numbers=none,xleftmargin=1pt,xrightmargin=1pt]
[ROLE]
You are an expert in hyperspectral remote sensing and in constructing
visual question answering (VQA) benchmarks for Earth observation.

[EVIDENCE]
You are given one hyperspectral image block:
(1) a false-color RGB visualization, for judging spatial layout and
    land-cover composition;
(2) spectral-spatial statistics for the same block, including the
    class-wise band summaries and other measurements relevant to this
    question type.
Use only the supplied image and statistics. Do not invent classes,
measurements, or spatial details.

[TASK: SPECTRAL FEATURE RECOGNITION]
Generate multiple-choice questions about spectral properties in this
block. Ground the answer in the supplied class-wise band statistics,
reflectance trends, or spectral indices. Choose plausible alternatives
that the supplied evidence can distinguish.

Constraints:
- The question must not reveal or hint at the answer. Do not embed the
  answer, an equivalent expression of it, or a decisive cue in the stem.
- Distractors must be plausible and mutually exclusive, and exactly one
  option must be correct under the given evidence.
- The question must be self-contained and answerable from this block only.

[OUTPUT]
Return ONLY a JSON array of QA objects. Each object has exactly the
three fields shown below; the number of options may vary by task:
[
  {
    "question": "...",
    "options": {"A": "...", "B": "...", "C": "...", "D": "..."},
    "answer": "<one of the option letters>"
  }
]
\end{lstlisting}
  \caption{\textbf{Representative MLLM-assisted QA-generation prompt for spectral feature recognition.}
  Task instructions and evidence fields vary across QA types. The model returns
  only question, options, and answer; paths and taxonomy identifiers are added
  during deterministic post-processing. The four-option object illustrates the
  output format rather than fixing the option count for every task.}
  \label{fig:prompt_template}
\end{figure}

\begin{figure}[!p]
  \centering
\begin{lstlisting}[basicstyle=\ttfamily\scriptsize,breaklines=true,frame=single,
                    numbers=none,xleftmargin=1pt,xrightmargin=1pt]
You are an expert in hyperspectral remote sensing analysis.
You are provided with {total_images} images and
{num_report} spectral analysis report(s).

EVIDENCE PRIORITY RULE:
- The FIRST {num_rgb} image(s) are pseudo-color RGB representations
  and serve as the PRIMARY evidence. Base your initial judgment
  entirely on spatial patterns, colors, and textures from these
  RGB images.
- The SUBSEQUENT {num_pca} image(s) are PCA-enhanced versions
  highlighting spectral variance. Use them ONLY when the RGB images
  are ambiguous or insufficient to distinguish between options.
- The spectral report(s) contain quantitative measurements. Consult
  them ONLY as a last resort when visual evidence from both RGB and
  PCA cannot resolve the answer.

ANALYSIS GUIDELINES:
- Start with RGB -> only escalate to PCA -> only escalate to Report if needed.
- Do NOT let the report override clear visual evidence from the RGB image.

{RGB and PCA image inputs}

Spectral Report Content:
{report_content}

Question:
{question}

Options:
{option_lines}

You MUST choose your answer ONLY from the available options: {valid_options}.
Output exactly ONE uppercase letter corresponding to your chosen option.
Do NOT provide any explanation, reasoning, or additional text.
Your entire response must be a single character.

Answer:
\end{lstlisting}
  \caption{\textbf{Representative fusion prompt for the full VSR$^{2}$ input.} The
  placeholders \{total\_images\}, \{num\_report\}, \{num\_rgb\}, \{num\_pca\},
  \{report\_content\}, \{question\}, \{option\_lines\} and \{valid\_options\}
  are populated from the current sample. Other input settings omit unavailable
  view descriptions and inputs while retaining the question, options, and
  one-letter output requirement.}
  \label{fig:fusion_prompt}
\end{figure}

\subsection{Quality Control of MLLM-based QA}

The generated QA candidates are not directly included in HM-Bench. Instead, each candidate undergoes expert verification against the corresponding hyperspectral evidence and predefined task requirements.

During expert verification, reviewers examine candidates from four aspects: (1) \textbf{evidence grounding}, i.e., whether the answer can be supported by spectral-spatial information extracted from the original hyperspectral sample; (2) \textbf{factual correctness}, i.e., whether the generated interpretation is consistent with available annotations, spectral statistics, or physical indicators; (3) \textbf{task relevance}, i.e., whether the question evaluates the intended hyperspectral capability rather than superficial visual understanding; and (4) \textbf{format compliance}, i.e., whether the QA pair follows the predefined benchmark schema and answer protocol.

Candidates are rejected if they contain unsupported spectral claims, trivial questions, invalid answer formats, or ambiguous task definitions. Representative rejection cases are summarized in Table~\ref{tab:rejection_examples}.

Importantly, MLLMs serve only as linguistic generation tools rather than annotation authorities. Final QA pairs are retained only when their answers can be traced back to explicit spectral-spatial evidence or verified domain knowledge, ensuring that benchmark supervision remains evidence-grounded rather than relying on unconstrained language generation.

\subsection{Evidence Sufficiency Across Task Categories}
\label{app:evidence_sufficiency}

An answer can be correct with respect to the original hyperspectral cube and its annotations without being uniquely recoverable from an RGB image, a PCA composite, or a summary report. We therefore distinguish three checks: \emph{ground-truth validity} against the cube, label map, or deterministic measurements; \emph{view-level sufficiency}, meaning that the actual inputs provide cues at the spatial and spectral granularity needed to separate the answer from its alternatives; and \emph{question clarity}, meaning that the wording and options admit a single supported interpretation. Table~\ref{tab:evidence_requirements} states the view-level evidence to examine for each task. It specifies assessment criteria, not measured pass rates for all retained questions.

\begin{table*}[!t]
\centering
\small
\setlength{\tabcolsep}{6pt}
\caption{\textbf{Evidence to assess view-level answerability by task.} The required cue must be present at the granularity requested by the question and separate the correct option from the alternatives. The table is an audit checklist, not a claim that every retained item passes under every input configuration.}
\label{tab:evidence_requirements}
\begin{tabular}{lp{0.75\textwidth}}
\toprule
Task & Evidence required in the delivered views \\
\midrule
SFR & Pixel- or region-aligned spectral signature, band statistics, or indices at the queried location \\
LCC & Discriminative spatial or spectral cues at the requested class granularity \\
PD & Evidence for both the presence and the absence of the target class, including small regions \\
Count & Individually separable targets or explicit quantitative region and area statistics \\
OLR & Identifiable objects with unambiguous relative positions \\
RD & Visible boundaries at the spatial precision required by the answer options \\
SAD & Local spectral deviation and a defined anomaly reference or criterion \\
SU & Endmember evidence and, where requested, mixture or abundance information \\
VHS & Spectral indicators or calibrated measures of vegetation health and stress \\
EPS & Relevant pollution indicators with a defined severity scale \\
BCI & Registered observations from both dates and evidence of the stated transition \\
CAL & Spatially identifiable change patterns at the requested location granularity \\
CSA & Temporal statistics or quantitative evidence of class transitions \\
\bottomrule
\end{tabular}
\end{table*}

\appsection{Details of VSR$^2$}
\label{app:vsr}

To expose hyperspectral information to off-the-shelf MLLMs without any retraining, we propose VSR$^2$, a training-free Visual--Spectral--Report Reasoning framework. VSR$^2$ converts each hyperspectral sample into three aligned views: an RGB image, a PCA-based spectral view, and a structured textual report. The three views are designed to be complementary: the RGB view preserves natural visual semantics and spatial layout, the PCA view summarizes dominant spectral variation in an image-compatible form, and the report view provides explicit spectral--spatial evidence through language. This section presents the detailed construction process of each view and the overall inference protocol.

\subsection{Framework Overview}

Given a hyperspectral cube $\mathbf{X} \in \mathbb{R}^{H \times W \times B}$, VSR$^2$ transforms it into three complementary representations that can be directly consumed by existing MLLMs. The overall design follows two principles: (1) preserving as much spectral evidence as possible in a training-free manner, and (2) maintaining compatibility with the native image--text interface of off-the-shelf models. Specifically, each sample is converted into an RGB view for visual semantics, a PCA view for compact spectral variation visualization, and a structured report for explicit numerical and spatial evidence. These representations are then used either individually or jointly under a unified prompting protocol.

\subsection{RGB View Construction}
\label{app:vsr_rgb}

The RGB view in VSR$^2$ provides a natural-image-like representation of hyperspectral samples, enabling off-the-shelf MLLMs to leverage their native visual understanding without retraining. Since hyperspectral images lack standard RGB channels, we construct a pseudo-RGB image by selecting three representative bands from the spectral cube.

\begin{figure}[t]
    \centering
    \includegraphics[width=0.8\linewidth]{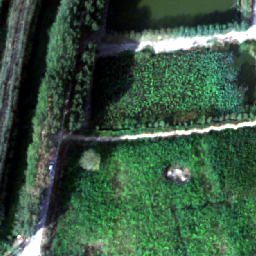}
    \caption{Example of the pseudo-RGB view generated from a hyperspectral sample.}
    \label{fig:rgb_view_example}
\end{figure}

Figure~\ref{fig:rgb_view_example} illustrates the spatial appearance preserved by this band-selection procedure; its colors are a pseudo-RGB rendering of selected HSI bands rather than a conventional camera photograph.

Given a hyperspectral cube $\mathbf{X} \in \mathbb{R}^{H \times W \times B}$, the core challenge is determining which band indices best correspond to the visible red, green, and blue wavelengths. We adopt a unified selection criterion: for each target wavelength $\lambda^* \in \{650, 550, 450\}$\,nm, the optimal band index is
\[
b^* = \arg\min_{b \in \{1,\dots,B\}} |\lambda_b - \lambda^*|,
\]
where $\lambda_b$ denotes the effective wavelength of band $b$. The four-level priority below governs how $\lambda_b$ is obtained or estimated, ensuring compatibility across heterogeneous datasets. 

\paragraph{Band Selection Priority.}
We determine $\lambda_b$ through the following hierarchical strategy:

\begin{itemize}[leftmargin=15pt]
    \item \textbf{Metadata-Driven Selection.}
    When explicit spectral calibration metadata is available, we directly use the recorded wavelength array as $\lambda_b$. This includes automatic unit normalization (e.g., $\mu$m to nm) and robust key matching across varying metadata conventions.

    \item \textbf{Dataset-Specific Prior Knowledge.}
    In the absence of embedded metadata, we retrieve the known spectral configuration from a curated dataset specification registry using the dataset's unique identifier, providing documented sensor characteristics as a reliable proxy.

    \item \textbf{Identity-Inferred Approximation.}
    If no explicit identifier is provided, we infer the dataset origin via filename pattern matching. Upon identification, $\lambda_b$ is reconstructed through linear interpolation over the sensor's documented spectral range.

    \item \textbf{Heuristic Fallback.}
    As a last resort for uncharacterized data, we assume the spectral dimension is approximately ordered by increasing wavelength and assign $\lambda_b$ proportionally across the documented or inferred spectral range. If even the spectral range is unknown, we default to selecting bands at the 30\%, 20\%, and 10\% positions of the spectral dimension for R, G, and B, respectively. 
\end{itemize}

\subsection{PCA View Construction}
\label{app:image_processing}

The PCA view in VSR$^2$ is designed to expose dominant hyperspectral variation to off-the-shelf MLLMs through a purely image-based interface. Since standard vision encoders cannot directly process a hyperspectral cube $\mathbf{X} \in \mathbb{R}^{H \times W \times B}$ with hundreds of spectral bands, we compress the spectral dimension using principal component analysis (PCA)~\citep{mackiewicz1993principal} and render the resulting component maps as a structured composite image. This design preserves major spectral variation while remaining fully compatible with the native visual input format of existing MLLMs.

\paragraph{Preprocessing and standardization.}
Given a hyperspectral cube $\mathbf{X}$, we first reshape it into a two-dimensional matrix $\mathbf{X}_{2D} \in \mathbb{R}^{HW \times B}$, where each row corresponds to the spectrum of one pixel. Before applying PCA, we perform two preprocessing steps to improve numerical stability and component quality:

\begin{itemize}[leftmargin=15pt]
    \item \textbf{Band filtering.} Bands with zero variance are removed, since they do not contribute meaningful information to covariance estimation and may introduce numerical instability.
    \item \textbf{Z-score normalization.} Each retained band is standardized to zero mean and unit variance:
    \begin{equation}
        \mathbf{X}_{\text{scaled}} = \frac{\mathbf{X}_{2D} - \boldsymbol{\mu}}{\boldsymbol{\sigma}},
    \end{equation}
    where $\boldsymbol{\mu}$ and $\boldsymbol{\sigma}$ denote the per-band mean and standard deviation, respectively.
\end{itemize}

This normalization is important because hyperspectral bands often have different dynamic ranges and variances; without standardization, PCA may be dominated by a small subset of high-energy bands rather than reflecting the overall spectral structure.
\begin{figure}[t]
    \centering
    \includegraphics[width=\columnwidth]{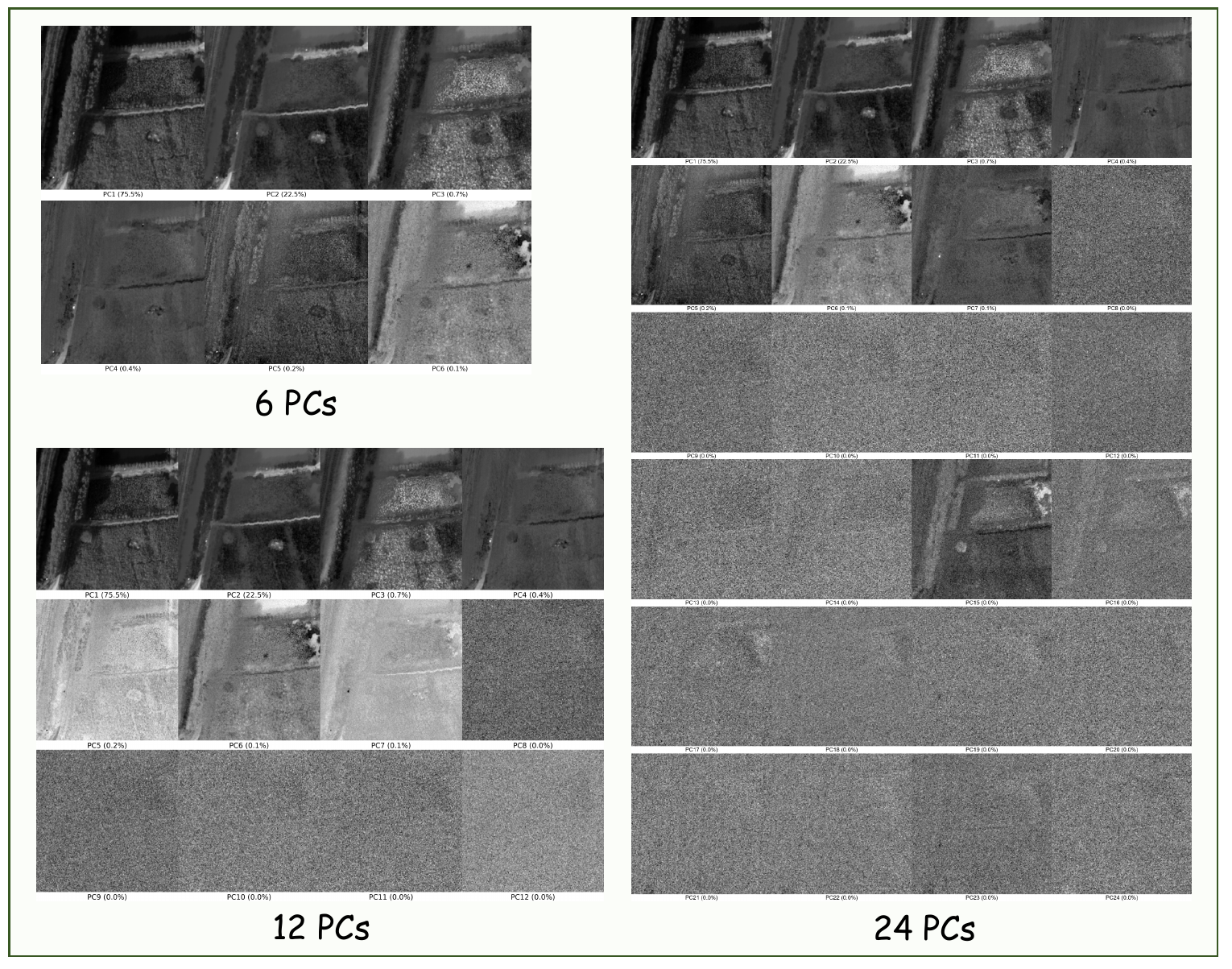}
    \caption{Examples of PCA-based spectral views constructed using the top 6, 12, and 24 principal components from the same hyperspectral sample. }
    \label{fig:pca_6_12_24_examples}
\end{figure}
\paragraph{Principal component extraction.}
We then apply PCA to the standardized matrix $\mathbf{X}_{\text{scaled}}$ and project the original spectra onto the principal directions. The resulting principal component scores are reshaped back to spatial maps of size $H \times W$, producing one image-like map for each principal component. Let $\mathrm{PC}_k \in \mathbb{R}^{H \times W}$ denote the $k$-th component map. Each map is independently normalized to the $[0,1]$ range and rendered as an 8-bit grayscale image to ensure consistent visual contrast across components.

\begin{table*}[t]
\centering
\caption{PCA explained variance summary across datasets in HM-Bench. We report the cumulative explained variance (EV, \%) captured by the top 1, 3, 6, 12, and 24 principal components, together with the number of principal components required to reach 95\% explained variance.}
\label{tab:pca_variance_summary}
\footnotesize
\setlength{\tabcolsep}{3pt}
\begin{tabular*}{\textwidth}{@{\extracolsep{\fill}}lccccccc@{}}
\toprule
Dataset & \#Bands & EV@1 & EV@3 & EV@6 & EV@12 & EV@24 & \#PCs for 95\% EV \\
\midrule
BayArea\_2013~\citep{cooper2020bayarea}           & 224 & 82.52 & 99.81 & 99.96 & 99.99 & 99.99 & 2  \\
BayArea\_2015~\citep{cooper2020bayarea}           & 224 & 54.10 & 99.44 & 99.81 & 99.93 & 99.96 & 2  \\
Botswana~\citep{hou2021hyperspectral}                & 145 & 94.62 & 99.40 & 99.67 & 99.78 & 99.87 & 2  \\
HanChuan~\citep{zhong2020whu}                & 274 & 92.21 & 99.45 & 99.67 & 99.74 & 99.81 & 2  \\
Hermiston\_2004~\citep{hou2021hermiston}         & 242 & 76.77 & 93.59 & 96.67 & 97.99 & 98.97 & 4  \\
Hermiston\_2007~\citep{hou2021hermiston}         & 242 & 73.11 & 90.17 & 95.29 & 97.19 & 98.56 & 6  \\
Holden~\citep{li2022stepwise}                  & 440 & 95.61 & 98.89 & 99.56 & 99.76 & 99.88 & 1  \\
HongHu~\citep{zhong2020whu}                  & 270 & 61.95 & 96.90 & 98.11 & 98.78 & 99.25 & 2  \\
Houston\_13~\citep{debes2014hyperspectral}             & 144 & 59.15 & 99.66 & 99.94 & 99.98 & 99.99 & 2  \\
Houston\_18~\citep{xu2019advanced}             & 48  & 74.62 & 99.65 & 99.93 & 99.98 & 99.99 & 2  \\
Indianpines~\citep{baumgardner2015220}             & 200 & 58.49 & 90.94 & 93.99 & 96.93 & 98.74 & 8  \\
Ksc~\citep{kennedy2024ksc}                     & 176 & 26.18 & 41.79 & 56.49 & 70.26 & 86.21 & 37 \\
LongKou~\citep{zhong2020whu}                 & 270 & 95.47 & 99.85 & 99.93 & 99.96 & 99.97 & 1  \\
Nilifossae~\citep{li2022stepwise}              & 425 & 84.03 & 98.71 & 99.62 & 99.86 & 99.94 & 3  \\
Paviac~\citep{hou2021hyperspectral}                  & 102 & 75.03 & 99.31 & 99.78 & 99.93 & 99.97 & 2  \\
Paviau~\citep{hou2021hyperspectral}                  & 103 & 48.65 & 98.42 & 99.35 & 99.81 & 99.94 & 2  \\
Salinas~\citep{hou2021hyperspectral}                 & 204 & 87.89 & 99.59 & 99.85 & 99.94 & 99.96 & 2  \\
SantaBarbara\_2013~\citep{lopez2019gpu}      & 224 & 80.99 & 94.45 & 96.39 & 98.09 & 99.38 & 4  \\
SantaBarbara\_2014~\citep{lopez2019gpu}      & 224 & 82.94 & 96.50 & 97.65 & 98.65 & 99.61 & 2  \\
Urban~\citep{zhu2017hyperspectral}                   & 162 & 68.84 & 98.68 & 99.66 & 99.87 & 99.94 & 2  \\
Utopia~\citep{li2022stepwise}                  & 432 & 70.64 & 96.78 & 98.15 & 99.02 & 99.47 & 3  \\
Washington~\citep{multispec_dc}              & 191 & 83.44 & 99.49 & 99.87 & 99.96 & 99.98 & 2  \\
Xiongan~\citep{yi2020aerial}                 & 250 & 60.56 & 99.34 & 99.70 & 99.75 & 99.77 & 2  \\
\bottomrule
\end{tabular*}
\end{table*}

\paragraph{Why 12 principal components?}
A key design question is how many principal components should be retained for the PCA view. To answer this, we analyze the explained variance across all datasets in HM-Bench. The complete per-dataset statistics are reported in Table~\ref{tab:pca_variance_summary}, including the cumulative explained variance captured by the top 1, 3, 6, 12, and 24 principal components, as well as the number of components required to reach 95\% explained variance.

Overall, the first few components already capture most of the spectral variance for the majority of datasets. In particular, the top 12 principal components typically retain a very high proportion of variance, while remaining compact enough to be organized into a visually interpretable composite. This makes 12 components a practical trade-off between spectral coverage and image complexity. Using too few components (e.g., 6) may discard discriminative spectral cues needed for challenging reasoning tasks, while using too many components (e.g., 24) enlarges the visual input and introduces lower-energy components that are often noisier and less semantically interpretable.

We further validate this choice empirically in Table~\ref{tab:pca_component_ablation}, which compares model performance under PCA views constructed from 6, 12, and 24 components. Across multiple MLLMs and multimodal input settings, using 12 components generally provides the best or most stable performance. This result is consistent with the explained-variance statistics and supports our choice of 12 principal components as the default setting in VSR$^2$.

\begin{table}[t]
\centering
\caption{Ablation on the number of principal components in the PCA view. Using 12 components yields the best or most stable performance across models and input configurations.}
\label{tab:pca_component_ablation}
\renewcommand{\arraystretch}{1.15}
\setlength{\tabcolsep}{6pt}
\small 
\begin{tabular}{lp{2.cm}ccc}
\toprule
Model & Input Mod. & \multicolumn{3}{c}{\# PCs} \\
\cmidrule(lr){3-5}
      &               & 6     & \textbf{12}    & 24    \\
\midrule
\multirow{3}{*}{Qwen3-VL-4B} 
    & PCA-only        & 39.57 & 40.96 & 40.34 \\
    & w/o Report      & 39.23 & 42.97 & 42.63 \\
    & VSR²            & 40.47 & 43.23 & 41.98 \\
\addlinespace[2pt]
\midrule
\multirow{3}{*}{InternVL3.5} 
    & PCA-only        & 37.08 & 43.08 & 38.10 \\
    & w/o Report      & 42.91 & 43.80 & 41.72 \\
    & VSR²            & 44.28 & 45.09 & 44.84 \\
\addlinespace[2pt]
\midrule
\multirow{3}{*}{LLaVA-v1.5} 
    & PCA-only        & 31.19 & 34.83 & 31.14 \\
    & w/o Report      & 36.83 & 37.06 & 35.15 \\
    & VSR²            & 38.95 & 40.60 & 40.18 \\
\bottomrule
\end{tabular}
\end{table}

\paragraph{Task-level sensitivity to the number of components.}
Table~\ref{tab:pca_task_sensitivity} complements the overall accuracies in Table~\ref{tab:pca_component_ablation} with the 13 task accuracies under full VSR$^2$. The 12-component setting obtains the highest accuracy in 30 of the 39 model--task comparisons. The best dimensionality is nevertheless task dependent: for example, Qwen3-VL-4B achieves its highest spectral feature recognition accuracy with 24 components. These results support 12 components as a default across the tested settings rather than as an optimum for every task. Task accuracies and question-weighted overall accuracy measure different quantities; the latter is reported separately in Table~\ref{tab:pca_component_ablation}.

\begin{table*}[t]
\centering
\scriptsize
\setlength{\tabcolsep}{2.5pt}
\caption{\textbf{Task-level sensitivity to PCA dimensionality under full VSR$^2$.} Values are accuracies (\%) for 6, 12, and 24 principal components. Task abbreviations follow Table~\ref{tab:full_results}; the best result within each model and task is bold.}
\label{tab:pca_task_sensitivity}
\begin{tabular}{llrrrrrrrrrrrrr}
\toprule
Model & PCs & SFR & LCC & PD & Count & OLR & RD & SAD & SU & VHS & EPS & BCI & CAL & CSA \\
\midrule
Qwen3-VL-4B & 6 & 39.95 & 25.60 & 34.24 & 42.55 & 30.75 & 34.03 & 67.95 & 39.54 & 53.28 & 64.35 & 38.35 & 13.61 & 40.83 \\
 & 12 & 42.03 & 28.07 & 36.27 & \textbf{46.50} & \textbf{34.17} & 36.02 & \textbf{70.02} & 41.26 & \textbf{57.32} & \textbf{67.41} & 40.45 & \textbf{17.87} & \textbf{44.57} \\
 & 24 & \textbf{48.36} & \textbf{30.25} & \textbf{42.70} & 37.84 & 30.85 & \textbf{36.67} & 64.39 & \textbf{41.35} & 44.41 & 60.58 & \textbf{42.14} & 10.03 & 40.67 \\
\addlinespace
InternVL3.5-14B & 6 & 43.21 & 28.24 & 48.48 & 46.54 & 28.39 & \textbf{50.08} & 67.27 & 42.87 & 60.48 & 53.40 & 40.48 & 20.58 & 40.00 \\
 & 12 & \textbf{44.33} & \textbf{29.53} & \textbf{50.25} & \textbf{48.88} & \textbf{30.57} & 46.29 & \textbf{68.93} & \textbf{44.53} & \textbf{61.09} & \textbf{56.07} & \textbf{42.04} & \textbf{21.53} & \textbf{42.12} \\
 & 24 & 43.47 & 27.93 & 48.82 & 45.70 & 28.39 & 46.92 & 65.16 & 42.65 & 60.53 & 53.94 & 40.08 & 20.41 & 41.67 \\
\addlinespace
LLaVA-v1.5-7B & 6 & 38.76 & \textbf{38.76} & 44.94 & 33.13 & 26.50 & 29.51 & 60.75 & 34.81 & 55.53 & 50.24 & 29.82 & 9.22 & 36.33 \\
 & 12 & \textbf{42.23} & 38.05 & \textbf{47.63} & \textbf{37.53} & \textbf{30.69} & \textbf{32.64} & \textbf{66.63} & \textbf{37.02} & 56.24 & \textbf{53.62} & \textbf{32.40} & \textbf{12.99} & \textbf{40.10} \\
 & 24 & 40.16 & 35.17 & 46.43 & 35.62 & 27.92 & 31.94 & 62.15 & 36.73 & \textbf{56.80} & 52.31 & 32.36 & 11.05 & 39.74 \\
\bottomrule
\end{tabular}
\end{table*}

\paragraph{Composite image construction.}
After extracting the top 12 principal components, we arrange the corresponding component maps into a single composite image using three rows and four columns, as illustrated in Fig.~\ref{fig:pca_6_12_24_examples}. Each panel preserves the original spatial structure of the component map, and the panels are ordered according to component rank. The explained variance ratio of each component can be annotated in the panel for interpretability. The final composite image therefore provides a compact visual summary of dominant spectral variation, while remaining easy for MLLMs to parse through their standard visual encoders.

Figure~\ref{fig:pca_6_12_24_examples} compares representative PCA composites constructed from 6, 12, and 24 principal components. The 6-component setting under-represents fine spectral variation, whereas 24 components introduce additional low-variance patterns with diminishing visual and task-level benefit. In contrast, retaining 12 principal components offers the most balanced representation between information preservation and redundancy reduction.
\begin{figure*}[t]
    \centering
    \includegraphics[width=\textwidth]{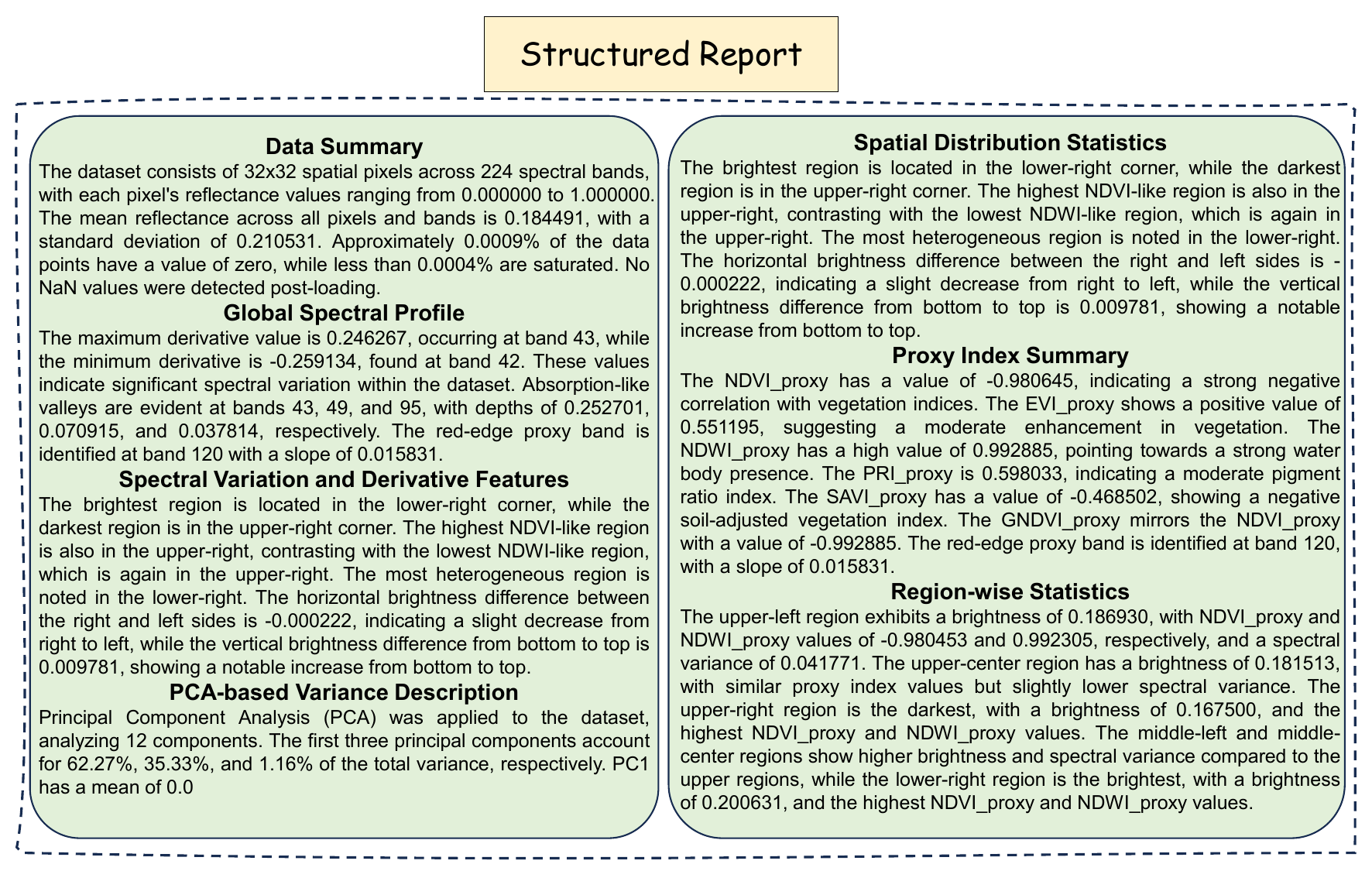}
    \caption{An example of structured report input generated from the same data block as the PCA composite image above.}
    \label{fig:structured_report_example}
\end{figure*}
\paragraph{Role in VSR$^2$.}
The PCA view complements the RGB view by exposing beyond-RGB spectral variation in an image-compatible form. While the RGB view preserves natural appearance and spatial semantics, the PCA view highlights latent spectral structure that may be invisible in pseudo-color rendering. This makes it particularly useful for tasks that require sensitivity to material differences, subtle region boundaries, or cross-band variation patterns. Importantly, the entire PCA view construction pipeline is training-free and can be applied uniformly across heterogeneous hyperspectral datasets.

\paragraph{Visible-spectrum control.}
To examine the contribution associated with non-visible bands, we also construct a control PCA view using only the visible-range bands selected from the original hyperspectral cube before applying PCA. We compare RGB plus this visible-spectrum PCA view with RGB plus the full-spectrum PCA view under the same evaluation protocol, excluding the report in both settings. Thus, the RGB input is retained while the spectral range used to construct the PCA view changes. Results for three representative MLLMs are reported in Table~\ref{tab:spectral_range_question_only}.

\subsection{Structured Report Generation}
\label{app:report_generation}

The report view in VSR$^2$ converts each hyperspectral sample into a structured textual description that summarizes quantitative spectral--spatial evidence in language form. Unlike generic captioning, this view is not intended to produce semantic scene descriptions or task-specific answers. Instead, it serves as an auxiliary evidence channel that exposes information from the hyperspectral cube in a form that language models can readily process, while minimizing hallucination and premature inference.

As shown in Fig.~\ref{fig:structured_report_example}, the report is generated through a two-stage pipeline: \textbf{quantitative feature extraction} followed by \textbf{controlled text generation}.

\paragraph{Stage 1: Quantitative feature extraction.}
Given a hyperspectral cube, we first derive a structured set of numerical descriptors that characterize its spectral and spatial properties. The descriptors are organized into a JSON-like intermediate representation and include the following categories:

\begin{itemize}[leftmargin=15pt]
    \item \textbf{Data quality indicators.} We record basic information such as cube shape, the number of invalid values (e.g., NaNs), and the proportion of saturated or degenerate values. The statistics help characterize the reliability and dynamic range of the sample.
    
    \item \textbf{Global spectral descriptors.} We compute summary statistics of the global spectral signature, including the mean spectrum, spectral entropy, and first-order derivative statistics that capture major slope changes and local extrema across wavelengths~\citep{rasti2020feature}. These features provide a compact description of the overall spectral behavior of the sample.
    
    \item \textbf{Proxy spectral indices.} To provide task-relevant cues without relying on dataset-specific semantic annotations, we compute a set of commonly used ratio-based indices, including EVI, NDVI, NDWI, PRI, SAVI, and GNDVI~\citep{rasti2020feature, hong2021multimodal}. Because the included datasets vary substantially in spectral resolution and wavelength coverage, these indices are computed using approximate band positions (e.g., blue, green, red, and near-infrared proxies defined by relative locations in the spectrum) rather than fixed sensor-specific bands. This design preserves cross-dataset applicability while retaining informative comparative signals.
    
    \item \textbf{Coarse spatial distribution statistics.} To describe regional heterogeneity, we divide the image into a $3 \times 3$ spatial grid and compute descriptive statistics such as brightness and variance for each sub-region. This allows the report to encode coarse spatial structure rather than only global spectral summaries.
\end{itemize}

Importantly, all extracted quantities are directly computed from the hyperspectral data itself and do not rely on class labels, external metadata, or manually annotated semantic descriptions.

\paragraph{Stage 2: Controlled text generation.}
The extracted quantitative features are then provided to LLaVA-1.5-7B to produce a structured textual report. The generation template presents data-summary, spectral-profile, proxy-index, spatial-grid, and PCA fields, while omitting file paths, dataset identifiers, class proportions, ground-truth labels, and the benchmark questions and answers. The goal of this stage is to convert numerical descriptors into a readable evidence summary that can be consumed together with other views.

To maximize objectivity and reproducibility, we employ a constrained generation protocol with the following principles:

\begin{itemize}[leftmargin=15pt]
    \item \textbf{Objective input only.} The model receives the computed numerical descriptors and derived statistics, without access to ground-truth labels, scene names, class proportions, or task-specific annotations.
    
    \item \textbf{Deterministic decoding.} We use deterministic generation (e.g., greedy decoding) to avoid stochastic variation across runs and ensure that the same hyperspectral sample always yields the same report under the same pipeline.
    
    \item \textbf{Constraint-aware prompting.} The prompting template explicitly instructs the model to summarize observed evidence rather than speculate about semantic categories or downstream answers. This helps reduce hallucination and keeps the report grounded in the input statistics.
\end{itemize}

\paragraph{Prompting rationale.}
The report view differs fundamentally from the RGB and PCA views: instead of asking the model to interpret an image, it presents the model with an explicit textual summary of spectral--spatial evidence. Accordingly, we use a report-oriented prompt that frames the input as a quantitative hyperspectral analysis and encourages evidence-based reasoning. This design makes the report view particularly valuable when spectral distinctions are difficult to infer reliably from visual appearance alone.

\paragraph{Role in VSR$^2$.}
The structured report complements the RGB and PCA views by exposing information that is difficult to encode visually, such as global statistical trends, spectral index values, derivative-based cues, and coarse regional summaries. In this sense, it acts as an interpretable bridge between raw hyperspectral measurements and language-based reasoning. Since the report is generated entirely from the input cube and does not require retraining or hyperspectral-specific supervision, it remains fully consistent with the training-free design of VSR$^2$.

\paragraph{Report reliability checks and manual review.}
The feature-only input policy does not guarantee that a language generator will obey every constraint. We therefore ran automated checks on all 2,648 generated reports and manually reviewed a stratified sample of 359 question--report pairs spanning the 13 tasks against their answer options. Table~\ref{tab:report_audit} separates the automated scans, manual review, and numerical verification. The manual review found no direct disclosure of a correct option, but identified 16 answer-indicative cues (4.5\%); this distinction matters because a verbalized proxy index can narrow an answer without reproducing it. We report inter-reviewer agreement before adjudication and will release the generation template and code, feature schema, check scripts, and reviewed annotations with the benchmark for inspection.

\begin{table*}[!t]
\centering
\small
\setlength{\tabcolsep}{5pt}
\caption{\textbf{Report reliability checks.} Automated scans cover all 2,648 reports; 359 QA-linked cases were manually reviewed. Agreement is measured on independent reviewer judgments before adjudication. Numerical matching allowed derived cumulative PCA values and rounding after case-wise review of the 25 initially unmatched claims.}
\label{tab:report_audit}
\begin{tabular}{p{0.35\textwidth}p{0.55\textwidth}}
\toprule
Check & Finding \\
\midrule
Automated answer-string scan & 0 direct-answer statements among 16,094 lexical-answer items \\
Automated semantic scan & 1,790/2,648 reports (67.6\%) contain proxy-index interpretations; 20/2,648 (0.76\%) contain residual prompt text \\
Manual review of QA-linked reports (359 cases) & 0 direct disclosures; 16 answer-indicative cues (4.5\%); 138 generic interpretations (38.4\%); 205 with no relevant cue (57.1\%) \\
Inter-reviewer agreement before adjudication & Answer-text presence: 99.2\%, $\kappa=0.80$; cue level: 81.9\%, $\kappa=0.65$ \\
Numerical fidelity & 94,852/94,853 claims valid or legitimately derived; one decimal-shift error (0.0011\% of claims), affecting one report \\
\bottomrule
\end{tabular}
\end{table*}

The 25 initially unmatched numerical claims comprise 12 cumulative PCA statistics, 12 rounding cases, and one decimal-shift error, which we identified for correction. The residual prompt text and answer-indicative cues show that prompt constraints are imperfect; these checks bound observed failures under the stated procedures rather than proving an absence of all indirect leakage. A controlled removal of generated descriptive wording is reported in Appendix~\ref{app:report_ablation}.

\subsection{Cross-View Alignment and Complementarity}

The three views in VSR$^2$ are spatially aligned and semantically complementary. All views are derived from the same hyperspectral sample, ensuring that they correspond to the same underlying scene content. The RGB view mainly supports visually grounded semantic understanding and spatial localization, the PCA view emphasizes dominant cross-band variation and is therefore particularly useful for spectrally sensitive tasks, and the report view provides explicit quantitative evidence that can complement visual observations. This design allows VSR$^2$ to expose beyond-RGB hyperspectral information through interfaces that current MLLMs can already process. Building upon these three complementary views, we next describe the inference protocol under which VSR² is evaluated.

\subsection{Inference Protocol of VSR$^2$}
\label{app:inference_protocol}

We evaluate VSR$^2$ under a unified zero-shot multiple-choice reasoning protocol. For each hyperspectral sample, the model receives one or more views derived from the same underlying cube, depending on the experimental setting. These views may include the pseudo-RGB image, the PCA view, and the structured report. No parameter updates, task-specific fine-tuning, or hyperspectral pretraining are introduced at inference time; thus, all gains come from representation design and inference-time prompting alone.

\paragraph{Input configurations.}
VSR$^2$ supports both single-view and multi-view inference. In the single-view setting, the model receives only one modality, such as RGB-only, PCA-only, or report-only input. In the multi-view setting, multiple representations of the same hyperspectral sample are concatenated into a unified prompt. The ordering of the inputs is fixed to ensure consistency across samples and models. When both RGB and PCA images are used, RGB images are placed first and PCA images are placed afterwards; when reports are also used, the report text is appended after the image inputs.

\paragraph{Fusion prompt design.}
To guide the model in combining heterogeneous evidence without over-relying on any single modality, we adopt a controlled fusion prompt with an explicit evidence-priority rule. In the full input configuration, the prompt informs the model that:
\begin{itemize}[leftmargin=15pt]
    \item the pseudo-RGB image(s) serve as the \textbf{primary evidence}, and the model should first form its judgment based on spatial patterns, colors, and textures;
    \item the PCA image(s) serve as \textbf{secondary evidence}, and should be consulted only when RGB evidence is ambiguous or insufficient;
    \item the structured report serves as \textbf{auxiliary textual evidence}, and should be used only as a final reference when the visual evidence from RGB and PCA remains inconclusive.
\end{itemize}

\textbf{Adaptation to the six input configurations.}
All six configurations use the same question, options, and single-letter answer requirement. Figure~\ref{fig:fusion_prompt} shows the full RGB+PCA+Report example. For the other settings, we omit the unavailable input sections and their evidence-priority instructions, and update the modality counts and ordering description. Report-only, PCA-only, and RGB-only retain one view; RGB+Report and RGB+PCA retain the indicated pair. These edits change prompt length and input budget, so RGB-only versus full VSR$^2$ remains a system-level comparison. The matched controls in Appendix~\ref{app:controlled_contributions} isolate narrower changes within a common prompt structure.

The full-input example in Fig.~\ref{fig:fusion_prompt} illustrates the priority rule; the QA-generation example in Fig.~\ref{fig:prompt_template} serves the separate benchmark-construction stage.

This evidence-priority design reflects the intended roles of the three views in VSR$^2$: RGB preserves natural appearance and coarse spatial semantics, PCA highlights latent spectral variation beyond visible color, and the report summarizes quantitative spectral--spatial descriptors in text form. By explicitly specifying this hierarchy, we encourage the model to use the modalities in a complementary rather than competitive manner.

\paragraph{Modality-specific prompting.}
Modality-specific variants of the representative prompt in Fig.~\ref{fig:fusion_prompt} retain the same question, options, and answer format. Each variant removes unavailable view sections and adjusts the modality counts and evidence-priority wording to match the supplied inputs. These adaptations change prompt length and, in the full configuration, add views.

Importantly, these prompts do not inject task-specific labels, hidden supervision, or external metadata. Their role is only to clarify the nature of each input view and to standardize how the model attends to complementary evidence.

\paragraph{Standardized output constraint.}
All experiments use the same multiple-choice answer format. Given a question and its candidate options, the model is explicitly constrained to output \emph{exactly one uppercase letter} corresponding to one of the valid options. No explanation, reasoning trace, or additional text is allowed. This output restriction reduces formatting variance across models and facilitates automatic evaluation.

\paragraph{Decoding and reproducibility.}
We use the decoding configuration described in Appendix~\ref{app:evaluation}, including temperature zero when supported by the model provider. These settings reduce sampling variability, but do not by themselves guarantee identical outputs across repeated calls or inference environments.

\paragraph{Discussion.}
The fusion prompt in VSR$^2$ should be viewed as an inference-time coordination mechanism rather than a source of external knowledge. It does not alter model parameters or provide hidden supervision; instead, it organizes multiple views of the same hyperspectral sample into a consistent reasoning protocol. This design allows us to evaluate whether off-the-shelf MLLMs can better exploit complementary hyperspectral representations under a training-free setting.

\appsection{Details of the Evaluation Setup}
\label{app:evaluation}

We evaluate all models under a unified zero-shot protocol. Since the view construction pipeline and fusion prompting strategy have been described in the preceding sections, we focus here on implementation and evaluation details.

\subsection{Environment and Deployment}
All experiments were conducted within the same evaluation period using a unified inference pipeline. Open-source models were deployed locally with vLLM on up to four NVIDIA GeForce RTX 4090 GPUs (24 GB each), under CUDA 12.4 and NVIDIA driver 550.142. Closed-source models were accessed through their official online APIs. No model was fine-tuned or adapted to hyperspectral data.

\subsection{Evaluated Models}
We evaluate 17 MLLM baselines together with one random baseline. The closed-source models include Claude-Sonnet-4-6~\citep{anthropic2026claudesonnet46}, GPT-5.4-Mini~\citep{openai2026gpt54mini}, and Grok-4~\citep{de2025grok}. The open-source models include InternVL2-8B~\citep{chen2024internvl}, InternVL2.5-4B~\citep{chen2024internvl}, InternVL3-9B~\citep{chen2024internvl}, InternVL3-14B~\citep{chen2024internvl}, InternVL3.5-14B~\citep{chen2024internvl}, Kimi-VL-A3B-Instruct~\citep{team2025kimi}, Qwen2.5-VL-7B~\citep{bai2023qwen}, Qwen3-VL-4B~\citep{bai2023qwen}, LLaVA-NeXT-8B~\citep{li2024llava}, LLaVA-v1.5-7B~\citep{liu2024improved}, Geochat-7B~\citep{kuckreja2024geochat}, GeoLLaVA-8K~\citep{wang2025geollava}, GLM-4.6V-Flash-9B~\citep{hong2025glm}, and Nemotron-Nano-VL-8B~\citep{deshmukh2025nvidia}.

\subsection{Inference Configuration}
Unless restricted by the model provider, decoding is deterministic with temperature set to 0 and maximum generation length set to 64 tokens. We use a generic neutral system instruction, while modality-specific constraints are implemented in the user prompt. For robustness, failed requests are retried up to 3 times, the request timeout is 120 seconds, and report inputs are truncated to at most 3000 characters when necessary.

We report one evaluation run per model and input configuration under the decoding settings described above. A single run does not measure run-to-run variability; temperature-zero decoding does not by itself guarantee identical outputs across all model providers or inference environments.

\subsection{Answer Extraction and Scoring}
All tasks are evaluated in a standardized multiple-choice format. The model is required to output exactly one uppercase option letter. During post-processing, we extract the first valid option letter from the response; outputs without a valid option letter are counted as incorrect. Accuracy is computed by exact match between the extracted answer and the ground-truth option.

\appsection{Additional Experimental Results}
\label{app:more_results}

Table~\ref{tab:overall_results} in the main text reports overall accuracy under six input settings. The supplementary Table~\ref{tab:full_results} breaks down all six settings by model and task, making variation across the 13 tasks visible without repeating the overall results.

\begin{table*}[!t]
\centering
\scriptsize
\setlength{\tabcolsep}{1.5pt}
\renewcommand{\arraystretch}{0.95}
\caption{\textbf{Task-wise results under all six input settings.} Accuracy (\%) of all evaluated models on 13 HM-Bench tasks, spanning single-view (Report-only, PCA-only, RGB-only), partial multi-view (RGB+Report, RGB+PCA) and full VSR$^{2}$ exposure. The table continues in the next panel; the bottom block reports the unweighted average over 17 models for each task and input setting. These task-wise averages differ from the question-weighted overall accuracies in Table~\ref{tab:overall_results}.}
\label{tab:full_results}
\begin{tabular}{@{}llccccccccccccc@{}}
\toprule
Model & Input & \multicolumn{2}{c}{FR} & \multicolumn{1}{c}{TQ} & \multicolumn{3}{c}{SL} & \multicolumn{2}{c}{CI} & \multicolumn{2}{c}{SE} & \multicolumn{3}{c}{CD} \\
\cmidrule(lr){3-4} \cmidrule(lr){5-5} \cmidrule(lr){6-8} \cmidrule(lr){9-10} \cmidrule(lr){11-12} \cmidrule(lr){13-15}
 & & SFR & LCC & PD & Count & OLR & RD & SAD & SU & VHS & EPS & BCI & CAL & CSA \\
\midrule
\multirow{6}{*}{Claude-Sonnet-4.6}
  & Report & 41.54 & 23.39 & 39.80 & 34.57 & 30.38 & 47.30 & 39.58 & 39.62 & 48.58 & 45.37 & 30.36 & 21.60 & 33.33 \\
  & PCA & 58.20 & 42.17 & 50.07 & 26.73 & 29.02 & 43.77 & 32.05 & 42.65 & 49.26 & 56.48 & 38.08 & 14.80 & 30.00 \\
  & RGB & 48.67 & 29.59 & 56.26 & 24.70 & 31.64 & 42.43 & 32.36 & 41.83 & 46.32 & 53.01 & 36.35 & 12.41 & 30.05 \\
  & RGB+Rep & 45.80 & 30.10 & 40.99 & 31.37 & 34.34 & 52.48 & 51.00 & 43.93 & 53.79 & 52.08 & 34.36 & 25.16 & 20.04 \\
  & RGB+PCA & 55.94 & 39.09 & 55.55 & 32.17 & 35.24 & 41.05 & 33.31 & 45.26 & 63.03 & 59.23 & 45.17 & 19.93 & 35.47 \\
  & VSR² & 51.20 & 40.63 & 50.19 & 33.71 & 38.27 & 44.04 & 40.57 & 49.70 & 65.04 & 59.59 & 43.66 & 21.06 & 34.43 \\
\cmidrule(lr){1-15}
\multirow{6}{*}{GPT-5.4-Mini}
  & Report & 39.60 & 19.77 & 32.96 & 39.05 & 27.24 & 43.77 & 60.27 & 37.32 & 47.85 & 44.52 & 31.96 & 20.41 & 29.17 \\
  & PCA & 50.60 & 35.48 & 46.06 & 32.05 & 28.55 & 42.00 & 64.18 & 47.30 & 41.79 & 51.31 & 35.55 & 12.41 & 35.00 \\
  & RGB & 42.61 & 31.61 & 49.93 & 27.22 & 32.22 & 39.70 & 45.90 & 38.36 & 47.23 & 50.31 & 32.62 & 13.61 & 29.17 \\
  & RGB+Rep & 43.84 & 29.38 & 45.00 & 34.68 & 32.80 & 41.43 & 63.78 & 40.73 & 55.45 & 50.53 & 38.46 & 24.42 & 27.00 \\
  & RGB+PCA & 49.99 & 39.85 & 51.25 & 38.02 & 35.38 & 36.49 & 64.79 & 48.36 & 53.32 & 58.30 & 40.16 & 16.76 & 35.45 \\
  & VSR² & 45.79 & 33.50 & 50.43 & 39.11 & 35.15 & 40.12 & 61.88 & 46.84 & 60.04 & 55.37 & 42.87 & 25.69 & 38.78 \\
\cmidrule(lr){1-15}
\multirow{6}{*}{Grok-4}
  & Report & 33.33 & 15.41 & 26.73 & 21.90 & 23.73 & 39.86 & 54.40 & 22.00 & 2.55 & 44.75 & 22.37 & 10.71 & 24.17 \\
  & PCA & 48.22 & 23.02 & 27.91 & 27.57 & 22.00 & 38.20 & 51.32 & 39.84 & 34.71 & 50.46 & 24.63 & 12.84 & 31.00 \\
  & RGB & 38.79 & 21.79 & 36.43 & 24.70 & 28.97 & 22.31 & 44.39 & 39.76 & 48.53 & 49.61 & 37.68 & 14.97 & 19.17 \\
  & RGB+Rep & 45.30 & 28.00 & 44.12 & 35.78 & 31.46 & 39.93 & 64.72 & 44.44 & 35.03 & 50.82 & 35.86 & 20.98 & 26.63 \\
  & RGB+PCA & 55.13 & 35.49 & 54.80 & 33.97 & 35.66 & 50.36 & 44.55 & 45.39 & 55.84 & 59.59 & 45.76 & 14.73 & 37.39 \\
  & VSR² & 44.72 & 25.79 & 41.86 & 31.97 & 33.57 & 36.58 & 58.22 & 40.91 & 48.16 & 53.83 & 38.69 & 23.55 & 35.64 \\
\cmidrule(lr){1-15}
\multirow{6}{*}{InternVL2-8B}
  & Report & 37.76 & 20.87 & 28.05 & 39.19 & 27.40 & 29.05 & 64.03 & 38.65 & 41.85 & 62.73 & 33.95 & 14.80 & 40.00 \\
  & PCA & 41.80 & 29.47 & 43.37 & 38.14 & 27.40 & 31.03 & 34.24 & 43.39 & 39.18 & 65.28 & 37.55 & 11.05 & 41.67 \\
  & RGB & 39.48 & 25.11 & 52.97 & 31.14 & 29.54 & 28.73 & 37.10 & 41.39 & 44.51 & 65.90 & 37.02 & 14.80 & 40.83 \\
  & RGB+Rep & 38.16 & 23.35 & 34.03 & 41.38 & 27.84 & 28.91 & 64.28 & 40.08 & 43.28 & 65.30 & 33.84 & 13.97 & 38.35 \\
  & RGB+PCA & 40.26 & 27.96 & 48.91 & 36.26 & 30.41 & 29.70 & 30.32 & 44.29 & 44.39 & 69.80 & 41.21 & 14.74 & 40.30 \\
  & VSR² & 39.39 & 25.73 & 39.39 & 40.16 & 26.90 & 28.75 & 61.68 & 42.33 & 44.01 & 66.41 & 39.94 & 13.74 & 40.13 \\
\cmidrule(lr){1-15}
\multirow{6}{*}{InternVL2.5-4B}
  & Report & 33.93 & 22.22 & 32.66 & 29.11 & 29.60 & 26.27 & 27.09 & 30.08 & 42.24 & 59.34 & 33.95 & 13.27 & 30.83 \\
  & PCA & 33.33 & 20.93 & 30.71 & 25.54 & 31.59 & 24.56 & 59.67 & 33.92 & 54.98 & 50.08 & 37.15 & 14.63 & 35.00 \\
  & RGB & 39.13 & 25.72 & 34.38 & 27.01 & 30.96 & 30.55 & 62.00 & 39.39 & 41.17 & 55.79 & 37.28 & 14.12 & 30.00 \\
  & RGB+Rep & 38.14 & 23.14 & 27.27 & 37.86 & 26.30 & 27.34 & 60.35 & 32.52 & 45.19 & 52.24 & 28.89 & 14.97 & 35.83 \\
  & RGB+PCA & 42.54 & 30.51 & 38.54 & 27.07 & 29.26 & 29.38 & 29.59 & 38.48 & 49.05 & 46.97 & 33.69 & 12.99 & 34.10 \\
  & VSR² & 39.22 & 24.28 & 30.16 & 44.37 & 26.19 & 25.65 & 62.62 & 31.42 & 43.62 & 52.88 & 32.01 & 14.76 & 39.79 \\
\cmidrule(lr){1-15}
\multirow{6}{*}{InternVL3-9B}
  & Report & 36.90 & 21.24 & 25.93 & 42.62 & 28.86 & 30.50 & 61.63 & 42.05 & 53.74 & 53.32 & 36.22 & 22.96 & 40.83 \\
  & PCA & 40.98 & 28.55 & 40.67 & 41.43 & 30.17 & 31.19 & 64.94 & 41.09 & 63.76 & 59.88 & 42.48 & 14.97 & 35.83 \\
  & RGB & 42.57 & 29.65 & 44.65 & 37.72 & 31.33 & 29.91 & 61.02 & 41.09 & 57.13 & 57.79 & 44.74 & 12.76 & 33.33 \\
  & RGB+Rep & 39.20 & 28.59 & 35.80 & 38.68 & 28.63 & 31.55 & 62.13 & 41.44 & 56.15 & 59.16 & 38.60 & 20.05 & 39.98 \\
  & RGB+PCA & 44.36 & 34.06 & 45.40 & 44.87 & 30.08 & 32.04 & 62.55 & 42.31 & 57.19 & 63.66 & 43.33 & 14.80 & 42.52 \\
  & VSR² & 41.02 & 29.85 & 36.48 & 39.13 & 28.73 & 30.68 & 61.78 & 41.62 & 58.46 & 58.18 & 40.69 & 18.62 & 40.91 \\
\cmidrule(lr){1-15}
\multirow{6}{*}{InternVL3-14B}
  & Report & 37.24 & 19.77 & 25.69 & 37.44 & 29.07 & 46.50 & 63.81 & 37.32 & 51.53 & 48.77 & 35.15 & 18.54 & 20.83 \\
  & PCA & 44.29 & 29.77 & 36.87 & 26.24 & 25.14 & 45.80 & 68.17 & 38.73 & 42.13 & 49.85 & 42.34 & 10.71 & 39.17 \\
  & RGB & 47.64 & 32.11 & 46.63 & 30.16 & 30.91 & 42.43 & 66.22 & 40.87 & 46.94 & 51.70 & 43.01 & 13.27 & 41.67 \\
  & RGB+Rep & 39.20 & 27.28 & 38.17 & 41.33 & 31.61 & 49.92 & 67.20 & 41.13 & 49.91 & 49.74 & 38.72 & 19.98 & 27.27 \\
  & RGB+PCA & 50.77 & 37.49 & 43.92 & 27.78 & 29.20 & 38.62 & 67.02 & 42.35 & 58.80 & 56.96 & 46.87 & 13.57 & 40.50 \\
  & VSR² & 43.81 & 29.78 & 34.88 & 36.24 & 30.71 & 49.32 & 65.67 & 41.39 & 46.92 & 52.65 & 40.39 & 18.63 & 30.28 \\
\cmidrule(lr){1-15}
\multirow{6}{*}{InternVL3.5-14B}
  & Report & 39.91 & 21.67 & 28.48 & 48.99 & 28.50 & 49.28 & 53.42 & 39.39 & 55.44 & 52.08 & 40.08 & 19.22 & 31.67 \\
  & PCA & 43.26 & 32.72 & 49.39 & 32.26 & 25.72 & 50.08 & 64.11 & 44.27 & 47.11 & 57.72 & 38.75 & 12.59 & 33.33 \\
  & RGB & 43.64 & 31.98 & 48.92 & 42.97 & 29.23 & 47.83 & 64.18 & 42.94 & 57.64 & 58.33 & 37.95 & 15.99 & 30.00 \\
  & RGB+Rep & 43.62 & 28.10 & 43.10 & 49.81 & 30.96 & 51.97 & 60.21 & 42.84 & 58.09 & 53.68 & 42.00 & 20.40 & 34.02 \\
  & RGB+PCA & 45.95 & 32.45 & 54.57 & 43.95 & 31.38 & 43.51 & 65.74 & 46.88 & 51.18 & 60.95 & 43.85 & 14.21 & 34.81 \\
  & VSR² & 44.33 & 29.53 & 50.25 & 48.88 & 30.57 & 46.29 & 68.93 & 44.53 & 61.09 & 56.07 & 42.04 & 21.53 & 42.12 \\
\cmidrule(lr){1-15}
\multirow{6}{*}{Kimi-VL-A3B}
  & Report & 36.90 & 22.96 & 31.48 & 33.52 & 27.40 & 40.02 & 63.36 & 36.14 & 52.15 & 52.47 & 37.02 & 14.80 & 41.67 \\
  & PCA & 35.88 & 27.75 & 52.53 & 27.36 & 25.41 & 37.24 & 32.58 & 39.47 & 40.20 & 55.02 & 46.21 & 15.31 & 40.00 \\
  & RGB & 41.29 & 29.71 & 54.78 & 27.57 & 30.64 & 47.57 & 29.35 & 39.99 & 42.13 & 54.24 & 38.48 & 14.29 & 41.67 \\
  & RGB+Rep & 44.16 & 28.70 & 50.81 & 29.63 & 31.39 & 44.60 & 33.36 & 39.68 & 56.36 & 54.22 & 35.88 & 17.88 & 42.06 \\
  & RGB+PCA & 44.65 & 31.83 & 60.06 & 30.05 & 32.36 & 49.74 & 35.13 & 41.43 & 45.47 & 56.72 & 40.77 & 16.84 & 42.55 \\
  & VSR² & 47.25 & 31.70 & 57.41 & 30.79 & 31.68 & 47.66 & 32.31 & 42.35 & 43.20 & 57.06 & 39.81 & 20.01 & 43.49 \\
\bottomrule
\end{tabular}
\end{table*}

\begin{table*}[!t]
\ContinuedFloat
\centering
\scriptsize
\setlength{\tabcolsep}{1.5pt}
\renewcommand{\arraystretch}{0.95}
\caption[]{\textbf{Task-wise results under all six input settings (continued).}}
\begin{tabular}{@{}llccccccccccccc@{}}
\toprule
Model & Input & \multicolumn{2}{c}{FR} & \multicolumn{1}{c}{TQ} & \multicolumn{3}{c}{SL} & \multicolumn{2}{c}{CI} & \multicolumn{2}{c}{SE} & \multicolumn{3}{c}{CD} \\
\cmidrule(lr){3-4} \cmidrule(lr){5-5} \cmidrule(lr){6-8} \cmidrule(lr){9-10} \cmidrule(lr){11-12} \cmidrule(lr){13-15}
 & & SFR & LCC & PD & Count & OLR & RD & SAD & SU & VHS & EPS & BCI & CAL & CSA \\
\midrule
\multirow{6}{*}{Qwen2.5-VL-7B}
  & Report & 31.31 & 14.30 & 26.20 & 28.83 & 23.10 & 26.70 & 57.71 & 28.82 & 52.72 & 44.75 & 16.25 & 13.27 & 33.33 \\
  & PCA & 31.74 & 18.35 & 24.41 & 26.82 & 19.75 & 34.56 & 56.28 & 36.88 & 21.57 & 43.52 & 25.70 & 11.90 & 42.50 \\
  & RGB & 39.35 & 25.29 & 37.44 & 34.41 & 31.53 & 33.55 & 63.51 & 38.95 & 41.73 & 50.77 & 43.14 & 16.16 & 33.33 \\
  & RGB+Rep & 39.70 & 23.47 & 29.93 & 38.45 & 29.40 & 30.46 & 64.79 & 37.58 & 55.34 & 50.58 & 40.17 & 15.95 & 36.63 \\
  & RGB+PCA & 45.12 & 29.83 & 51.07 & 25.93 & 33.22 & 33.14 & 63.76 & 41.53 & 44.85 & 53.68 & 43.22 & 15.82 & 29.71 \\
  & VSR² & 40.68 & 26.76 & 32.14 & 38.17 & 29.12 & 33.53 & 64.46 & 37.32 & 50.67 & 59.91 & 43.34 & 16.38 & 39.90 \\
\cmidrule(lr){1-15}
\multirow{6}{*}{Qwen3-VL-4B}
  & Report & 36.25 & 20.20 & 25.19 & 44.37 & 27.03 & 24.34 & 60.27 & 37.40 & 56.29 & 59.72 & 36.62 & 13.95 & 35.00 \\
  & PCA & 51.89 & 37.63 & 36.06 & 32.33 & 27.50 & 28.04 & 66.29 & 41.98 & 43.60 & 66.05 & 44.87 & 10.20 & 37.50 \\
  & RGB & 46.43 & 28.73 & 41.75 & 35.34 & 31.59 & 31.03 & 65.31 & 43.16 & 55.83 & 59.34 & 41.01 & 13.78 & 40.83 \\
  & RGB+Rep & 39.31 & 24.33 & 31.94 & 45.67 & 31.51 & 31.90 & 65.63 & 40.08 & 57.58 & 56.41 & 33.95 & 19.44 & 33.72 \\
  & RGB+PCA & 52.65 & 32.07 & 50.03 & 37.84 & 36.26 & 36.93 & 45.66 & 46.93 & 50.63 & 60.74 & 47.09 & 15.25 & 46.55 \\
  & VSR² & 42.03 & 28.07 & 36.27 & 46.50 & 34.17 & 36.02 & 70.02 & 41.26 & 57.32 & 67.41 & 40.45 & 17.87 & 44.57 \\
\cmidrule(lr){1-15}
\multirow{6}{*}{LLaVA-NeXT-8B}
  & Report & 40.08 & 20.20 & 31.45 & 37.72 & 26.30 & 29.37 & 61.55 & 34.44 & 54.19 & 57.72 & 26.23 & 13.10 & 41.67 \\
  & PCA & 39.82 & 24.43 & 41.52 & 38.35 & 29.65 & 32.58 & 64.48 & 39.76 & 39.58 & 58.56 & 37.55 & 12.76 & 46.67 \\
  & RGB & 40.94 & 24.55 & 38.22 & 41.85 & 30.38 & 36.12 & 63.28 & 39.17 & 53.34 & 59.95 & 32.89 & 12.76 & 47.50 \\
  & RGB+Rep & 40.60 & 22.98 & 34.38 & 41.85 & 29.32 & 33.05 & 61.16 & 35.91 & 57.39 & 58.55 & 29.30 & 16.81 & 45.14 \\
  & RGB+PCA & 43.11 & 25.36 & 60.03 & 35.53 & 31.61 & 33.68 & 50.68 & 39.99 & 46.67 & 60.94 & 32.18 & 15.66 & 43.89 \\
  & VSR² & 38.32 & 23.01 & 37.46 & 42.32 & 29.58 & 32.49 & 62.43 & 35.03 & 56.64 & 57.97 & 32.06 & 15.66 & 47.70 \\
\cmidrule(lr){1-15}
\multirow{6}{*}{LLaVA-v1.5-7B}
  & Report & 39.69 & 44.51 & 64.11 & 37.72 & 23.99 & 27.07 & 21.97 & 27.94 & 30.63 & 16.67 & 30.23 & 10.03 & 30.00 \\
  & PCA & 37.46 & 43.09 & 61.65 & 36.46 & 23.57 & 26.48 & 20.99 & 27.05 & 29.11 & 17.67 & 29.29 & 9.69 & 28.33 \\
  & RGB & 38.85 & 44.34 & 64.35 & 37.94 & 22.01 & 26.73 & 21.79 & 28.44 & 30.67 & 17.69 & 31.08 & 10.02 & 29.13 \\
  & RGB+Rep & 43.20 & 48.53 & 67.96 & 41.74 & 27.96 & 31.09 & 25.92 & 32.11 & 34.71 & 20.30 & 34.78 & 14.05 & 34.02 \\
  & RGB+PCA & 41.56 & 44.45 & 53.69 & 31.96 & 29.26 & 30.26 & 34.43 & 38.33 & 57.83 & 45.33 & 30.42 & 10.73 & 33.52 \\
  & VSR² & 42.23 & 38.05 & 47.63 & 37.53 & 30.69 & 32.64 & 66.63 & 37.02 & 56.24 & 53.62 & 32.40 & 12.99 & 40.10 \\
\cmidrule(lr){1-15}
\multirow{6}{*}{Geochat-7B}
  & Report & 37.33 & 43.28 & 62.02 & 36.88 & 23.36 & 25.41 & 20.32 & 27.86 & 30.69 & 16.36 & 30.49 & 10.03 & 28.33 \\
  & PCA & 39.39 & 44.51 & 64.07 & 37.72 & 23.94 & 27.07 & 21.90 & 28.01 & 31.65 & 16.74 & 30.76 & 10.37 & 30.00 \\
  & RGB & 36.60 & 28.12 & 66.26 & 41.85 & 28.08 & 30.66 & 26.41 & 33.41 & 33.92 & 38.04 & 23.83 & 11.56 & 27.50 \\
  & RGB+Rep & 46.56 & 48.08 & 68.86 & 41.97 & 29.34 & 32.45 & 26.82 & 32.48 & 34.83 & 24.54 & 33.68 & 14.28 & 34.25 \\
  & RGB+PCA & 54.79 & 33.20 & 60.85 & 42.29 & 31.06 & 29.56 & 65.84 & 36.43 & 48.26 & 40.43 & 31.34 & 19.57 & 36.57 \\
  & VSR² & 49.31 & 46.10 & 70.28 & 45.99 & 31.87 & 34.28 & 30.30 & 33.76 & 40.83 & 51.80 & 32.52 & 14.11 & 33.59 \\
\cmidrule(lr){1-15}
\multirow{6}{*}{GeoLLaVA-8K}
  & Report & 41.71 & 34.99 & 41.25 & 43.11 & 23.73 & 32.26 & 45.97 & 35.70 & 46.38 & 59.72 & 33.42 & 12.24 & 39.17 \\
  & PCA & 42.44 & 32.78 & 40.17 & 42.48 & 23.31 & 30.02 & 40.56 & 36.81 & 47.68 & 56.48 & 34.09 & 12.07 & 35.00 \\
  & RGB & 47.77 & 45.30 & 38.89 & 41.36 & 23.99 & 31.19 & 54.18 & 37.77 & 43.43 & 46.22 & 34.62 & 11.22 & 28.33 \\
  & RGB+Rep & 50.20 & 47.43 & 35.56 & 46.94 & 24.50 & 36.07 & 55.48 & 38.18 & 44.43 & 53.76 & 36.87 & 11.82 & 33.12 \\
  & RGB+PCA & 51.89 & 47.77 & 39.83 & 47.05 & 26.29 & 35.26 & 56.76 & 41.37 & 44.51 & 61.01 & 36.33 & 12.01 & 30.31 \\
  & VSR² & 52.76 & 48.91 & 38.87 & 50.78 & 26.90 & 38.24 & 57.30 & 41.91 & 47.20 & 64.57 & 36.29 & 11.93 & 29.57 \\
\cmidrule(lr){1-15}
\multirow{6}{*}{GLM-4.6V-Flash-9B}
  & Report & 38.62 & 20.38 & 30.71 & 39.82 & 28.55 & 44.25 & 60.72 & 38.58 & 51.93 & 46.22 & 30.09 & 18.37 & 25.83 \\
  & PCA & 44.93 & 31.18 & 40.81 & 34.57 & 28.92 & 47.19 & 62.75 & 44.49 & 42.98 & 62.42 & 44.61 & 9.52 & 36.67 \\
  & RGB & 45.40 & 34.99 & 46.77 & 42.62 & 31.43 & 49.09 & 61.78 & 44.42 & 45.83 & 59.57 & 42.34 & 15.14 & 37.50 \\
  & RGB+Rep & 43.56 & 25.89 & 42.86 & 44.03 & 33.65 & 52.19 & 64.92 & 42.97 & 58.61 & 53.40 & 37.74 & 20.86 & 24.82 \\
  & RGB+PCA & 47.15 & 37.43 & 46.01 & 36.39 & 31.47 & 43.67 & 60.99 & 47.74 & 48.68 & 63.59 & 48.94 & 12.28 & 38.07 \\
  & VSR² & 50.53 & 40.13 & 49.33 & 34.64 & 32.25 & 41.79 & 65.18 & 47.72 & 47.35 & 59.92 & 47.23 & 12.36 & 38.49 \\
\cmidrule(lr){1-15}
\multirow{6}{*}{Nemotron-Nano-VL-8B}
  & Report & 38.83 & 21.85 & 38.55 & 31.70 & 29.65 & 23.70 & 59.44 & 39.84 & 48.13 & 52.78 & 36.62 & 18.20 & 35.83 \\
  & PCA & 39.82 & 25.66 & 42.69 & 26.87 & 30.02 & 28.30 & 65.69 & 40.87 & 43.71 & 56.10 & 44.61 & 13.95 & 30.00 \\
  & RGB & 40.12 & 24.55 & 36.63 & 30.02 & 29.54 & 31.84 & 60.72 & 39.10 & 44.90 & 56.25 & 35.29 & 14.80 & 33.33 \\
  & RGB+Rep & 41.09 & 26.37 & 42.66 & 33.81 & 30.52 & 27.52 & 67.68 & 41.80 & 54.84 & 52.33 & 36.80 & 17.04 & 40.71 \\
  & RGB+PCA & 43.14 & 27.66 & 48.55 & 27.64 & 32.67 & 29.75 & 30.34 & 41.73 & 45.25 & 55.91 & 42.57 & 16.62 & 33.49 \\
  & VSR² & 42.82 & 27.45 & 44.55 & 30.90 & 30.77 & 28.95 & 66.35 & 41.39 & 49.96 & 55.13 & 39.56 & 15.44 & 41.64 \\
\midrule
\multirow{6}{*}{Average}
  & Report & 37.70 & 23.94 & 34.78 & 36.86 & 26.93 & 34.45 & 51.50 & 34.89 & 45.11 & 48.08 & 31.82 & 15.62 & 33.04 \\
  & PCA & 42.59 & 31.03 & 42.88 & 32.52 & 26.57 & 35.18 & 51.19 & 39.21 & 41.94 & 51.39 & 37.31 & 12.34 & 35.75 \\
  & RGB & 42.31 & 30.18 & 46.78 & 34.03 & 29.65 & 35.39 & 50.56 & 39.41 & 45.96 & 52.03 & 37.02 & 13.63 & 33.73 \\
  & RGB+Rep & 42.45 & 30.22 & 41.97 & 39.70 & 30.09 & 37.82 & 56.44 & 39.29 & 50.06 & 50.45 & 35.88 & 18.12 & 33.74 \\
  & RGB+PCA & 47.59 & 34.50 & 50.77 & 35.22 & 31.81 & 36.66 & 49.50 & 42.87 & 50.88 & 57.28 & 40.76 & 15.09 & 37.36 \\
  & VSR² & 44.44 & 32.31 & 43.98 & 39.48 & 31.01 & 36.88 & 58.61 & 40.97 & 51.57 & 57.79 & 39.06 & 17.31 & 38.89 \\
\bottomrule
\end{tabular}
\end{table*}

\begin{table*}[!tp]
\centering
\caption{\textbf{Controlled comparisons on three representative MLLMs.} Accuracy is in percent; differences are in percentage points. The spectral-range control compares RGB plus visible-spectrum PCA with RGB plus full-spectrum PCA, with no report in either setting. The question-only control removes all image and report inputs while retaining the original prompts and model settings. $\Delta_{\mathrm{spec}}$ is full-spectrum minus visible-spectrum PCA; $\Delta_{\mathrm{input}}$ is full VSR$^2$ minus question-only accuracy.}
\label{tab:spectral_range_question_only}
\small
\setlength{\tabcolsep}{5pt}
\begin{tabular}{lrrrrrr}
\toprule
& \multicolumn{3}{c}{Spectral-range control} & \multicolumn{3}{c}{Question-only control} \\
\cmidrule(lr){2-4}\cmidrule(lr){5-7}
Model & RGB+Vis-PCA & RGB+Full-PCA & $\Delta_{\mathrm{spec}}$ & Question-only & VSR$^2$ & $\Delta_{\mathrm{input}}$ \\
\midrule
InternVL2-8B     & 37.49 & 38.35 & +0.86 & 27.29 & 39.12 & 11.83 \\
Qwen2.5-VL-7B    & 38.83 & 39.30 & +0.47 & 26.61 & 39.41 & 12.80 \\
Qwen3-VL-4B      & 42.46 & 42.97 & +0.51 & 27.15 & 43.23 & 16.08 \\
\bottomrule
\end{tabular}
\end{table*}

\paragraph{Overall performance and the effect of fusion.}
As Table~\ref{tab:overall_results} shows, multi-view input generally improves performance over single-view input. Averaged over all evaluated models, the full VSR$^{2}$ setting achieves the highest mean accuracy (40.95\%), outperforming Report-only (35.84\%), PCA-only (38.10\%), and RGB-only (38.90\%). This is a system-level result under common questions and scoring; Table~\ref{tab:r2_control_deltas} gives more narrowly matched evidence for the spectral-range and report-content contributions on three models.

At the same time, the gain from fusion is not uniform across modalities. The average result of \emph{w/o Report} (40.79\%) is close to the full VSR$^{2}$ result (40.95\%), whereas \emph{w/o PCA} reaches 38.94\%. This suggests that, among the two auxiliary views, the PCA view contributes more consistently than the report view when combined with RGB. In other words, the largest fusion benefit appears to come from adding PCA-based spectral variation cues to the RGB image, while the report provides a smaller but still occasionally useful complementary signal.

\paragraph{Relative utility of the three single views.}
Among the three single-view settings, RGB-only achieves the highest mean model accuracy among single views (38.90\%), followed by PCA-only (38.10\%), with Report-only trailing behind (35.84\%). This ranking is consistent with the design intuition behind VSR$^{2}$: pseudo-RGB images are the most naturally aligned with the visual pretraining distribution of existing MLLMs, PCA provides additional spectral discrimination beyond visible appearance, and text reports offer more abstract evidence that may be harder for general-purpose MLLMs to exploit reliably in a zero-shot setting.

This observation also supports the evidence-priority rule used in our fusion prompt. Since RGB is empirically the strongest standalone source, it is reasonable to treat it as the primary evidence and use PCA and report views as complementary references rather than equally weighted inputs.

\paragraph{Best-performing models and open-source competitiveness.}
The strongest overall model in the full VSR$^{2}$ setting is InternVL3.5-14B, which reaches 45.09\% accuracy. Notably, this open-source model also ranks first in several other settings, including Report-only, PCA-only, \emph{w/o PCA}, and \emph{w/o Report}, indicating strong robustness across heterogeneous inputs. Among closed-source models, GPT-5.4-Mini achieves the best overall fusion result (44.28\%), while Claude-Sonnet-4-6 performs competitively and Grok-4 shows a less consistent pattern across settings.

The performance across all models suggest that strong open-source MLLMs can already match or exceed the performance of proprietary systems on hyperspectral reasoning under a unified zero-shot protocol, especially when the view design and prompting strategy are carefully controlled.

\begin{figure*}[t]
    \centering
    \includegraphics[width=\textwidth,trim=0 145 0 155,clip]{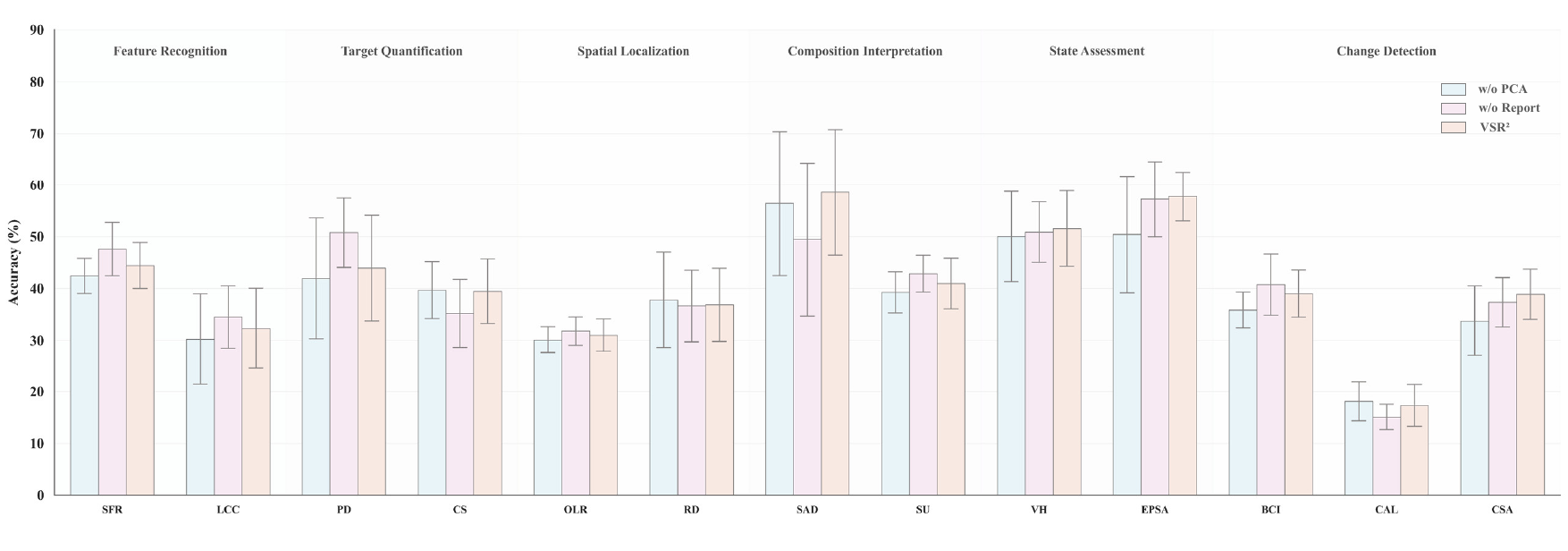}
    \caption{\textbf{Task-wise performance of partial and full multi-view settings.} Bars show the unweighted mean accuracy over 17 models; error bars indicate one standard deviation across models. The settings are RGB+Report (w/o PCA), RGB+PCA (w/o Report), and full VSR$^2$. CS, VH, and EPSA correspond to Count, VHS, and EPS in Table~\ref{tab:full_results}.}
    \label{fig:taskwise_vsr_fusion}
\end{figure*}

\paragraph{Task-wise performance of multi-view settings.}
Figure~\ref{fig:taskwise_vsr_fusion} complements the single-view comparison in Fig.~\ref{fig:result}(a) by showing full VSR$^2$ alongside its two partial-fusion variants. Averaged across models, full VSR$^2$ exceeds RGB+PCA on Count and SAD, while it is lower on PD, illustrating that the impact of the report view varies by task. The error bars describe variability across models, not confidence intervals for paired gains; detailed per-model results are given in Table~\ref{tab:full_results}.

\paragraph{Task-level differences across modalities.}
The task-wise results in Table~\ref{tab:full_results} reveal substantial heterogeneity in modality preference. RGB-only performs best on average for several tasks, including PD, OLR, RD, SU, VHS, and EPS. These tasks likely depend more strongly on spatial layout, morphology, texture, or pseudo-color appearance, all of which are directly accessible in RGB visualizations and better aligned with the native capabilities of current vision-language models.

PCA-only is the strongest single view on several other tasks, such as SFR, LCC, BCI, and CSA, and stays competitive with RGB on SU and RD. This indicates that PCA views can expose discriminative spectral variance that is not always obvious in RGB renderings. In tasks that require subtle material separation or latent spectral distinction, PCA can therefore provide a useful complementary perspective even when it is not the strongest standalone view overall.

Report-only, while weakest on average, is not uniformly uninformative. It achieves the best mean performance on Count, SAD, and CAL among the three single-view settings, suggesting that text-based summaries may help in tasks that benefit from explicit quantitative or categorical evidence rather than purely visual inspection. This is consistent with the role of the report view as a compact textual abstraction of spectral--spatial information.

\paragraph{Spectral-range and question-only controls.}
We conduct two additional controlled evaluations on three representative MLLMs (Table~\ref{tab:spectral_range_question_only}). For the spectral-range comparison, both settings include the same RGB view and exclude the report; only the bands supplied to PCA change. Full-spectrum PCA improves overall accuracy over visible-spectrum PCA by 0.86, 0.47, and 0.51 percentage points for InternVL2-8B, Qwen2.5-VL-7B, and Qwen3-VL-4B, respectively. These positive but modest differences provide evidence that non-visible bands can add useful information in the tested models; they do not establish the size or consistency of this contribution across all 17 models or every task.

For the question-only control, we retain the original prompts and model settings while removing all image and report inputs. Accuracy falls to 27.29\%, 26.61\%, and 27.15\%, close to the 25.13\% random baseline and 11.83--16.08 percentage points below the corresponding full VSR$^2$ results. This indicates that the supplied views contribute substantially to performance in these three models, beyond what is available from the question text alone.

\paragraph{Alternative training-free spectral representations.}
To complement the component ablations, we evaluate selected spectral bands and spectral curves on InternVL2-8B, Qwen2.5-VL-7B, and Qwen3-VL-4B using the same benchmark questions and answer-scoring protocol. Each alternative is evaluated as a standalone visual input and as a replacement for the PCA view while retaining RGB and the report. Table~\ref{tab:external_hsi_baselines} reports question-weighted overall accuracy for each model and the equally weighted mean across the three models. The compared inputs differ in visual representation and, for standalone versus fusion settings, in the number and type of views; these results therefore characterize system-level performance rather than isolating spectral content alone.

The full VSR$^2$ mean exceeds the selected-band and spectral-curve fusion alternatives by 1.33 and 1.52 percentage points, respectively. Rankings vary across models: selected-band fusion slightly exceeds VSR$^2$ for InternVL2-8B, and spectral-curve fusion exceeds it for Qwen2.5-VL-7B. These results support the utility of the multi-view interface without establishing PCA as universally optimal.

\begin{table*}[t]
\centering
\small
\setlength{\tabcolsep}{7pt}
\caption{\textbf{Alternative training-free spectral representations.} Question-weighted overall accuracy (\%) on three models; Mean is the unweighted mean of their displayed accuracies. Bands and curves are evaluated alone and as replacements for PCA with RGB and report retained. Question-only and VSR$^2$ are included as reference settings.}
\label{tab:external_hsi_baselines}
\begin{tabular}{lrrrr}
\toprule
Input & InternVL2-8B & Qwen2.5-VL-7B & Qwen3-VL-4B & Mean \\
\midrule
Question-only & 27.29 & 26.61 & 27.15 & 27.02 \\
Selected bands only & 37.91 & 30.17 & 38.43 & 35.50 \\
Spectral curves only & 38.10 & 36.97 & 37.04 & 37.37 \\
RGB + report + selected bands & 39.20 & 38.52 & 40.05 & 39.26 \\
RGB + report + spectral curves & 39.06 & 39.63 & 38.51 & 39.07 \\
VSR$^2$ & 39.12 & 39.41 & 43.23 & 40.59 \\
\bottomrule
\end{tabular}
\end{table*}

\paragraph{Task-wise controlled contrasts.}
\label{app:controlled_contributions}
Table~\ref{tab:r2_control_deltas} reports three contrasts for the same three models across all 13 tasks. The spectral-range contrast replaces visible-band PCA with full-spectrum PCA, keeping RGB, the number of images, and the report-free prompt structure fixed. Full-spectrum PCA improves LCC, PD, OLR, SU, VHS, and BCI on all three models, but reduces EPS on all three; its contribution is therefore task dependent. The report-content contrast retains RGB, full-spectrum PCA, and the report field, replacing only an empty report with the generated report. Its overall gains are +2.93, +1.41, and +0.15 points, with consistent gains on Count, SAD, VHS, CAL, OLR, and CSA and consistent declines on PD. This control isolates the effect of report content within a fixed field structure, although the filled report necessarily uses more text tokens than an empty field.

The full VSR$^2$ versus RGB-only contrast shows the effect of the complete system, including additional views and modality-specific prompt clauses; it does not by itself attribute the gain to non-visible bands. Count improves and PD declines on all three models, illustrating why the overall score should be read alongside the task-level results. Original task scores for the six standard settings are already listed in Table~\ref{tab:full_results}, and the overall visible-PCA and question-only controls are in Table~\ref{tab:spectral_range_question_only}; Table~\ref{tab:r2_control_deltas} therefore gives differences without repeating those scores. Paired scene-level uncertainty for these contrasts cannot be recovered from the available task-aggregated control results, so the scene intervals in Table~\ref{tab:r2_task_scene} describe full VSR$^2$ accuracy rather than uncertainty in these differences.

\begin{table*}[!t]
\centering
\scriptsize
\setlength{\tabcolsep}{2pt}
\caption{\textbf{Task-wise controlled contrasts on three MLLMs.} Differences are in percentage points; positive values favor the first setting. Full--Vis PCA compares RGB+full-spectrum PCA with RGB+visible-spectrum PCA. Report--Empty compares full VSR$^2$ with RGB+PCA+an empty report field. VSR$^2$--RGB is the full-system comparison. All three contrasts use the same questions and scoring; the first two keep the visual input count fixed.}
\label{tab:r2_control_deltas}
\resizebox{\textwidth}{!}{%
\begin{tabular}{llrrrrrrrrrrrrrr}
\toprule
Contrast & Model & SFR & LCC & PD & Count & OLR & RD & SAD & SU & VHS & EPS & BCI & CAL & CSA & Overall \\
\midrule
Full--Vis PCA & InternVL2-8B & +0.01 & +0.58 & +5.51 & +1.13 & +2.28 & +1.66 & +4.06 & +3.27 & +1.47 & -0.96 & +1.00 & -0.23 & +1.97 & +0.86 \\
 & Qwen2.5-VL-7B & +4.93 & +3.88 & +9.78 & +1.39 & +4.89 & -2.62 & +1.93 & +6.31 & +2.08 & -14.07 & +3.61 & +1.51 & -11.79 & +0.47 \\
 & Qwen3-VL-4B & +6.17 & +2.97 & +6.97 & -1.49 & +6.03 & -3.95 & -18.90 & +6.06 & +4.03 & -2.99 & +5.81 & +5.05 & +13.22 & +0.51 \\
\addlinespace[3pt]
Report--Empty & InternVL2-8B & -3.82 & -4.87 & -4.77 & +11.06 & +3.22 & +4.17 & +33.42 & +2.74 & +5.77 & +0.76 & +3.31 & +2.70 & +4.68 & +2.93 \\
 & Qwen2.5-VL-7B & +1.27 & +0.52 & -10.91 & +15.47 & +1.31 & -1.44 & +4.16 & +1.77 & +9.01 & -0.74 & +4.06 & +4.60 & +1.07 & +1.41 \\
 & Qwen3-VL-4B & -7.24 & -3.18 & -7.43 & +7.31 & +2.37 & -1.75 & +4.41 & -1.16 & +11.74 & +4.45 & -2.83 & +7.33 & +3.74 & +0.15 \\
\addlinespace[3pt]
VSR$^2$--RGB & InternVL2-8B & -0.09 & +0.62 & -13.58 & +9.02 & -2.64 & +0.02 & +24.58 & +0.94 & -0.50 & +0.51 & +2.92 & -1.06 & -0.70 & +2.49 \\
 & Qwen2.5-VL-7B & +1.33 & +1.47 & -5.30 & +3.76 & -2.41 & -0.02 & +0.95 & -1.63 & +8.94 & +9.14 & +0.20 & +0.22 & +6.57 & +1.22 \\
 & Qwen3-VL-4B & -4.40 & -0.66 & -5.48 & +11.16 & +2.58 & +4.99 & +4.71 & -1.90 & +1.49 & +8.07 & -0.56 & +4.09 & +3.74 & +3.15 \\
\bottomrule
\end{tabular}}
\end{table*}

\paragraph{Complementarity rather than simple redundancy.}
An important pattern emerging from the two main result tables is that the three views are complementary, but not symmetrically so. RGB provides the strongest general foundation, PCA supplies the most reliable additional signal beyond RGB, and reports offer more selective benefits that seem to matter primarily in cases where visual evidence remains ambiguous. This asymmetry explains why the full VSR$^{2}$ setting yields the best average result, while the gain over \emph{w/o Report} is relatively small: the report does not dominate performance globally, but it can still help resolve difficult cases for some models and tasks.

This finding also indicates that effective hyperspectral reasoning for general MLLMs may depend less on adding more modalities indiscriminately and more on presenting complementary evidence in a structured and prioritized manner.

\paragraph{Model-specific observations.}
Several individual models exhibit distinctive behavior. InternVL3.5-14B stands out as the most stable model across almost all settings, suggesting strong robustness to both visual and textual view transformations. GLM-4.6V-Flash-9B achieves the best RGB-only result (44.46\%), pointing to particularly strong performance when the task can be addressed through visual cues. GPT-5.4-Mini performs especially well in PCA-only and full-fusion settings, indicating good capacity for integrating alternative view representations. By contrast, Grok-4 performs noticeably worse in the full VSR$^{2}$ setting than in \emph{w/o Report}, implying that additional textual evidence may sometimes introduce distraction rather than useful clarification.

Overall, the supplementary results reinforce the central motivation of VSR$^{2}$: different hyperspectral views expose different aspects of the same scene, and their value depends both on the task and on the reasoning characteristics of the underlying MLLM.

\begin{table*}[!t]
\centering
\scriptsize
\setlength{\tabcolsep}{3pt}
\caption{\textbf{Accuracy by original scene group for 13 models under full VSR$^2$.} The 23 source datasets form 19 groups after grouping acquisitions of the same scene across years. $N$ is the number of distinct questions per group; each question has 13 model responses. Pooled accuracy weights questions within each scene. The ranked list continues in the right panel.}
\label{tab:r2_scene}
\resizebox{\textwidth}{!}{%
\begin{tabular}{lllrr@{\hspace{12pt}}lllrr}
\toprule
Scene group & Sensor & Category & $N$ & Acc. & Scene group & Sensor & Category & $N$ & Acc. \\
\midrule
Xiongan & VNIR & Agriculture & 3003 & 65.22 & SantaBarbara & AVIRIS & Urban & 820 & 30.83 \\
Salinas & AVIRIS & Agriculture & 1080 & 63.38 & Holden & CRISM & Mars & 290 & 30.77 \\
Houston & CASI-1500 & Urban & 1002 & 52.93 & HanChuan & Headwall & Agriculture & 563 & 30.52 \\
Hermiston & EO-1 & Agriculture & 384 & 43.29 & LongKou & Headwall & Urban & 552 & 29.86 \\
Washington & HYDICE & Urban & 612 & 42.70 & BayArea & AVIRIS & Urban & 6000 & 29.39 \\
PaviaC & ROSIS & Urban & 768 & 38.80 & HongHu & Headwall & Nature & 565 & 27.47 \\
PaviaU & ROSIS & Urban & 707 & 36.30 & Nilifossae & CRISM & Mars & 450 & 26.56 \\
KSC & AVIRIS & Nature & 676 & 35.50 & Indianpines & AVIRIS & Agriculture & 450 & 24.43 \\
Urban & HYDICE & Urban & 747 & 33.88 & Utopia & CRISM & Mars & 150 & 22.05 \\
Botswana & EO-1 & Nature & 518 & 32.71 &  &  &  &  &  \\
\bottomrule
\end{tabular}}
\end{table*}

\begin{table*}[!t]
\centering
\scriptsize
\setlength{\tabcolsep}{2pt}
\caption{\textbf{Full sensor-by-task breakdown for the same 13-model VSR$^2$ subset.} Each task cell first averages model accuracy within a scene and then gives equal weight to scenes from that sensor; --- means that the task is absent. The pooled column instead weights all questions. Sensor and scene type are not disentangled here; VNIR and CASI-1500 each occur in one scene.}
\label{tab:r2_sensor_task}
\resizebox{\textwidth}{!}{%
\begin{tabular}{lrrrrrrrrrrrrrrrr}
\toprule
Sensor & Scenes & $N$ & Pooled & SFR & LCC & PD & Count & OLR & RD & SAD & SU & VHS & EPS & BCI & CAL & CSA \\
\midrule
VNIR & 1 & 3003 & 65.22 & 64.5 & 50.5 & 57.7 & 33.7 & 74.7 & 55.8 & 81.1 & 69.4 & 70.8 & 91.8 & 60.1 & --- & --- \\
CASI-1500 & 1 & 1002 & 52.93 & 58.3 & 50.7 & 44.2 & 36.9 & 50.2 & 49.9 & 43.3 & 48.8 & 76.8 & 76.0 & --- & --- & 33.3 \\
HYDICE & 2 & 1359 & 37.85 & 40.6 & 28.9 & 39.2 & 27.2 & 22.9 & 23.9 & 40.3 & 26.2 & 80.7 & 59.1 & --- & --- & --- \\
ROSIS & 2 & 1475 & 37.60 & 42.8 & 29.8 & 50.1 & 31.5 & 31.4 & 25.8 & 26.5 & 37.7 & 78.1 & 26.0 & --- & --- & --- \\
EO-1 & 2 & 902 & 37.22 & 30.8 & 27.8 & 45.0 & 24.8 & 27.7 & 29.1 & 42.7 & 19.7 & 34.8 & 13.7 & 53.5 & 25.5 & 50.9 \\
AVIRIS & 5 & 9026 & 33.80 & 41.0 & 23.7 & 44.3 & 31.2 & 29.5 & 38.1 & 55.4 & 43.6 & 40.2 & 39.3 & 30.2 & 19.9 & 26.9 \\
Headwall & 3 & 1680 & 29.28 & 28.0 & 41.1 & 33.1 & 30.9 & 23.1 & 23.2 & 26.9 & 39.7 & 24.0 & 28.0 & --- & --- & --- \\
CRISM & 3 & 890 & 27.17 & 29.8 & 47.5 & 18.2 & 16.5 & 30.4 & 20.8 & 29.2 & 26.3 & --- & 26.5 & --- & --- & --- \\
\bottomrule
\end{tabular}}
\end{table*}

\begin{table*}[!t]
\centering
\small
\setlength{\tabcolsep}{8pt}
\caption{\textbf{Task accuracy and scene variation for the 13-model full VSR$^2$ subset.} Scene-equal accuracy averages task accuracy over scenes supporting that task. Pooled accuracy weights questions. The 95\% intervals use a Student $t$ interval over scene accuracies, treating scenes as the unit; intervals are omitted where fewer than five scenes support a task.}
\label{tab:r2_task_scene}
\begin{tabular}{lrrrrr}
\toprule
Task & Scenes & $N$ & Scene-equal & Pooled & 95\% scene CI \\
\midrule
SFR & 17 & 2328 & 39.34 & 41.05 & [29.8, 48.9] \\
LCC & 17 & 1629 & 34.29 & 27.61 & [28.1, 40.4] \\
PD & 17 & 2970 & 38.62 & 40.68 & [31.3, 46.0] \\
Count & 18 & 1429 & 28.85 & 36.76 & [24.5, 33.2] \\
OLR & 17 & 1909 & 31.75 & 29.23 & [24.1, 39.4] \\
RD & 18 & 1869 & 31.39 & 33.67 & [23.2, 39.6] \\
SAD & 16 & 1329 & 42.18 & 57.88 & [30.2, 54.2] \\
SU & 16 & 1353 & 39.32 & 38.95 & [30.1, 48.6] \\
VHS & 15 & 1766 & 51.53 & 51.47 & [37.2, 65.9] \\
EPS & 15 & 1296 & 41.29 & 56.01 & [25.7, 56.9] \\
BCI & 4 & 751 & 43.49 & 37.09 & --- \\
CAL & 3 & 588 & 21.80 & 16.30 & --- \\
CSA & 3 & 120 & 37.04 & 38.46 & --- \\
\midrule
Overall & 19 & 19337 & 36.66 & 39.71 & [30.8, 42.5] \\
\bottomrule
\end{tabular}
\end{table*}

\begin{table*}[!t]
\centering
\scriptsize
\setlength{\tabcolsep}{2pt}
\caption{\textbf{Complete task-wise report-wording ablation on three MLLMs.} The ablated reports retain their structure and all quantitative fields but omit generated interpretive descriptions. Values are accuracies (\%); $\Delta$ is ablated minus original overall accuracy. Original task-wise results for the same model--input pairs are in Table~\ref{tab:full_results}.}
\label{tab:r2_report_ablation}
\resizebox{\textwidth}{!}{%
\begin{tabular}{llrrrrrrrrrrrrrrrr}
\toprule
Model & Input & SFR & LCC & PD & Count & OLR & RD & SAD & SU & VHS & EPS & BCI & CAL & CSA & Original & Ablated & $\Delta$ \\
\midrule
InternVL2-8B & RGB+Report & 38.70 & 26.64 & 35.62 & 24.56 & 29.54 & 29.27 & 60.35 & 40.50 & 40.83 & 68.21 & 35.55 & 15.31 & 38.33 & 37.90 & 37.19 & -0.71 \\
 & VSR$^2$ & 42.27 & 28.84 & 38.88 & 25.65 & 29.52 & 28.15 & 66.21 & 43.25 & 41.94 & 68.50 & 39.61 & 14.04 & 38.93 & 39.12 & 38.91 & -0.21 \\
\addlinespace[2pt]
Qwen2.5-VL-7B & RGB+Report & 38.62 & 23.94 & 30.84 & 23.16 & 31.48 & 40.66 & 62.98 & 37.77 & 58.89 & 49.31 & 41.28 & 18.20 & 30.00 & 37.88 & 37.47 & -0.41 \\
 & VSR$^2$ & 40.13 & 26.56 & 35.32 & 29.63 & 29.54 & 38.30 & 63.15 & 41.89 & 60.93 & 55.42 & 43.14 & 18.30 & 30.20 & 39.41 & 39.42 & +0.01 \\
\addlinespace[2pt]
Qwen3-VL-4B & RGB+Report & 36.81 & 25.11 & 33.97 & 39.75 & 29.91 & 40.98 & 63.13 & 37.32 & 53.00 & 48.23 & 42.74 & 17.18 & 28.33 & 39.34 & 38.19 & -1.15 \\
 & VSR$^2$ & 43.05 & 30.40 & 35.68 & 39.48 & 34.12 & 45.57 & 68.79 & 42.43 & 58.13 & 53.54 & 48.87 & 17.95 & 36.50 & 43.23 & 42.65 & -0.58 \\
\bottomrule
\end{tabular}}
\end{table*}

\subsection{Source-Stratified Results}
\label{app:source_stratification}

The 19,337 questions arise from tiles and multiple questions within the same original scene, so treating every question as an independent replicate understates uncertainty across sources. We regroup the 23 source datasets into 19 original-scene groups, merging acquisitions of the same scene across years, and aggregate the archived full VSR$^2$ predictions of 13 models (251,381 model--question observations). These 13 models form a defined subset of the 17-model six-view comparison; scene identifiers are not retained in the available six-view aggregate table, so the source-level analysis is confined to the archived full-view predictions.

Table~\ref{tab:r2_scene} reports scene-level results, Table~\ref{tab:r2_sensor_task} separates tasks by sensor, and Table~\ref{tab:r2_task_scene} summarizes variation across scenes within each task. The question-pooled accuracy is 39.71\%, whereas the average of the 19 scene-level pooled accuracies is 36.66\%. Treating the scene as the unit yields a Student-$t$ 95\% interval of [30.8, 42.5]\% around this scene-equal mean. The per-task scene-equal scores use a separate denominator consisting only of scenes that support each task.

The source gradient is substantial: pooled accuracy ranges from 65.22\% on Xiongan to 22.05\% on Utopia. With LongKou classified as urban and HongHu as nature, agriculture reaches 56.4\% when questions are pooled by scene category, compared with 32.1\% for nature scenes, 33.7\% for urban scenes, and 27.2\% for Martian scenes. This heterogeneity makes the benchmark diagnostic of where current models struggle, while the gap between pooled and scene-equal scores shows that the overall score is sensitive to source composition. Table~\ref{tab:r2_sensor_task} shows that sensor-level pooled accuracy ranges from 65.22\% for VNIR to 27.17\% for CRISM, but this breakdown is descriptive rather than a causal sensor comparison: VNIR and CASI-1500 each appear in one scene, and CRISM appears only in the Martian scenes. Task coverage is also uneven, with only three scene groups supporting CSA and three supporting CAL. The interval describes variation over the included scenes and does not estimate transfer to unseen sensors. Accordingly, the scene interval and support counts should accompany generalization claims.

\subsection{Report Descriptive-Wording Ablation}
\label{app:report_ablation}

To test whether natural-language interpretation drives the report results, we remove the LLaVA-generated descriptive wording while retaining report structure and all quantitative fields from the intermediate features, including proxy indices, regional statistics, and PCA summaries. Table~\ref{tab:r2_report_ablation} gives the complete 13-task results for RGB+Report and full VSR$^2$ on three representative MLLMs. Across these models, the mean change is $-0.76$ percentage points for RGB+Report and $-0.26$ for full VSR$^2$. The task-level changes are bidirectional: counting declines while region delineation improves on average. Descriptive wording can make legitimate statistics easier to use, but can also introduce task-dependent cues. Therefore, a small overall change is evidence that the main result is not dominated by the removed wording in these three models, rather than proof that every remaining report field is free of indirect answer cues.

\appsection{Limitations and Future Work}
\label{app:limitations}

Although HM-Bench and VSR$^2$ provide a systematic starting point for studying hyperspectral understanding with multimodal large language models (MLLMs), they still have several limitations in benchmark coverage, representation fidelity, and evaluation scope. We view the current benchmark and framework as an initial step toward more general and native HSI understanding.

\subsection{Benchmark Limitations}

HM-Bench covers a diverse set of hyperspectral reasoning tasks, but it is still limited in scale and domain breadth compared with the full complexity of real-world HSI applications. The current benchmark focuses on 13 task categories and a controlled evaluation protocol, which is useful for standardized comparison, but it cannot fully represent the wide range of sensor types, acquisition conditions, geographic regions, and application scenarios encountered in practice.

In addition, although our task design aims to cover multiple forms of spectral-spatial reasoning, some real deployment settings may require more specialized knowledge, longer reasoning chains, or finer-grained annotations than those currently provided. Therefore, the present version of HM-Bench should be regarded as a strong initial benchmark rather than a complete evaluation suite for hyperspectral intelligence.

\subsection{Framework Limitations}

VSR$^2$ is designed as a training-free and model-agnostic framework, which makes it practical for evaluating existing MLLMs. However, this design also introduces an inherent limitation: current general-purpose MLLMs do not directly consume raw hyperspectral cubes. Instead, they rely on transformed views, including pseudo-RGB renderings, PCA-based visualizations, and structured textual reports. While these views make hyperspectral data accessible to current MLLMs, they inevitably compress or discard part of the original spectral information.

Moreover, the effectiveness of multi-view prompting still depends on the reasoning behavior of the underlying model. As shown in our experiments, additional views do not always lead to monotonic gains for every model, suggesting that current MLLMs may not yet fully calibrate or reconcile heterogeneous evidence from different hyperspectral representations. Thus, VSR$^2$ should be understood as a practical bridging framework rather than a native solution to hyperspectral modeling.

\subsection{Future Directions}

An important direction for future work is to move from indirect HSI understanding toward direct hyperspectral modeling. In particular, we plan to explore specialized hyperspectral encoders that can process raw HSI cubes directly and interface with MLLMs through learned visual or multimodal token spaces. Such an encoder could preserve richer spectral-spatial information than hand-crafted view transformations and may enable more faithful reasoning over the original data.

Beyond encoder design, future work may also expand HM-Bench along several axes, including larger benchmark scale, broader task diversity, more challenging expert-level reasoning scenarios, and more realistic evaluation settings. We are also interested in studying whether lightweight adaptation, instruction tuning, or cross-modal alignment strategies can further improve the use of hyperspectral evidence in MLLMs.

More broadly, we hope this work can motivate future research on native hyperspectral foundation models and on general multimodal systems capable of supporting downstream applications such as environmental monitoring, land-cover analysis, precision agriculture, resource surveying, and disaster assessment.

\FloatBarrier
\appsection{Case Study}
\label{app:casestudy}
We show representative \textbf{InternVL3.5-14B}~\citep{chen2024internvl} predictions under PCA-only, Report-only, RGB-only, and full VSR$^2$. Figures~\ref{fig:case1}, \ref{fig:case2}, \ref{fig:case3}, \ref{fig:case4}, and~\ref{fig:case5} illustrate how selected options can vary across views; task-level accuracies are reported in Table~\ref{tab:full_results}.

\noindent
\begin{minipage}{\linewidth}
\centering
\includegraphics[width=0.9\columnwidth]{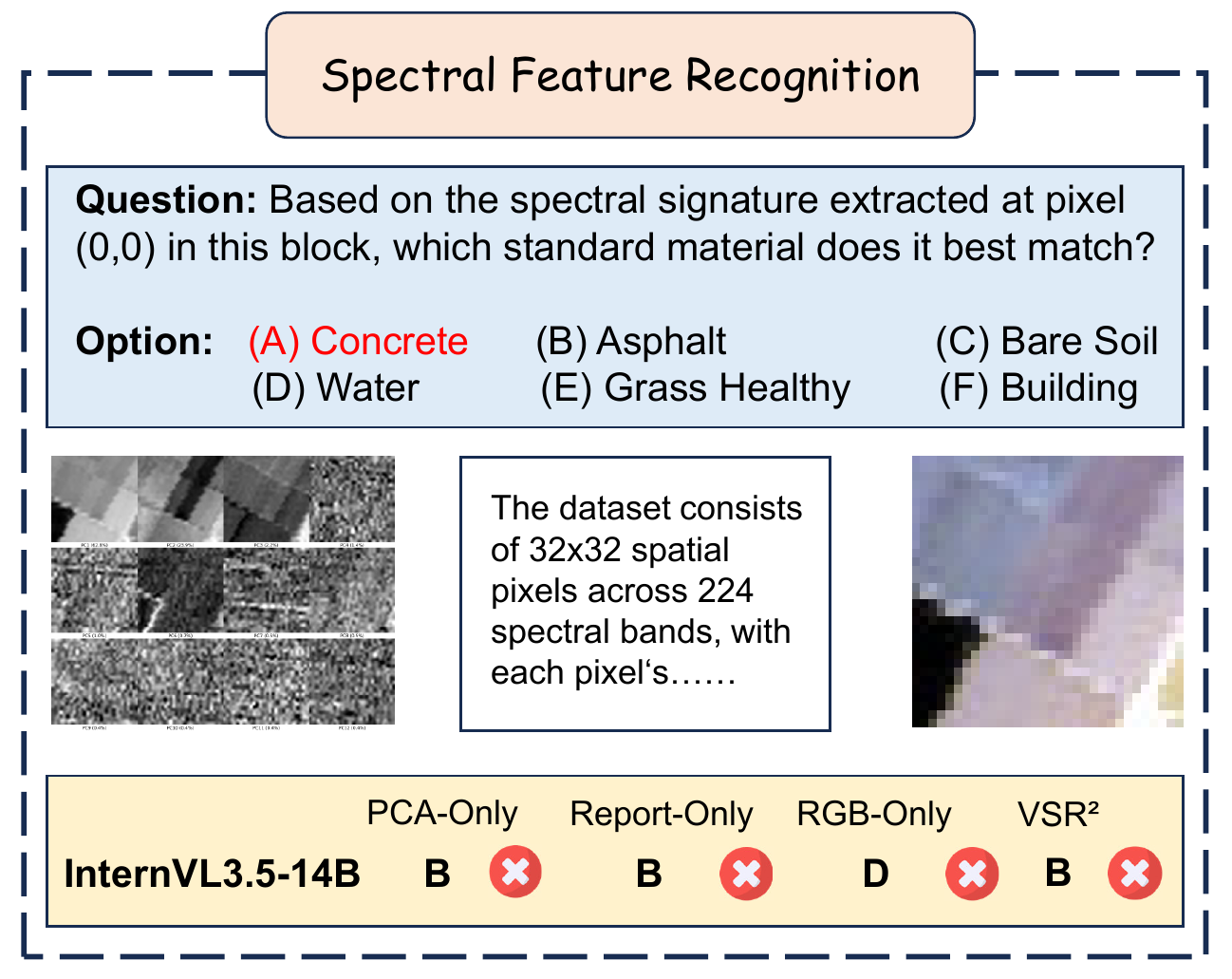}
\captionof{figure}{A question case of the Spectral Feature Recognition}
\label{fig:case1}
\end{minipage}

\vspace{2em}

\noindent
\begin{minipage}{\linewidth}
\centering
\includegraphics[width=0.9\columnwidth]{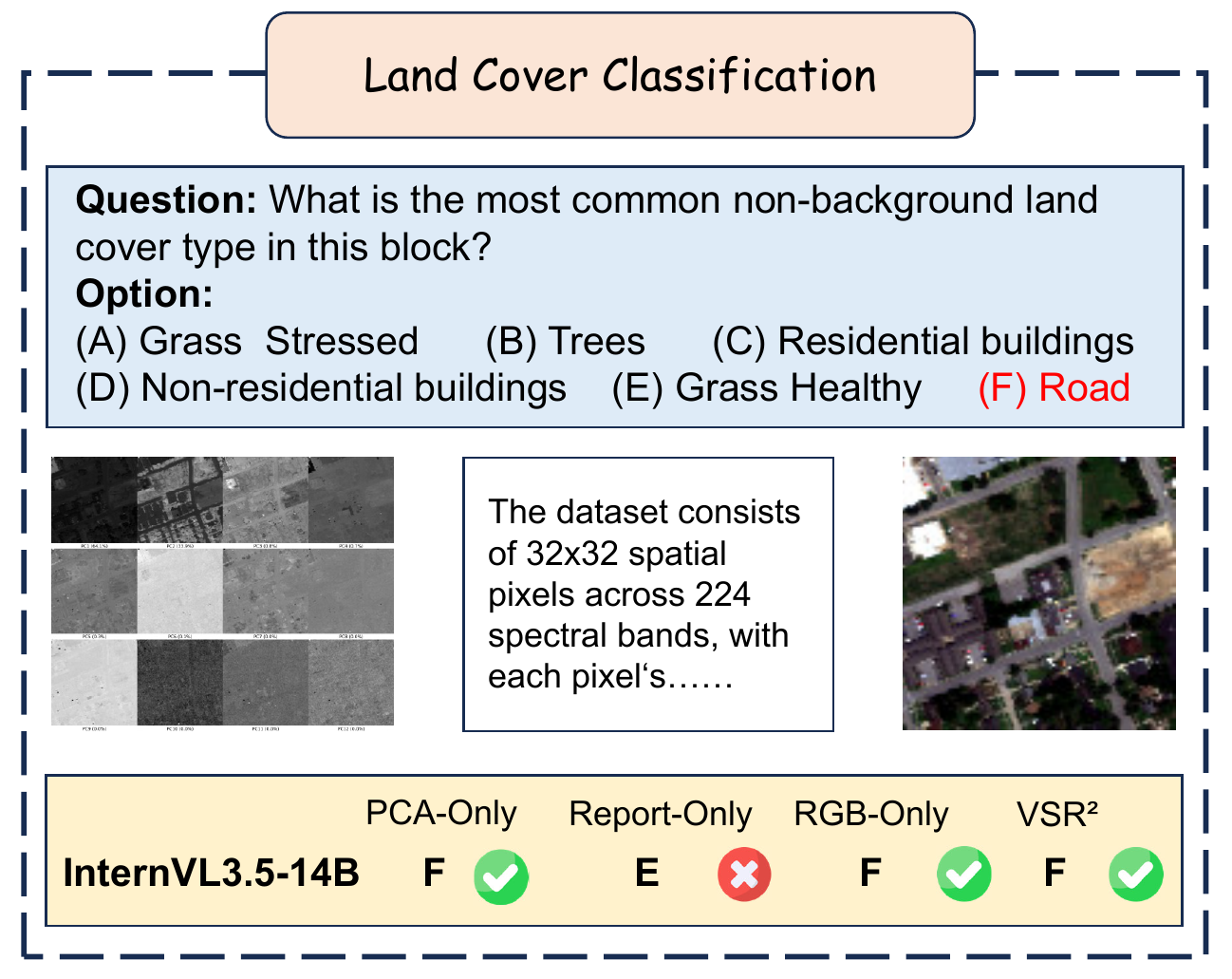}
\captionof{figure}{A question case of the Land Cover Classification }
\label{fig:case2}
\end{minipage}

\vspace{2em}

\noindent
\begin{minipage}{\linewidth}
\centering
\includegraphics[width=0.9\columnwidth]{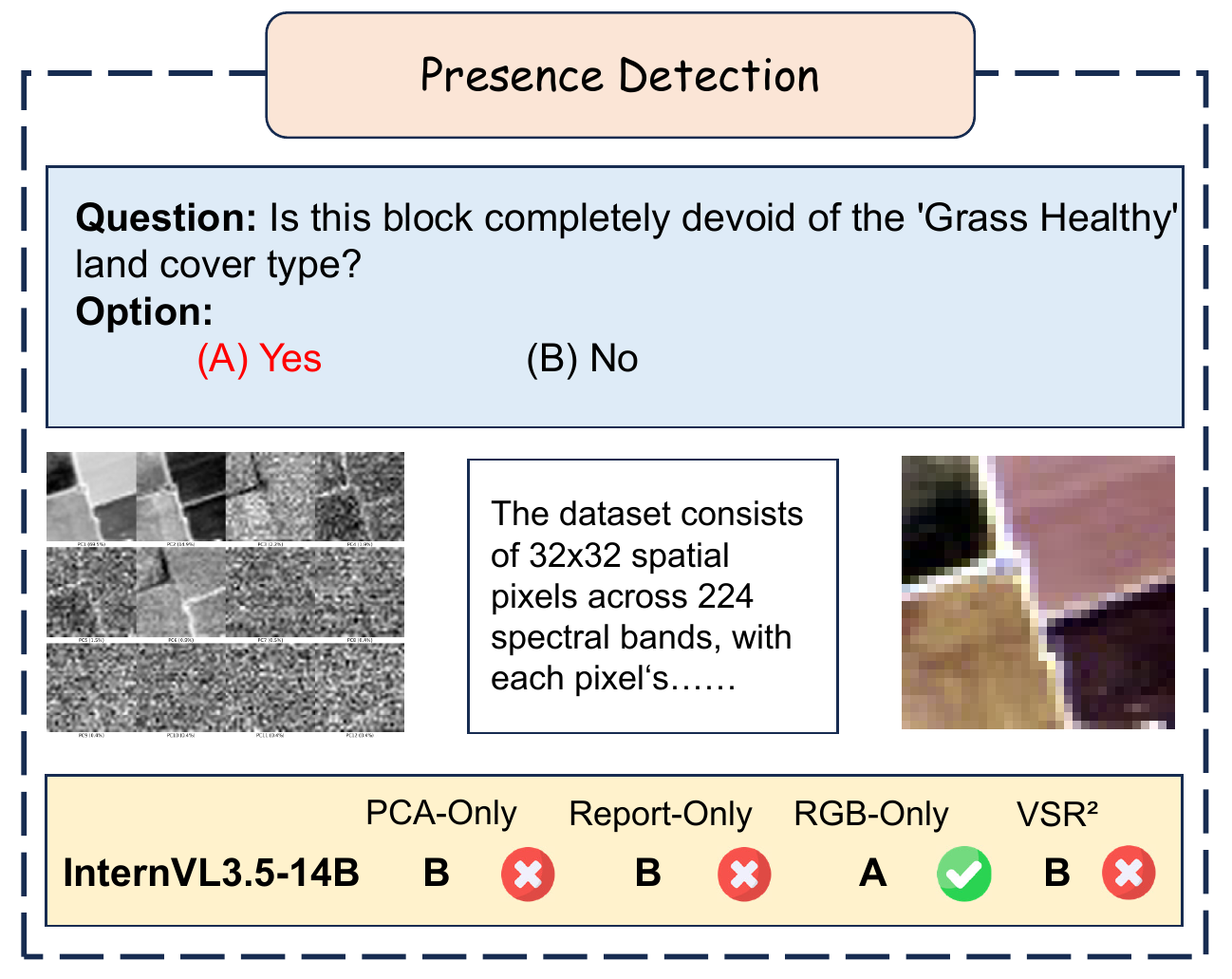}
\captionof{figure}{A question case of the Presence Detection}
\label{fig:case3}
\end{minipage}

\vspace{2em}

\noindent
\begin{minipage}{\linewidth}
\centering
\includegraphics[width=0.9\columnwidth]{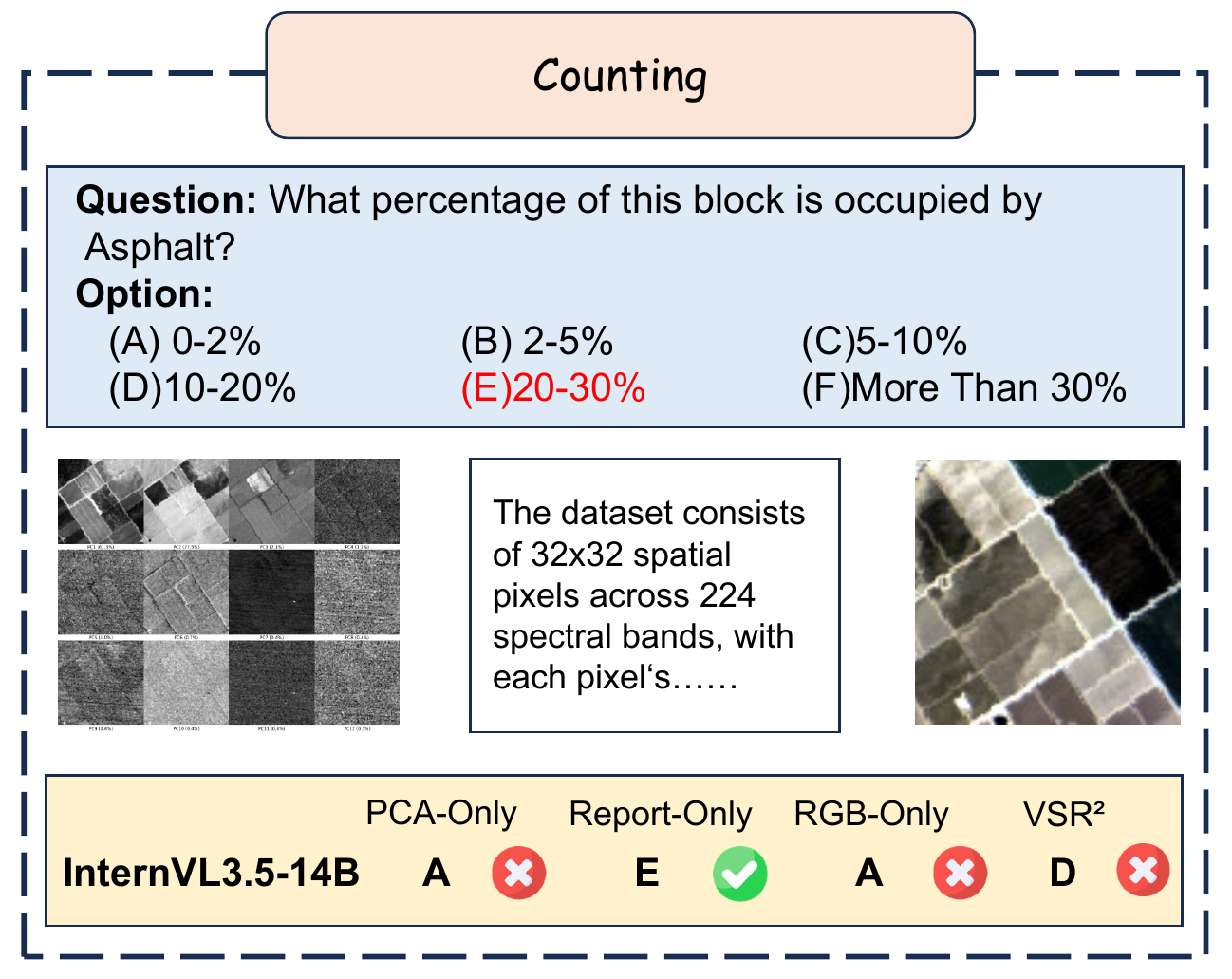}
\captionof{figure}{A question case of the Counting}
\label{fig:case4}
\end{minipage}

\vspace{2em}

\noindent
\begin{minipage}{\linewidth}
\centering
\includegraphics[width=0.9\columnwidth]{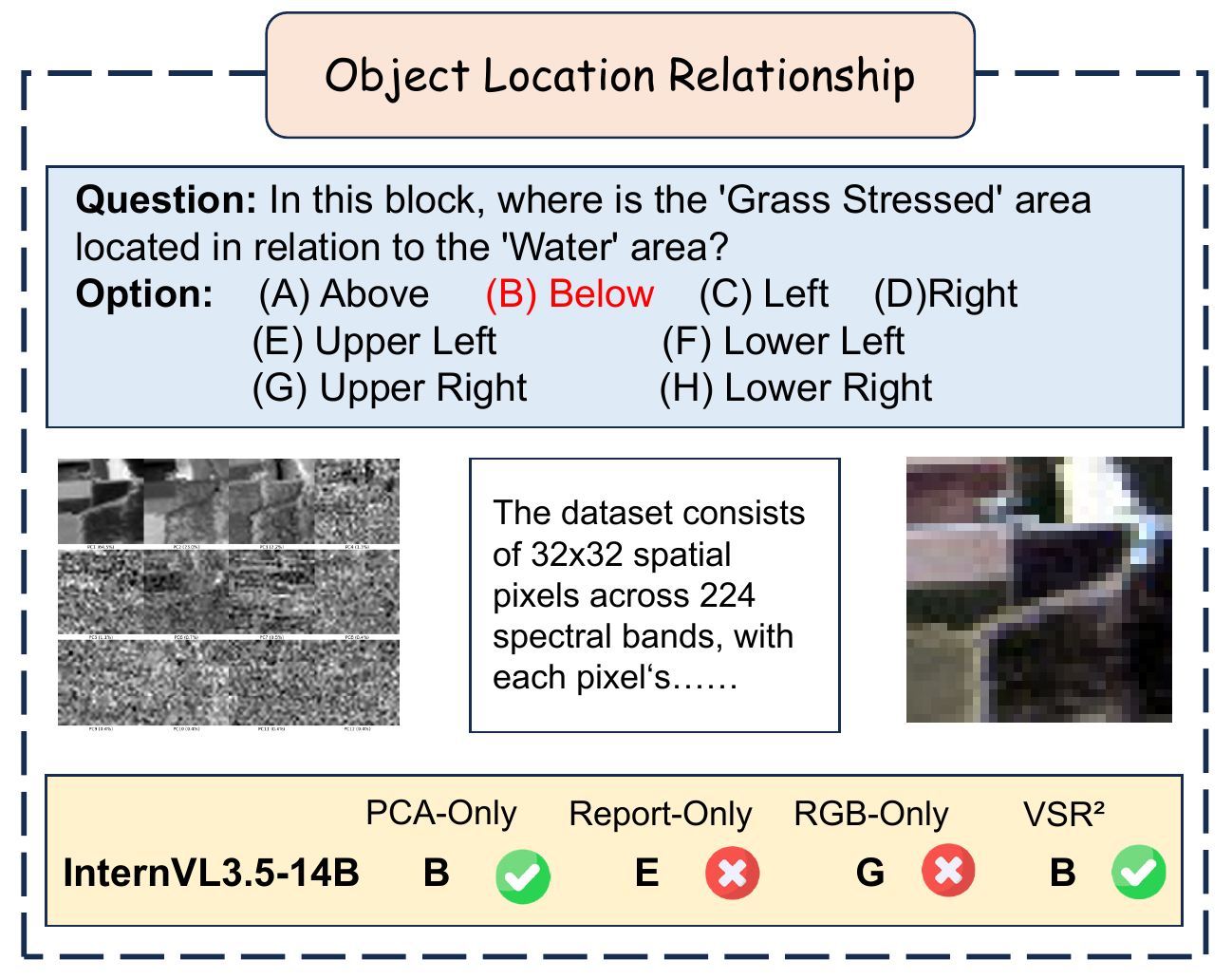}
\captionof{figure}{A question case of the Object Location Relationship}
\label{fig:case5}
\end{minipage}

\vspace{2em}

\noindent
\begin{minipage}{\linewidth}
\centering
\includegraphics[width=0.9\columnwidth]{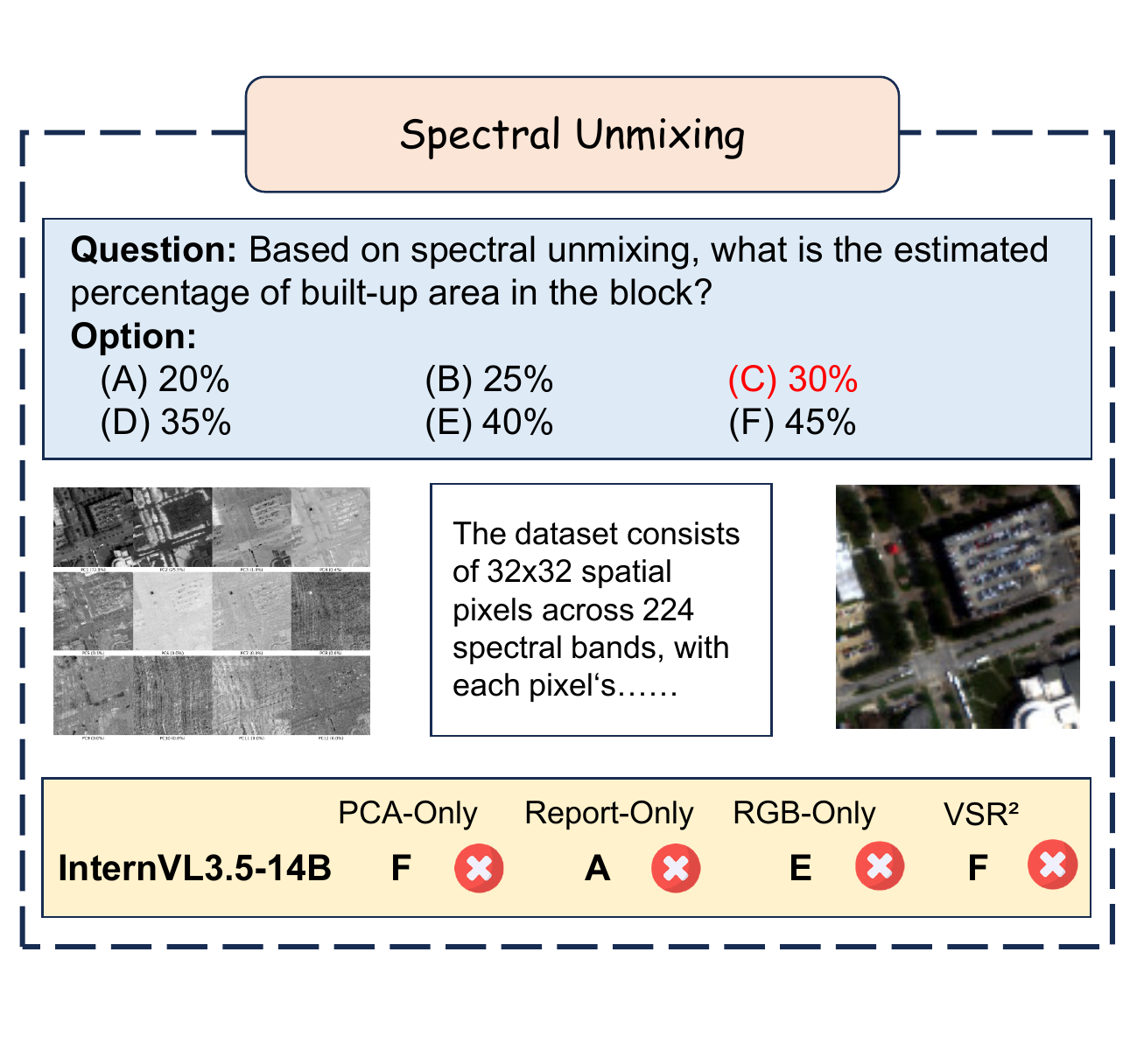}
\vspace{-2em}
\captionof{figure}{A question case of the Spectral Unmixing }
\label{fig:case8}
\end{minipage}

Figure~\ref{fig:case8} shows a spectral-unmixing question. Two related cases illustrate the task's range: a BayArea\_2015 boundary pixel (block 559) has \emph{Grass Stressed} and \emph{Grass Healthy} as its reference endmembers, while a Houston\_13 block (block 0) has a reference \emph{Road} abundance of approximately 0.68\%. Figures~\ref{fig:case9}, \ref{fig:case6}, \ref{fig:case10}, and~\ref{fig:case7} give further examples from other tasks, with both varying and consistent options across views.

\vspace{2em}

\noindent
\begin{minipage}{\linewidth}
\centering
\includegraphics[width=0.8\columnwidth]{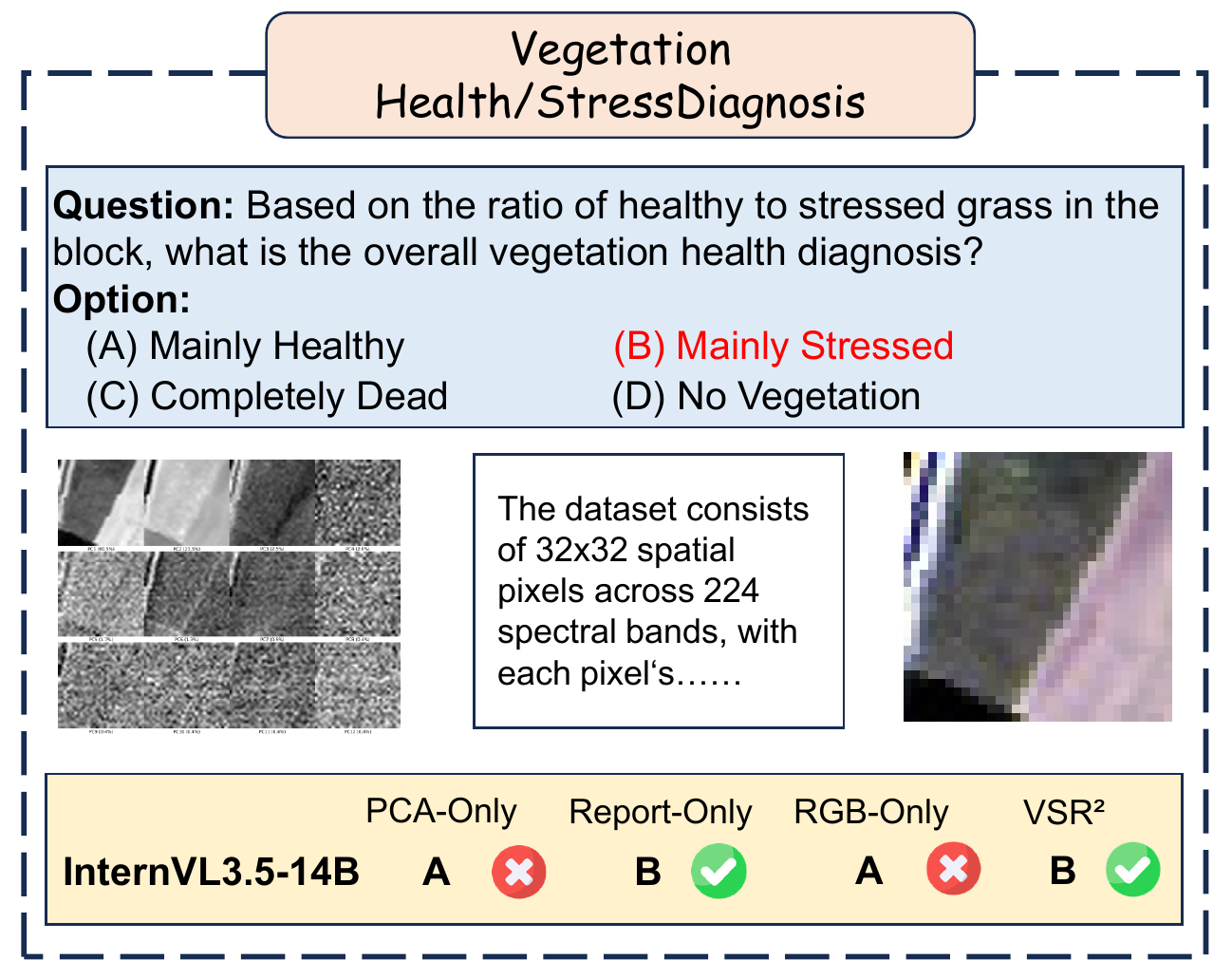}
\captionof{figure}{A question case of the Vegetation Health/Stress Diagnosis }
\label{fig:case9}
\end{minipage}

\vspace{2em}

\noindent
\begin{minipage}{\linewidth}
\centering
\includegraphics[width=0.8\columnwidth]{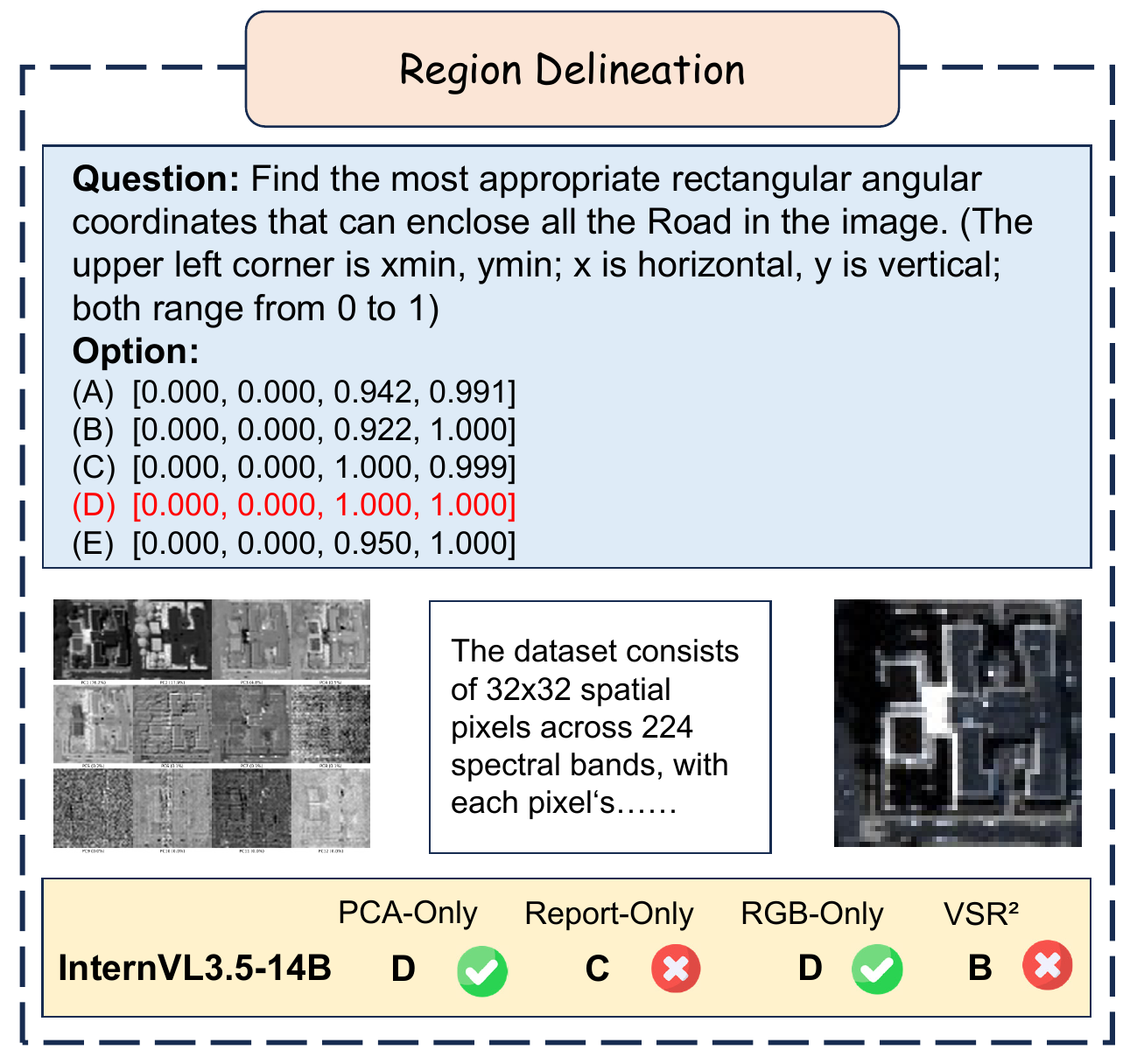}
\captionof{figure}{A question case of the Region Delineation }
\label{fig:case6}
\end{minipage}
\vspace{2em}

\noindent
\begin{minipage}{\linewidth}
\centering
\includegraphics[width=0.8\columnwidth]{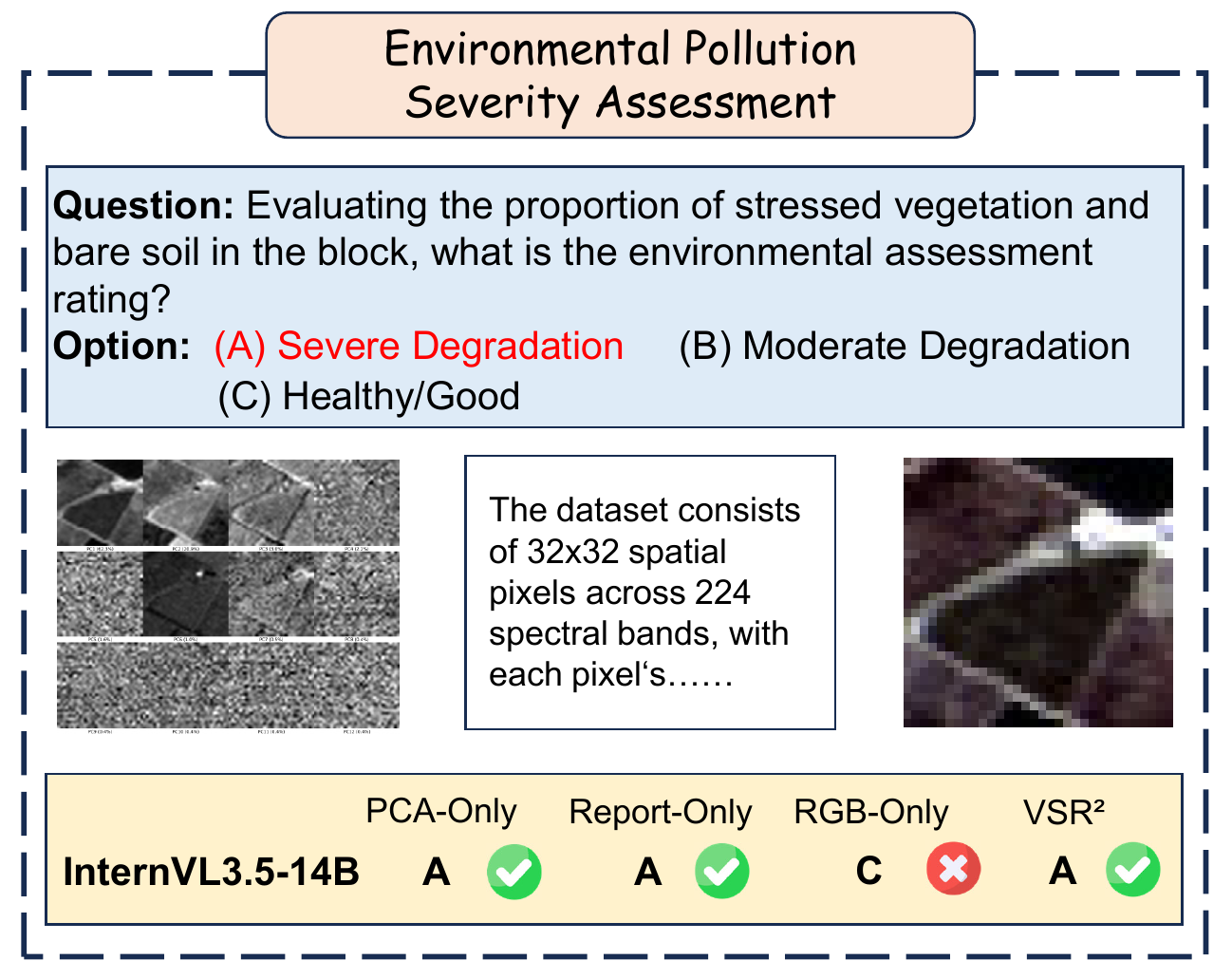}
\captionof{figure}{A question case of the Environmental Pollution Severity Assessment }
\label{fig:case10}
\end{minipage}
\vspace{2em}

\noindent
\begin{minipage}{\linewidth}
\centering
\includegraphics[width=0.8\columnwidth]{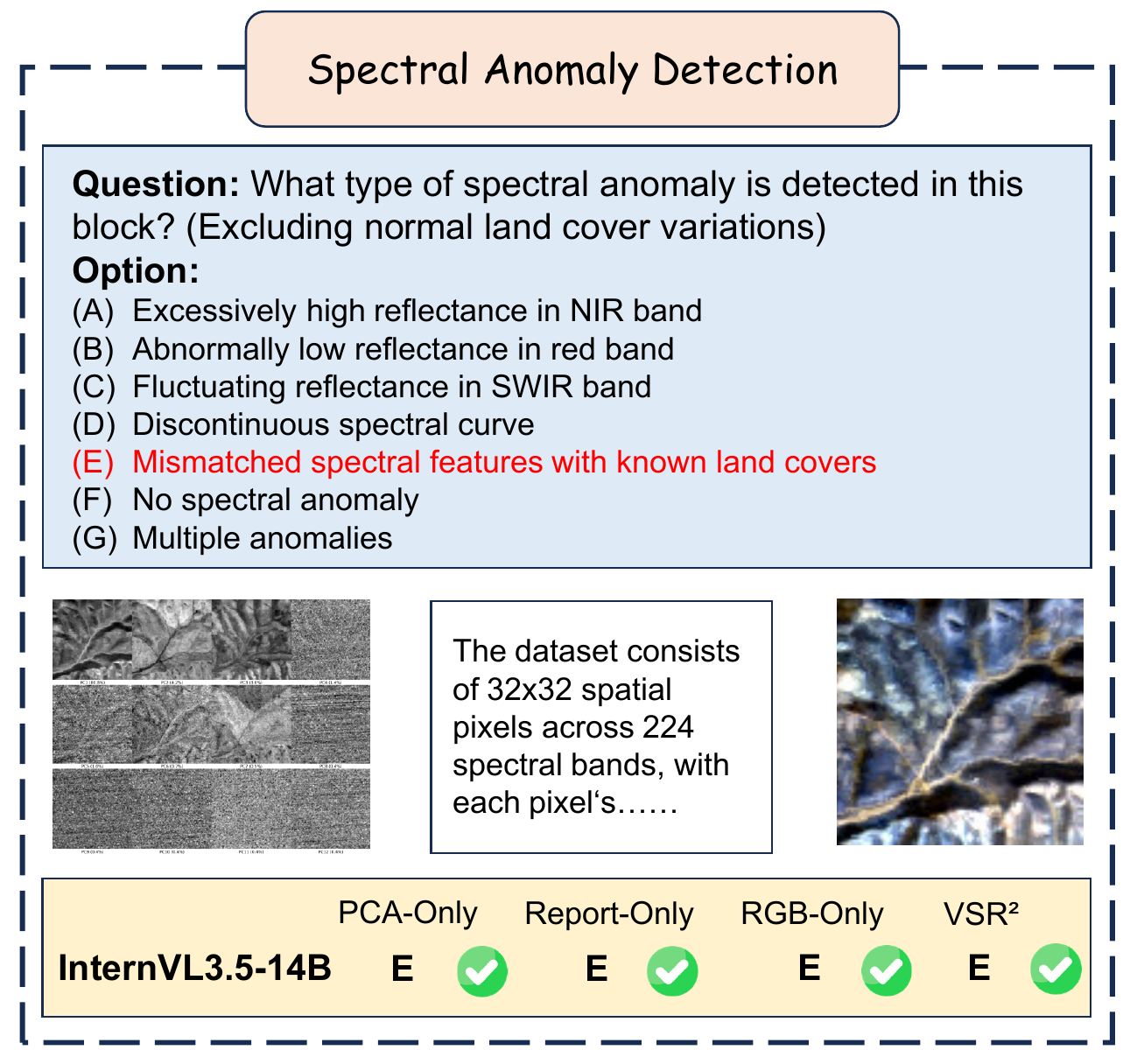}
\captionof{figure}{A question case of the Spectral Anomaly Detection}
\label{fig:case7}
\end{minipage}

\vspace{2em}

Figures~\ref{fig:case11}, \ref{fig:case12}, and~\ref{fig:case13} show the three change-detection tasks and their predictions under the four inputs.

\noindent
\begin{minipage}{\linewidth}
\centering
\includegraphics[width=0.8\columnwidth]{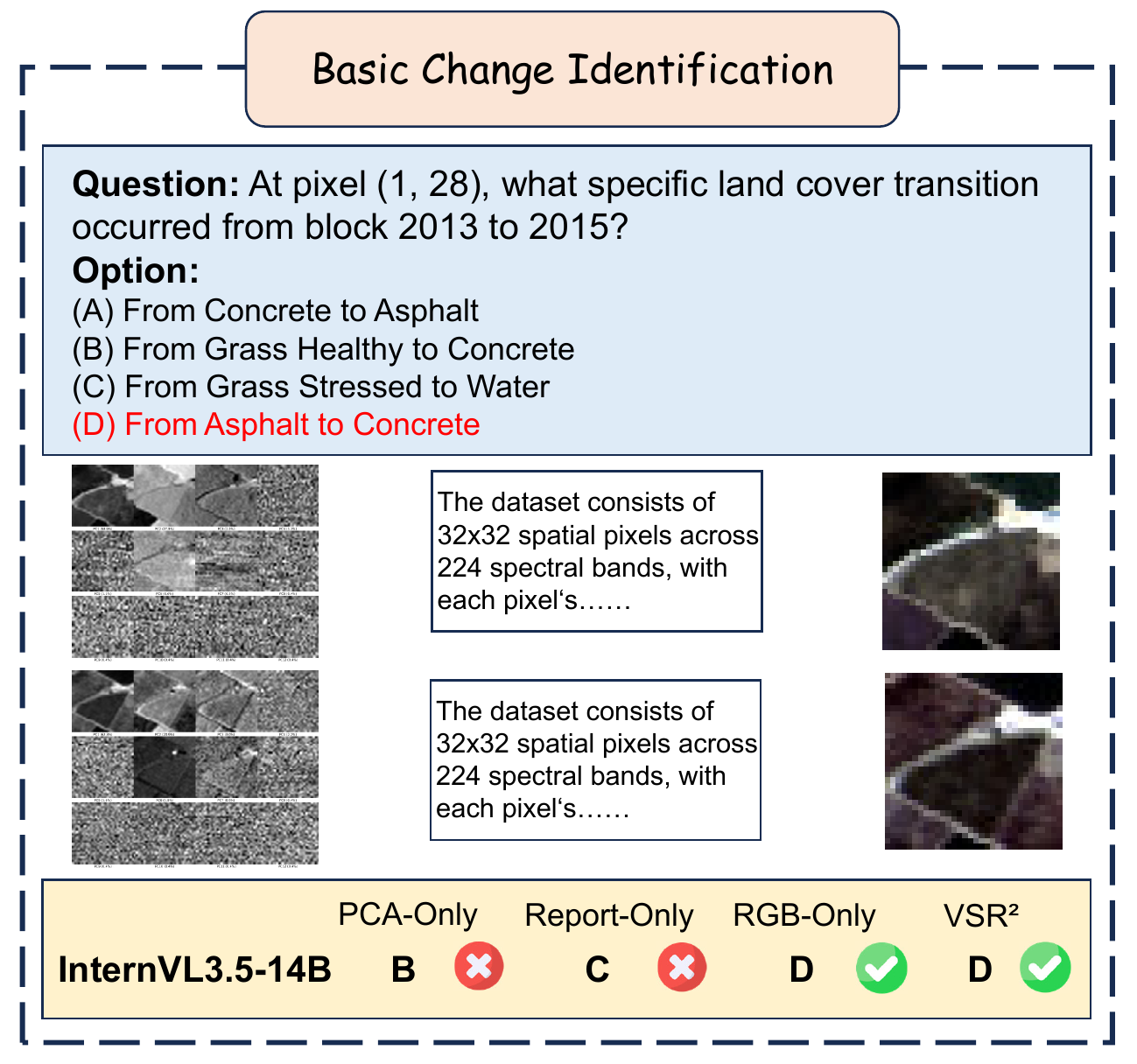}
\captionof{figure}{A question case of the Basic Change Identification}
\label{fig:case11}
\end{minipage}

\vspace{2em}

\noindent
\begin{minipage}{\linewidth}
\centering
\includegraphics[width=0.8\columnwidth]{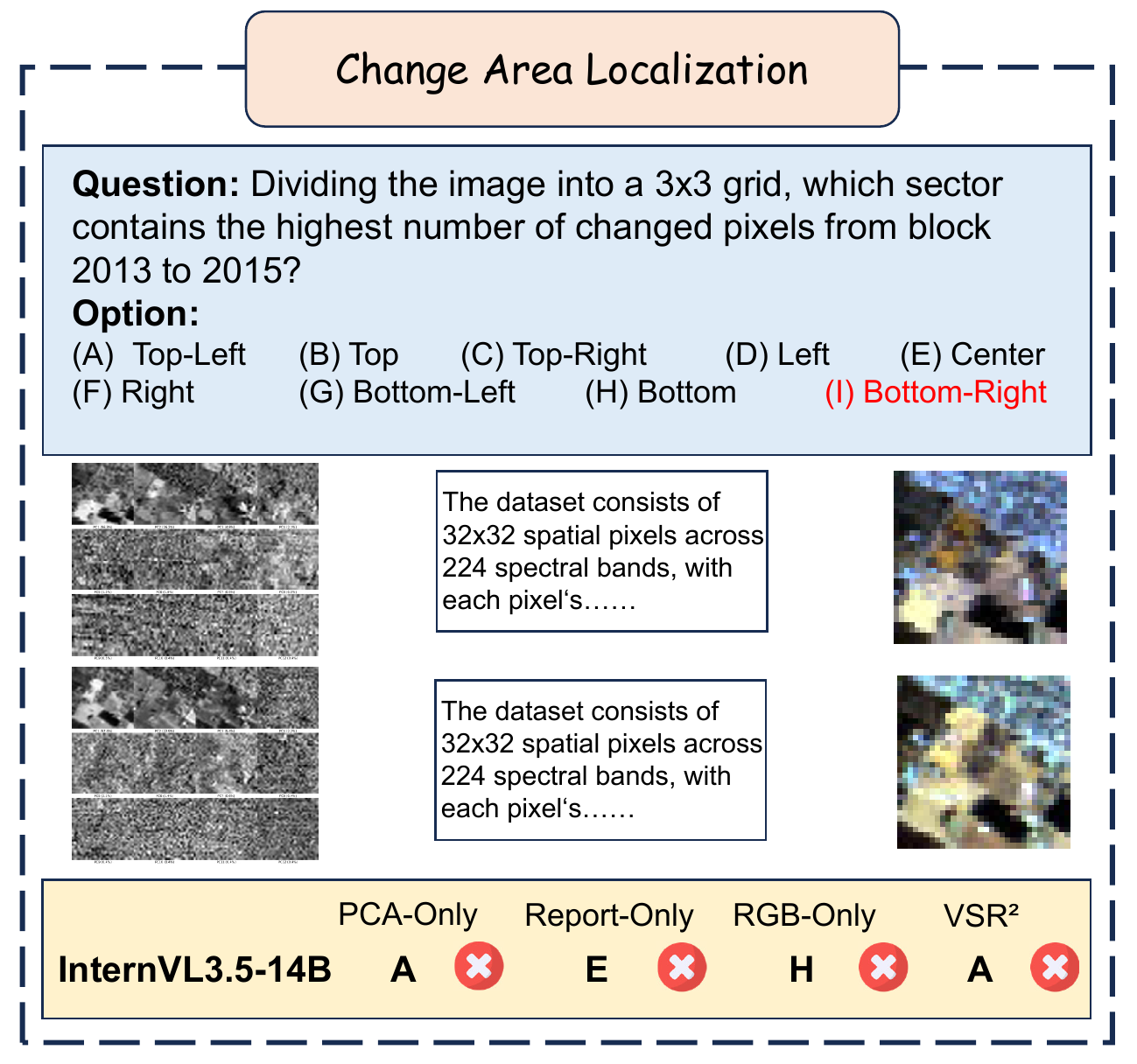}
\captionof{figure}{A question case of the Change Area Localization }
\label{fig:case12}
\end{minipage}

\vspace{2em}

\noindent
\begin{minipage}{\linewidth}
\centering
\includegraphics[width=0.9\columnwidth]{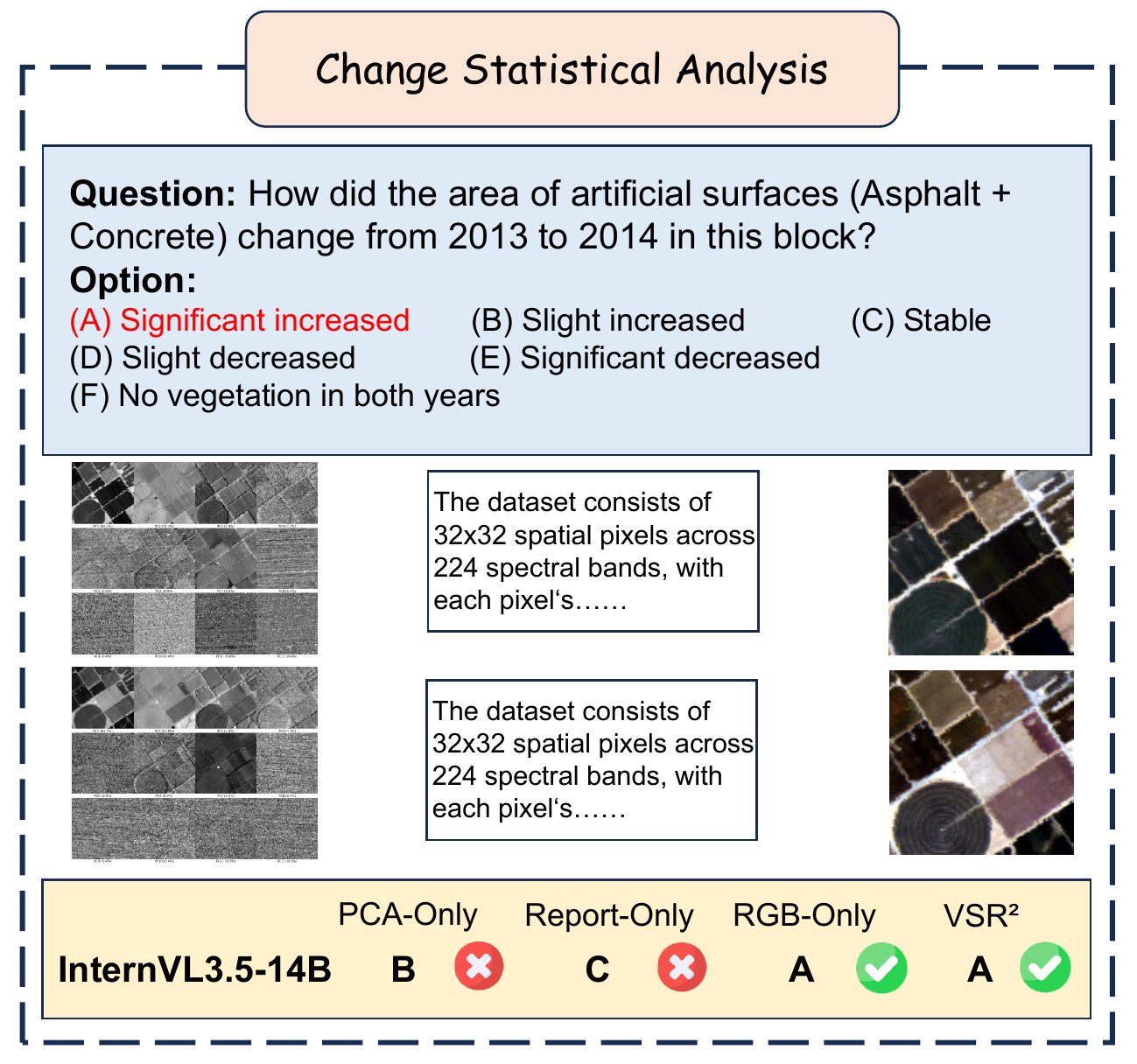}
\captionof{figure}{A question case of the Change Statistical Analysis}
\label{fig:case13}
\end{minipage}

\end{document}